\documentclass[mnsc,nonblindrev]{informs3_hide}
\OneAndAHalfSpacedXI % default for MS

\usepackage[english]{babel}
\usepackage[autostyle, english = american]{csquotes}
\MakeOuterQuote{"}
\usepackage[colorlinks,linkcolor=blue, citecolor=blue]{hyperref}
\usepackage[normalem]{ulem}

\usepackage{cleveref}
\crefname{subsection}{section}{subsections}
\usepackage{eqnarray}
\usepackage{algorithm,algpseudocode}
\usepackage{amsmath, bbm, enumitem, xparse, bm, lipsum}
\usepackage{amssymb}
\usepackage{comment}
\usepackage{mathtools}
\usepackage{subcaption}
\usepackage{booktabs}

\newcommand{\eps}{\varepsilon}

\usepackage[dvipsnames]{xcolor}

\usepackage{xparse}

\NewDocumentEnvironment{myproof}{o}
{\IfNoValueTF{#1}{\paragraph{{Proof.} }} {\paragraph{{#1.} }} }
{\hfill$\Halmos$}

\usepackage{natbib,bbm,multirow,multicol}
 \bibpunct[, ]{(}{)}{,}{a}{}{,}%
 \def\bibfont{\small}%
\TheoremsNumberedThrough     % Preferred (Theorem 1, Lemma 1, Theorem 2)
\ECRepeatTheorems

\EquationsNumberedThrough    % Default: (1), (2), ...
\allowdisplaybreaks

\begin{document}
%%%%%%%%%%%%%%%%

%\setcounter{page}{1}
%\DOI{10.1287/xxxx.0000.0000}%

\RUNAUTHOR{Zong, Jiang}

\RUNTITLE{Online Semi-infinite Linear Programming}

\TITLE{
Online Semi-infinite Linear Programming: Efficient Algorithms via Function Approximation
}

\ARTICLEAUTHORS{%
\AUTHOR{$\text{Yiming Zong}^{\dag}$, $\text{Jiashuo Jiang}^{\dag}$}

\AFF{\  \\
$\dag~$Department of Industrial Engineering \& Decision Analytics, Hong Kong University of Science and Technology
}
}

\ABSTRACT{
We consider the dynamic resource allocation problem where the decision space is finite-dimensional, yet the solution must satisfy a large or even infinite number of constraints revealed via streaming data or oracle feedback. We model this challenge as an Online Semi-infinite Linear Programming (OSILP) problem and develop a novel LP formulation to solve it approximately. Specifically, we employ function approximation to reduce the number of constraints to a constant $q$. This addresses a key limitation of traditional online LP algorithms, whose regret bounds typically depend on the number of constraints, leading to poor performance in this setting. We propose a dual-based algorithm to solve our new formulation, which offers broad applicability through the selection of appropriate potential functions. We analyze this algorithm under two classical input models—stochastic input and random permutation—establishing regret bounds of $O(q\sqrt{T})$ and $O\left(\left(q+q\log{T})\sqrt{T}\right)\right)$ respectively. Note that both regret bounds are independent of the number of constraints, which demonstrates the potential of our approach to handle a large or infinite number of constraints. Furthermore, we investigate the potential to improve upon the $O(q\sqrt{T})$ regret and propose a two-stage algorithm, achieving $O(q\log{T} + q/\eps)$ regret under more stringent assumptions. We also extend our algorithms to the general function setting. A series of experiments validates that our algorithms outperform existing methods when confronted with a large number of constraints.
}

\KEYWORDS{Semi-infinite Programming, Sublinear Regret, Online Optimization, Primal-dual Update}


\maketitle
%%%%%%%%%%%%%%%%%%%%%%%%%%%%%%%%%%%%%%%%%%%%%%%%%%%%%%%%%%%%%%%%%%%%%%

%\subsection{Problem Formulation}

\section{Introduction}

%\subsection{Problem Formulation}

In this paper, we investigate the general problem of Online Semi-Infinite Linear Programming (OSILP) over a finite horizon of discrete time periods. This setting represents a significant generalization of classical online optimization: while the decision variables are finite-dimensional, the constraints they must satisfy are drawn from an infinite or continuous index set, revealed sequentially.
At each time step $t$, the decision maker observes a stochastic input $(r_t, \bm{a}_t)$, drawn independently from an underlying distribution, and must immediately make a decision $x_t$. The primary objective is to maximize the total cumulative reward subject to the revealed constraints.

Our formulation is motivated by the increasing prevalence of high-dimensional streaming data where constraints are not merely numerous but effectively continuous. By bridging the gap between online learning and semi-infinite programming, our framework unifies and extends a broad range of classical models and modern applications:

1. \textbf{Robust and Distributionally Robust Optimization}: Our model naturally encapsulates robust linear programming with uncertainty sets \citep{ben1998robust,bertsimas2004price} and distributionally robust optimization (DRO) problems involving generalized moments \citep{delage2010distributionally,mohajerin2018data}. In these settings, the "infinite" constraints correspond to the worst-case scenarios within an uncertainty set. 

2. \textbf{Spatio-Temporal Resource Allocation}: In continuous facility location and $p$-center--type covering problems \citep{drezner2004facility}, constraints are often spatially indexed over a continuous region. Similarly, in spectrum allocation for wireless networks, interference constraints must be satisfied across a continuous frequency band and spatial domain, which fits our semi-infinite framework.

3. \textbf{Control Systems and Engineering}: Scenario-based methods in control often rely on constraints discovered via streaming data \citep{calafiore2006scenario,campi2008exact}. Our model is applicable to real-time safety-critical control, where a system must maintain safety margins (constraints) across a continuous state space while optimizing performance.

Furthermore, our formulation generalizes the standard Online Linear Programming (OLP) problem, which is recovered when the constraint index set is finite. Following the traditional online LP literature, we analyze our algorithms under two canonical settings: the Stochastic Input Model, where the input data $(r_t,\bm{a}_t)$ are generated i.i.d. from an unknown distribution $\mathcal{P}$, and the Random Permutation Model, where an adversary can fix a finite collection $S = \{(r_t, \bm{a}_t)\}_{t=1}^T$ beforehand and then reveal it in a uniformly random order. We evaluate performance using two metrics: \textit{regret} and \textit{constraint violation}. We adopt the offline optimal solution as our benchmark, which assumes knowledge of the entire sequence $(r_{\pi(t)},\bm{a}_{\pi(t)})$ in advance to maximize the objective $\sum{t=1}^T r_t x_t$. The regret measures the gap between the offline optimal objective value and the decision maker's cumulative reward, while the constraint violation measures the feasibility of our proposed solution.

\subsection{Main Results and Contributions}
While many existing online LP algorithms rely on a primal–dual paradigm (e.g., \cite{li2023simple}), these methods cannot be directly transferred to the semi-infinite setting. In standard online LP, dual variables are updated via multiplicative weights or projection within a finite-dimensional Euclidean space. In contrast, the dual variable in an online semi-infinite LP is a non-negative measure. Direct application of multiplicative weights or gradient descent is infeasible in this uncountably infinite-dimensional space. Moreover, their regret bounds depend on the number of constraints, which can be infinitely large under our setting.

To overcome this dimensionality mismatch, we employ function approximation. A key innovation of our work is the utilization of \textit{non-negative basis functions} to parameterize the dual space. This approach offers two distinct advantages: 1) it allows us to optimize over non-negative weights, a constraint that is computationally straightforward to maintain during updates; and 2) the objective value of the approximate problem serves as an upper bound to the original problem, simplifying the derivation of performance guarantees.

%Many online LP algorithms adopt a primal–dual paradigm (e.g. \cite{li2020simple}), however, such an idea cannot be directly applied to online semi-infinite LP. In online LP, the dual variables are updated via multiplicative-weights schemes or projection in a finite-dimensional Euclidean space; while in online semi-infinite LP, the dual variable is a nonnegative measure, which means direct multiplicative-weights or gradient descent updates are not implementable in such an uncountably infinite-dimensional space. Therefore, we consider function approximations to handle this problem. A key innovation is that we utilize \textit{non-negative basis functions}, which has several benefits: i) we can simply focus on non-negative weights, which is easy to satisfy during the update; 2) the objective value of the approximate problem is equivalent or larger than the original problem, which enables us to only focus on the approximate problem when deriving the upper bounds of our algorithm.

Leveraging this novel formulation, we provide the following main contributions.

1. \textbf{A Simple Algorithm Design and Sublinear Regret Bounds}: We first design a simple Gradient Descent algorithm for the stochastic input model and then extend this to a general Mirror Descent framework applicable to both stochastic and random permutation models.
We prove that our algorithm achieves an $O(q\sqrt{T})$ sublinear regret bound under the stochastic input model and an $O((q+q\log{T})\sqrt{T})$ regret bound under the random permutation model, where $q$ denotes the dimension of the basis functions. %Note that both our regret bounds are independent of the number of constraints, which demonstrate the potential of our algorithm for handling the semi-infinite problems.

2. \textbf{An Improved Two-Stage Algorithm}: Under a Global Polyhedral Growth (GPG) assumption with parameter $\epsilon$, we propose an enhanced two-stage algorithm. This method first accelerates the convergence of the dual variable to a neighborhood of the optimal solution, and subsequently refines it. This approach achieves an improved regret bound of $O(q\log{T}+q/\epsilon)$, significantly outperforming the standard $O(q\sqrt{T})$ rate.
 
3. \textbf{Scalability via Constraint-Cardinality-Independence}: A central bottleneck in extending standard online LP methods to semi-infinite problems is the explicit dependence on the constraint cardinality $m$. In particular, many classical online LP algorithms incur linear or polynomial dependence on $m$, which hinders their scalability. Our framework removes this dependence by applying function approximation to the dual variable: instead of maintaining a per-constraint dual variable, we approximate the non-negative dual measure using $q$-dimensional non-negative basis functions and update only their weights. Consequently, our regret bounds only depend on $q$, which is a known constant and independent of $m$. This demonstrates the potential of our algorithm for handling the semi-infinite problems. We compare our contributions with important prior works under similar formulation in the following \Cref{tab:comparison}.

\begin{table}[htbp]
\centering
\caption{Comparison of our work to important prior literature}
\label{tab:comparison}
\renewcommand{\arraystretch}{1.3}
\begin{tabular}{ccc}
\toprule
& Model & Regret \\
\midrule

\midrule
\citet{li2022online} & stochastic input & $O(m^2\log{T}\log{\log{T}})$ \\
\midrule
\multirow{2}{*}{\citet{li2023simple}}  & stochastic input &$O(m\sqrt{T})$ \\
                                       & random permutation & $O((m+\log{T})\sqrt{T})$ \\   
\midrule
\citet{gao2025beyond} & stochastic input & $O(m\log{T})$ \\
\midrule
\multirow{2}{*}{Our work}  & stochastic input & $O(q\sqrt{T})$, $O(q\log{T}+\frac{q}{\epsilon})$\\
                           & random permutation & $O((q+q\log{T})\sqrt{T})$ \\
\bottomrule
\end{tabular}
\end{table}

In addition to the above-mentioned main contributions, we extend our two online LP algorithms to a general function setting, where the objective and constraints are governed by concave functions $f$ and convex function $g$ rather than a linear parametrization, and analyze their theoretical guarantees. Such an extension substantially broadens our algorithms' applicability to nonlinear constraints or objectives and rich policy classes. 
Finally we conduct numerical experiments to demonstrate the superior performance of our algorithms, particularly in scenarios characterized by a massive number of constraints where traditional methods struggle.

In summary, our work provides the first algorithmic framework for solving the online semi-infinite LP problems, and we justify our algorithms both theoretically via rigorous performance guarantees and empirically through extensive numerical experiments. 

\subsection{Related Literature}
\paragraph{Semi-Infinite Linear Programming}
Semi-Infinite Linear Programming (SILP) addresses linear optimization problems with finite decision variables but infinite inequality constraints. Classical SILP algorithms typically follow a finite-approximation–correction paradigm: solve a master problem over a finite subset of constraints, then identify a new violating index and augment the active set. Several previous literature develops new algorithms based on the central cutting plane method \citep{betro2004accelerated, mehrotra2014cutting, zhang2013entropy}, where any sufficiently informative violated or nearly tight constraint can generate an effective cut. Recently, many literature focuses on the adaptive discretization \citep{still2001discretization, mitsos2011global, jungen2022adaptive, reemtsen2025semi}. In addition, \citet{oustry2025convex} introduces a novel method that employing an inexact oracle to handle convex semi-infinite programmings.

\paragraph{Online Linear Programming/Resource Allocation}
Online Linear Programming/resource allocation has long been investigated and has various applications, such as displaying ad allocation \citep{mehta2007adwords}, revenue management \citep{talluri2006theory,jasin2012re, jiang2025degeneracy}, covering and packing problems \citep{buchbinder2005online,buchbinder2006improved,feldman2010online} and more. 
A classical and prevalent framework for designing algorithm is the dual-based method, where the decision is made based on an estimated dual prices. Some literature follow the sample-then-price scheme, that is the algorithm samples an initial prefix to learn dual prices and then fixes them for the rest of the horizon \citep{devanur2009adwords,feldman2010online,molinaro2014geometry,devanur2012online}. In contrast, other literature adopt the dynamic price learning scheme, which re-solves and updates prices periodically, improving robustness and tightening dependence on horizon \citep{agrawal2014dynamic,li2022online,chen2015dynamic}.  In addition, there are also some papers focusing on primal-guided methods \citep{kesselheim2014primal} and first-order method without repeating solving dual price \citep{agrawal2014fast,balseiro2020dual}.

\paragraph{Online Convex Optimization and Constrained Variants.}
Our work is also related to online convex optimization (OCO). In the standard OCO framework, the decision maker chooses the decision before observing the loss function, and then incurs loss. This model is mainly motivated by machine learning applications such as online linear regression and online support vector machines \citep{hazan2016introduction}. Compared with our setting, standard OCO typically adopts a weaker static regret benchmark, where the comparator is fixed over time. By contrast, our benchmark is the offline optimum of the realized instance, which makes standard OCO algorithms and analyses not directly applicable.
Several works study dynamic regret in non-stationary or adversarial environments \citep{besbes2015non,hall2013dynamical,jadbabaie2015online}, but they focus on unconstrained problems. There is also a line of work on online convex optimization with constraints (OCOwC), where the constraint functions are either static \citep{jenatton2016adaptive,yuan2018online,yi2021regret} or i.i.d. over time \citep{neely2017online}. The extension to Markovian environment has also been studied \citep{jiang2023constant, li2025revenue, jiang2024achieving} under various metrics with various applications. To the best of our knowledge, existing OCOwC results do not allow for a large or infinite number of constraints, which is the problem we address in this paper.

\section{Problem Formulation}
We consider the general online semi-infinite LP, which takes the following form:
\begin{equation} \label{lp:primal}
\begin{aligned}
    \max\ &\bm{r}^{\top}\bm{x} \\ 
    \mathrm{s.t.\ } &A\bm{x} \le \bm{b} \\
    & 0 \le x_i \le 1,~~i=1,...,T
\end{aligned}
\end{equation}
where the input data $(r_t, \bm{a}_t)$ arrives at each time step $t$ and the right-hand-side $\bm{b}$ and time horizon $T$ are known beforehand. Here the dimension of columns of matrix $A$ and vector $\bm{b}$ are infinite and we use $m$ to denote it for convenience. The dual of this semi-infinite LP \eqref{lp:primal} is:
\begin{equation} \label{lp:dual}
\begin{aligned}
    \min\ &\bm{b}^{\top}\bm{u} + \bm{1}^{\top}\bm{s} \\ 
    \mathrm{s.t.\ } &A^{\top}\bm{u} + \bm{s}\ge \bm{r} \\
    & \bm{u} \ge 0, ~\bm{s} \ge 0
\end{aligned}
\end{equation}

Following traditional online LP literature, our paper mainly discusses two types of models: stochastic input model and random permutation model, and we formally state the definitions for these two models in the following:
\begin{definition} \label{def:models}
    The input data $\{(r_t,\bm{a}_t)\}_{t=1}^T$ satisfy one of the following standard models:

    1. (\textbf{Stochastic Input Model}) $(r_t,\bm{a}_t)$ are i.i.d. sampled from an unknown distribution $\mathcal{P}$.

    2. (\textbf{Random Permutation Model}) The multiset $\{(r_t,\bm{a}_t)\}_{t=1}^T$ is fixed in advance and arrives in a uniformly random order.
\end{definition}

\subsection{Assumptions}
Now we introduce some basic assumptions throughout the paper. We assume that both input data $\{r_t, \bm{a}_t\}_{t=1}^T$ and right-hand-side $\bm{b} = T\bm{d}$ are bounded. Furthermore, our paper develops a new LP formulation via function approximations with non-negative basis functions $\Phi\in\mathbb{R}_{\ge0}^{m\times q}$, which will be specified later. In semi-infinite LP, it is common to assume that the dual measure is finite, which can be translated that the term $\bm{a}_t^\top \Phi$ and $\bm{d}^\top \Phi$ are also bounded. We formally conclude these assumptions in the following:
\begin{assumption} \label{asp:bounded_data}
For all $t=1,...,T$,

1. (\textbf{Bounded Data}) Let right-hand-size $\bm{b} = T\bm{d}$ and there exist known constants $\overline{r} > 0$, $\overline{a} > 0$ and $\overline{d} >\underline{d} >0$ such that, 
\[
r_t \le \overline{r},~~ \left\|\bm{a}_t\right\|_\infty \le \overline{a},~~\underline{d} \le \left\|\bm{d}\right\|_\infty \le \overline{d}.
\]

2. (\textbf{Finite Dual Measure}) With respect to the basis function $\Phi$, there exist positive constants $0 < \underline{D} \le\overline{D}$ and $\overline{C} > 0$ such that, 
\[
\underline{D} \le \left\|\bm{d}^\top \Phi\right\|_\infty \le \overline{D},~~\left\|\bm{a}_t^\top \Phi\right\|_\infty \le \overline{C}.
\]
\end{assumption}

\subsection{Performance Measure}
For a given sequence of input data $S = \{(r_t, \bm{a}_t)\}_{t=1}^T$, we denote the offline optimal solution of the semi-infinite LP \eqref{lp:primal} as $\bm{x}^* = (x_1^*,...x_T^*)$ and the corresponding optimal value as $R_T^*$. Similarly, we denote the actual decision of each time step $t$ as $x_t(\pi)$ for a fixed policy $\pi$ and the objective value as $R_T(\pi)$. That is, 
\[
R^*_T = \sum_{t=1}^T r_tx_t^*,~~R_T(\pi) = \sum_{t=1}^T r_tx_t(\pi).
\]
Then we can give the formal definition of the optimality gap:
\[
\text{Reg}_T(\mathcal{P},\pi) = R_T^* - R_T(\pi)
\]
where $\mathcal{P} = (\mathcal{P}_1,...,\mathcal{P}_T)$ are the unknown distribution of input data $S = \{(r_t, \bm{a}_t)\}_{t=1}^T$. For the stochastic input model, these distribution $\mathcal{P} = (\mathcal{P}_1,...,\mathcal{P}_T)$ are all identical; while for the random permutation model, each $\mathcal{P}_t$ is the conditional empirical distribution of the yet-unseen items (i.e. sampling without replacement).
Following traditional online LP literature, we aim to find an upper bound for the expectation of the optimality gap for all distributions $\mathcal{P}$, and we define this performance measure as \textit{regret}:
\[
\text{Reg}_T(\pi) = \sup\limits_{\mathcal{P}\in\Xi} \mathbb{E}[\text{Reg}_T(\mathcal{P},\pi)]
\]
Furthermore, we consider a supplementary performance measure: \textit{constraints violation}. In classical Online LP with a finite number of constraints, violation is typically measured by the Euclidean norm of the residual vector. However, in the semi-infinite setting, the constraint index set is continuous, rendering the standard Euclidean norm ill-defined or computationally intractable. To address this, we adopt a dual-based measurement approach. Instead of checking pointwise violation, we measure the magnitude of violation "tested" against our dual basis functions. This concept draws from the weak duality in functional analysis, where a constraint is considered satisfied if its integrated value against all valid test functions is non-positive.

Let $\mathcal{U} = \{\left\|\bm{u}\right\|_2 \le \overline{u}: \bm{u} = \Phi \bm{w} \ge 0\}$ denote the domain of valid dual test functions within the span of our non-negative basis $\Phi$. We formally define the \textit{constraint violation} as the worst-case weighted violation over this set:
\[
v(\pi) = \max_{u\in\mathcal{U}}\mathbb{E}\left[||\left[\bm{u}^\top(A\bm{x}(\pi) - \bm{b})\right]^+||_2\right].
\]

\section{The new LP formulation via Function Approximation}
\label{sec:new_lp}

%\subsection{Previous derivation}
%In real-world online LP applications, it is common to encounter too many constraints. On the one hand, classical online LP algorithms do not perform well since their regret bound is linearly dependent on the number of constraints; on the other hand, traditional semi infinite LP algorithms fails because they need to know the distributions of the constraints $A$ and coefficients $\bm{r}$, which are unknown in online settings. Therefore, an efficient algorithm to solve online LP with large or even infinite number of constraints is of vital importance. In this section, we introduce our new LP formulation via function approximation.
Large-scale problems remain challenging due to their complexity and the high computational cost of solving them directly, where function approximation has been recently investigated as a powerful tool to handle these issues (e.g., \cite{liu2022understanding,jiang2025adaptive}).
In this section, we introduce our new LP formulation via function approximation for the challenging OSILP problems.
Denote the optimal solutions to the primal LP \eqref{lp:primal} and dual LP \eqref{lp:dual} as $\bm{x}^*, \bm{u}^*$ and $\bm{s}^*$. According to the complementary slackness, we have the following equations:
\begin{equation} \label{eq:comple}
x^*_t = 
\begin{cases}
1, & r_t > \bm{a_t}^{\top}\bm{u}^* \\
0, & r_t < \bm{a_t}^{\top}\bm{u}^*
\end{cases} ~~ \text{and} ~~s_t^* = [r_t - \bm{a_t}^{\top}\bm{u}^*]^+
\end{equation}
for $t = 1,...,T$, and $x_t$ can be an arbitrary value between $0$ and $1$ when $r_t = \bm{a_t}^{\top}\bm{u}^*$. 
Therefore, it is suffice to make a decision $x_t$ if the decision maker can obtain the accurate value of $\bm{u}^*$. Note that the dimension of dual variable $\bm{u}^*$ is also infinite, so we use function approximation to approximate it. To be specific, we assume that there exists a variable $z$ such that the dual variable $\bm{u}$ can be well approximated in the following formulation:
\begin{equation} \label{approx_p}
\bm{u}(z) \approx \sum_{k = 1}^q \phi(z)_kw_k = \Phi \bm{w}
\end{equation}

A key step is that we use \textit{non-negative basis functions} (e.g. Radial Basis Functions) to construct the basis function $\Phi$.
This technique brings several benefits. First, for an arbitrary basis function $\Phi$, which may have both negative and positive elements, it is non-trivial to project the weight $w$ into the feasible set $\Phi \bm{w} \ge 0$ during the update process. However, if we choose \textit{non-negative basis functions}, $\bm{w} \ge 0$ is a sufficient condition to satisfy the original constraint $\Phi \bm{w} \ge 0$, and projecting $\bm{w}$ into non-negative sets is trivial. 
Second, although these two constraints are not equivalent and there exists a gap between optimal values of these two problems, we can show that such a gap does not affect our final regret upper bound. 
More specifically, let $\bm{k}:= \Phi \bm{w}$. Since basis function $\Phi$ may not constitute an orthogonal basis for the space $\mathbb{R}_{\ge0}^m$, the domain for $\bm{k}$ is $\mathcal{K} \subseteq \mathbb{R}_{\ge0}^m$; while the domain for the original dual variable $\bm{u}$ is exactly $R_{\ge0}^m$. In other words, the domain of our approximate variable via basis function $\Phi$ is a subset of the domain of our target.

Let $\bm{d} = \frac{\bm{b}}{T}$ and combine function approximations \eqref{approx_p} and complementary slackness \eqref{eq:comple}, we can construct the following problem, which is actually an approximate to the original dual LP \eqref{lp:dual}:
\begin{equation} \label{lp:approx_dual}
\begin{aligned}
    \min_{\bm{w}}\ &f_{T,\Phi}(\bm{w}) = \bm{d}^{\top}\Phi \bm{w} + \frac{1}{T}\sum^T_{t=1}(r_t - \bm{a}_t^\top\Phi \bm{w})^+ \\
    \mathrm{s.t.\ } &\bm{w} \ge 0
\end{aligned}
\end{equation}
Assume that the column-coefficient pair $(r_t, \bm{a_t})$ are sampled from some unknown distribution $\mathcal{P}$ and we can consider fluid relaxation: 
\begin{equation} \label{lp:approx_stochastic}
\begin{aligned}
    \min_{\bm{w}}\ &f_\Phi(\bm{w}) = \bm{d}^{\top}\Phi \bm{w} + \mathbb{E}_{(r,a) \sim \mathcal{P}}\left[(r - \bm{a}^\top\Phi \bm{w})^+\right] \\
    \mathrm{s.t.\ } &\bm{w} \ge 0
\end{aligned}
\end{equation}
From another perspective, problem \eqref{lp:approx_dual} can also be viewed as the sample average approximation of the stochastic problem \eqref{lp:approx_stochastic}.
Applying similar idea to the dual problem \eqref{lp:dual}, we can obtain the following stochastic problem:
\begin{equation} \label{lp:dual_fluid}
\begin{aligned}
    \min_{\bm{u}}\ &f(\bm{u}) = \bm{d}^{\top}\bm{u} + \mathbb{E}_{(r,a) \sim \mathcal{P}}\left[(r - \bm{a}^\top\bm{u})^+\right] \\
    \mathrm{s.t.\ } &\bm{u} \ge 0
\end{aligned}
\end{equation}
Denote $\bm{w}^*$ as the optimal solution for the approximate dual problem \eqref{lp:approx_dual} and $\bm{u}^*$ is the optimal solution for the fluid relaxation dual LP \eqref{lp:dual_fluid}, and the following lemma help us build a bridge between $f(\bm{u}^*)$ and $f_{T,\Phi}(\bm{w}^*)$:
\begin{lemma} [\textbf{forklore}] \label{lemma:w_p}
It holds that $f_{T,\Phi}(\bm{w}^*) \ge f(\bm{u}^*).$
\end{lemma}

Note that our goal is to find an upper bound for regret and constraints violation, \Cref{lemma:w_p}
naturally bridges the original optimal value $f(\bm{u}^*)$ to the approximate optimal value $f_{T,\Phi}(\bm{w}^*)$, and from now on we can simply focus on dealing with problem \eqref{lp:approx_stochastic} and $f_{T,\Phi}(\bm{w}^*)$.

\section{A Dual-based First-order Algorithm} \label{sec:simple}
We begin with a data-driven first-order algorithm under the stochastic input model and later we extend it to a more general version under both the stochastic input model and the random permutation model. The formal definition of stochastic input model is presented in \Cref{def:models} and our main algorithm is presented in \Cref{alg:main}. 

Our algorithm is motivated from the dual-based algorithm, that is, in each time step $t$, the algorithm makes a decision $x_t$ according to the optimal dual variable $\bm{p}^*$, which is approximated by $\Phi \bm{w}_t$ in our algorithm. Then we update $\bm{w}_t$ to adjust our approximation to $\bm{p}^*$ using the sub-gradient descent method, and project it into the non-negative set. The update \eqref{main:update_w} is the sub-gradient of $t$-th term evaluated at $\bm{w}_t$:
\[
\begin{aligned}
&\left.\partial_{\bm{w}}\left(\bm{d}^\top\Phi \bm{w} + (r_t - \bm{a}_t^\top \Phi \bm{w})^+\right)\right|_{\bm{w}=\bm{w}_t} \\
&= \left.\bm{d}^\top\Phi - \bm{a}_t^\top\Phi I(r_t > \bm{a}_t \Phi \bm{w})\right|_{\bm{w}=\bm{w}_t} \\
& = \bm{d}^\top\Phi - \bm{a}_t^\top\Phi x_t
\end{aligned}
\]
where the last equation uses the definition of $x_t$ in Step $4$ of \Cref{alg:main}.

\begin{algorithm}[ht!]
%\small
\caption{Main algorithm}
\label{alg:main}
%\SetAlgoLined
\begin{algorithmic}[1]
\State \textbf{Input:} $\bm{d} = \frac{\bm{b}}{T}$.
%\State For each state-action pair $(s,a)$, we sample the model $\mathcal{M}$ for $N_0$ times.
\State Initialize $\bm{w}_1 = 0$.

\For{$t=1,...T$}
\State Observe input data $(r_t, \bm{a}_t)$ and set \[x_t = 
            \begin{cases}
            1, ~~ r_t > \bm{a_t}^\top \Phi \bm{w_t} \\
            0, ~~ r_t \le \bm{a_t}^\top \Phi \bm{w_t}
            \end{cases}\]

\State Update $ \label{main:update_w}
\bm{w}_{t+1} = \prod_{\bm{w}\ge0}\left[\bm{w}_t +\gamma_t(\bm{a_t}^\top\Phi x_t - \bm{d}^\top\Phi)\right]$
\EndFor
\State \textbf{Output}: $\bm{x}=(x_1,...x_T)$.
\end{algorithmic}
\end{algorithm}

Now we provide another perspective to illustrate \Cref{alg:main}. Consider the following LP:
\begin{equation} \label{lp:another}
\begin{aligned}
    \max\ &\bm{r}^{\top}\bm{x} \\ 
    \mathrm{s.t.\ } &\Phi^\top A\bm{x}\le \Phi^\top \bm{b} \\
    & 0 \le x_t \le 1,~~t=1,...,T
\end{aligned}
\end{equation}

Compared with the original primal formulation \eqref{lp:primal}, the constraint system in \eqref{lp:another} is obtained by projecting the original constraints onto the subspace spanned by $\Phi$. Hence, \eqref{lp:another} can be viewed as a constraint-reduced (or projected) surrogate of \eqref{lp:primal}: it achieves the approximate objective but replaces the potentially large set of constraints by a lower-dimensional set encoded through $\Phi^\top A$ and $\Phi^\top \bm{b}$.

The dual of \eqref{lp:another} coincides with the approximate dual problem \eqref{lp:approx_dual} by identifying $\bm{d}=\frac{\bm{b}}{T}$, where the associated dual variable is $\bm{w}$. Under this interpretation, \Cref{alg:main} can be viewed as a standard primal--dual method for solving the projected LP \eqref{lp:another}: the primal update seeks to improve the reward under the reduced constraints, while the dual update adjusts $\bm{w}$ to penalize violations of $\Phi^\top A\bm{x}\le \Phi^\top \bm{b}$. This perspective will be useful for relating the algorithmic iterates to both the approximate dual objective and the extent of constraint satisfaction in the original space.

Before presenting our formal result, we first illustrate that the dual variable $\bm{w}_t$ is upper bounded during the update, which is presented in \Cref{lemma:w_upper_bound}.
\Cref{lemma:w_upper_bound} is highly dependent on \Cref{asp:bounded_data}, and its main purpose is to help us derive an upper bound for constraints violation.

\begin{lemma} \label{lemma:w_upper_bound}
Under the stochastic input model, if the step size $\gamma_t \le 1$ for $t=1,...,n$ in \Cref{alg:main}, then
\begin{equation}
\begin{aligned}
    ||w_t||_2 & \le \frac{q(\overline{C}+\overline{D})^2 +2r_t}{2\underline{D}} +q(\overline{C}+\overline{D})
\end{aligned}
\end{equation}
where $w_t$'s are specified in \Cref{alg:main}.
\end{lemma}

Now we present our main result in \Cref{thm:simple} and leave the full proof to appendix. Such a $O(\sqrt{T})$ regret bound has been achieved in previous online LP literature, however, most of their results are also linear dependent on the constraint dimension $m$, which leads to poor performance confronted with large or even infinite constraints. Our algorithm is the first to achieve a $O(q\sqrt{T})$ regret bound under the online SILP setting and our bound is only dependent on the dimension of weight $w$, which is controllable and small in real applications.

\begin{theorem} \label{thm:simple}
    Under the stochastic input assumption and \Cref{asp:bounded_data}, if step size $\gamma_t = \frac{1}{\sqrt{T}}$ for $t=1,..,T$ in \Cref{alg:main}, the regret and expected constraints violation of \Cref{alg:main} satisfy
    \begin{equation}
        \text{Reg}_T(\pi_{\text{ALG1}}) \le O(q\sqrt{T})~~\text{and}~~v(\pi_{\text{ALG1}}) \le O(q^2\sqrt{T}). \nonumber
    \end{equation}
\end{theorem}

\section{Mirror-Descent-Based Method} \label{sec:mirror}
\subsection{Stochastic Input Model}
In this subsection, we extend the \Cref{alg:main} to a mirror-descent-based algorithm and discuss its regret bound under both stochastic input model and random permutation model. For a given potential function $\psi$, we denote by $D_{\psi}(u||v)$ the corresponding Bregman divergence, defined as
\[
D_{\psi}(u||v) = \psi(u) - \psi(v) - \langle \nabla\psi(v), u-v\rangle.
\]
Note that when choosing $\psi(u) = \frac{1}{2}\left\|u\right\|_2^2$, the Mirror Descent will degenerate to the standard gradient descent. However, there are several scenarios that choosing other potential functions are better. For instance, negative entropy function $\psi(u) = -\sum_{i=1}^n u_i\log{u_i}$ has better performance for the sparse solution LP problem. Therefore, developing a Mirror Descent Based algorithm is of vital importance. 

Before presenting our formal algorithm, we first introduce a standard assumption for the potential function $\psi$, together with a boundedness assumption on the dual iterates $\bm{w}_t$. We assume that $\psi$ is coordinate-wisely separable and $\alpha$-strongly convex with respect to a given norm, a setting that covers many commonly used potential functions.
When mirror descent reduces to Euclidean projection, (i.g. $\psi(u)=\frac{1}{2}u^2$) we have already established that the dual variables $\bm{w}_t$ admit an upper bound throughout the iterations. However, for a general potential function, it is typically not possible to derive an explicit or closed-form upper bound for the dual variables in the same way as in the Euclidean case. In light of this difficulty, we introduce a more general assumption that the dual variables remain bounded over the course of the algorithm’s updates once the potential function is determined.
\begin{assumption} [Mirror map and bounded dual iterates] \label{assump:sepa}
    The potential function $\psi(u)$  is coordinate-wisely separable, that is, $\psi(u) = \sum_{j=1}^p\psi_j(u_j)$ where $\psi_j:\mathbb{R}_+ \rightarrow\mathbb{R}$ is an univariable function and $\alpha$-strongly convex with respect to a given norm $\left\|\cdot\right\|$. Moreover, the dual variable generated by \Cref{alg:mirror_sto} remains uniformly bounded throughout the iterations: there exists a known constant $\overline{W}$ such that $\left\|\bm{w}_t\right\|_{\infty} \le \overline{W}$.
\end{assumption}

We present our formal algorithm in \Cref{alg:mirror_sto}, which is mainly based on \Cref{alg:main} with the modification on the update formula for dual variable $\bm{w}_t$. Now we discuss the regret bound of \Cref{alg:mirror_sto} under stochastic input model first, which is given in \Cref{thm:mirror_bound}. Our bound is dependent on the concrete potential function $\psi$ with respect to a norm $||\cdot||$. To bound these arbitrary norm, we give the upper bound under the $l_2$ norm. Therefore for any given norm $||\cdot||$, $||\bm{g}_t||_*^2$ will always have an upper bound, which is a constant multiple depending on that particular norm and the dimension. In our final bound, we omit this term and obtain the $O(q\sqrt{T})$ regret bound.
\begin{algorithm}[ht!]
%\small
\caption{Mirror-descent-based algorithm}
\label{alg:mirror_sto}
%\SetAlgoLined
\begin{algorithmic}[1]
\State \textbf{Input:} $\bm{d} = \frac{\bm{b}}{T}$.
%\State For each state-action pair $(s,a)$, we sample the model $\mathcal{M}$ for $N_0$ times.
\State Initialize $\bm{w}_1 = 0$.
\State Parameters: potential function $\psi$; step size $\gamma_t$
\For{$t=1,...,T$}
\State Observe input data $(r_t, \bm{a}_t)$ and set \[x_t = 
            \begin{cases}
            1, ~~ r_t > \bm{a}_t^\top \Phi \bm{w}_t \\
            0, ~~ r_t \le \bm{a}_t^\top \Phi \bm{w}_t
            \end{cases}\]

\State Update $\bm{g}_t = \bm{d}^\top\Phi - \bm{a}_t^\top\Phi x_t$.
\State Update $\bm{w}_{t+1} = \arg \min\limits_{\bm{w}\ge0} \{ \langle \gamma_t \bm{g}_t, \bm{w} \rangle + D_{\psi}(\bm{w}||\bm{w}_t) \}$.
\EndFor
\State \textbf{Output}: $\bm{x}=(x_1,...x_T)$.
\end{algorithmic}
\end{algorithm}

\begin{theorem} \label{thm:mirror_bound}
    Under the stochastic input model, \Cref{asp:bounded_data}, and \Cref{assump:sepa}, if step size $\gamma_t = \frac{1}{\sqrt{T}}$ for $t=1,..,T$ in \Cref{alg:mirror_sto}, the regret and expected constraints violation of \Cref{alg:mirror_sto} satisfy
    \begin{equation}
        \text{Reg}_T(\pi_{\text{ALG2}}) \le  O(q\sqrt{T})~~\text{and}~~v(\pi_{\text{ALG2}}) \le O(\sqrt{qT}). \nonumber
    \end{equation}
\end{theorem}

\subsection{Random Permutation Model}
In this subsection, we discuss a more general setting: Random Permutation Model. In this model, coefficient pair $(r_t,\bm{a}_t)$ arrives in a random order and their value can be chosen adversarially at the start. 
To simplify the exposition and exclude degeneracy, we work under a standard general-position convention. In particular, following the common perturbation argument \citep{devanur2009adwords}, one may independently perturb each reward $r_i$ by an arbitrarily small continuous noise term (e.g. $\epsilon\sim\text{U}[0,\beta]$). Since the dual variable $\bm{w}_t$ lies in $\mathbb{R}^q_{\ge0}$, no vector $\bm{w}_t$ can satisfy more than $q$ independent equalities simultaneously; consequently, after this perturbation, with probability one that any $\bm{w}$ can only satisfy at most $q$ independent equations at the same time. Hence, this non-degeneracy condition can be imposed without loss of generality. Under this convention, we can define the following thresholding rule:
\begin{equation} \label{thsld:per}
x_i(\bm{w}^*) = 
\begin{cases}
1, ~~r_i > \bm{a}_i^\top\Phi \bm{w}^* \\
0, ~~r_i \le \bm{a}_i^\top\Phi \bm{w}^*
\end{cases}
\end{equation}
where $\bm{x}(\bm{w}^*) = (x_1(\bm{w}^*),...,x_n(\bm{w}^*))$.
\begin{lemma} \label{lemma:optimal_value}
Denote $x^*_i$ as the optimal solution for the approximate primal LP relaxation \eqref{lp:another} and $x_i(\bm{w}^*)$ is defined in \eqref{thsld:per}, then we have $x_i(\bm{w}^*)\le x^*_i$ for all $i=1,...,n$. Besides, under the general-position convention above, $x_i(\bm{w}^*)$ and $x_i^*$ differs for no more than $q$ values of $i$.
\end{lemma}

Denote the optimal value of approximate primal LP \eqref{lp:another} $R_\Phi^*$. \Cref{lemma:optimal_value} tells us that the difference of $\sum_{i=1}^n r_ix_i(\bm{w}^*)$ and $R_\Phi^*$ can be bounded by $q\overline{r}$. That is,
\[
|\sum_{i=1}^n r_ix_i(\bm{w}^*) - R_\Phi^*| \le q\overline{r}
\]

Then we consider a scaled version of the approximate primal LP relaxation \eqref{lp:another}:
\begin{equation} \label{lp:approx_scaled}
\begin{aligned}
    \max\ \sum_{i=1}^s~r_i^{\top}x_i
    ~~\mathrm{s.t.\ } \sum_{i=1}^sa_{ji}^{\top}x_i \le \frac{sb_j}{T},
     0 \le \bm{x} \le 1
\end{aligned}
\end{equation}
for $s=1,..,T$. Denote the optimal value of scaled version of the approximate primal LP \eqref{lp:another} $R_s^*$. The following proposition helps us relate $R^*_s$ to $R^*_\Phi$:
\begin{proposition} \label{prop:permut}
    For $s>\max\{16\overline{C}^2,e^{16\overline{C}^2},e\}$,the following inequality holds:
    \begin{equation}
        \frac{1}{s}\mathbb{E}[R^*_s] \ge \frac{1}{T}R^*_\Phi - \frac{q\overline{r}}{T} - \frac{\overline{r}\log{s}}{\underline{D}\sqrt{s}} - \frac{q\overline{r}}{s}
    \end{equation}
    for all $s\le T \in \mathbb{N}^+$.
\end{proposition}

Now we present our formal result in \Cref{thm:permu}. Since Random Permutation Model is a more challenging than under the Stochastic Input Model, our regret bound is weaker than prior results; however, these two settings exhibit comparable performance in terms of constraint violations.
\begin{theorem} \label{thm:permu}
Under the random permutation model, \Cref{asp:bounded_data}, and \Cref{assump:sepa}, if step size $\gamma_t = \frac{1}{\sqrt{T}}$ for $t=1,..,n$ in \Cref{alg:mirror_sto}, then the regret and expected constraints violation satisfy
    \begin{equation} \nonumber
    \begin{aligned}
        \text{Reg}_T(\pi_{\text{ALG2}}) \le O\left((q+q\log{T})\sqrt{qT}\right) ~~\text{and}~~
        v(\pi_{\text{ALG2}}) \le O(\sqrt{qT})
    \end{aligned}
    \end{equation}
    where $||\cdot||_*$ denotes the dual norm of $||\cdot||$ and $g_t$ satisfies $||\bm{g}_t||_2^2 \le q(\overline{C}+\overline{D})^2$.
\end{theorem}

\section{Achieving $\log{T}$ Regret} \label{sec:log_regret}
In previous sections, we have introduced two first order algorithms achieving $O(q\sqrt{T})$ regret bound. However, can we develop a more efficient algorithm outperforming $O(q\sqrt{T})$ regret under the stochastic input model? 

The answer is \textit{yes}. In this section, we will propose a novel algorithm that achieves $O(q\log{T}+\frac{q}{\eps})$ regret, exceeding previous $O(m\sqrt{T})$ regret bound in most literature. 
This novel algorithm has two stages: i) accelerating stage, rapidly driving the iterates into a neighborhood of the optimal solution; ii) refinement stage, applying first order methods to refine the solution and ensure accurate optimality.
Before formally presenting the algorithm, we first introduce two common and important assumptions, which play key roles in our later analysis:
\begin{assumption} [General Position Gap] \label{asp:GPG}
Denote $\bm{w}^*$ as the unique optimal solution of  \eqref{lp:approx_stochastic}. Let $\mathcal{B}_\epsilon(d)$ be the ball centered at $d$ with radius $\epsilon$
under the $\ell_2$ norm, i.e.,
$$\mathcal{B}_\epsilon(d) = \{\hat{d} : \|\hat{d} - d\|_2 \le \epsilon\}.$$
There exists $\epsilon > 0$ such that for all $\hat{d} \in \mathcal{B}_\epsilon(d)$, $\bm{w}^*$ remains the optimal solution.
\end{assumption}

Note that this General Position Gap (GPG) assumption is equivalent to the general non-degeneracy assumption. The general non-degeneracy assumption mainly includes following two parts: i) For a given RHS vector $d$ the primal LP admits a unique optimal basic solution; ii) This optimal basis is nondegenerate, which means all basic variables, including the associated slack variables, are strictly positive ( i.e., no basic variable is equal to zero). From the geometric perspective, this means that the RHS vector $d$ does not lie on the degeneracy boundary of the feasible region; instead, it lies in the interior of a well-behaved region in which the optimal basis is unique and stable. Therefore, there must exist $\eps > 0$ such that as long as $d$ falls into a small region $\mathcal{B}_\epsilon(d) = \{\hat{d} : \|\hat{d} - d\|_2 \le \epsilon\}$, this unique optimal basic solution remains optimal, which is exactly equivalent to our GPG definition.

\begin{assumption} [H\"{o}lder Error Bound] \label{asp:holder}
For all $\bm{w} \ge 0$, there exist some $\lambda > 0$ and $\theta \in (0,1]$ such that
$$f(\bm{w}) - f(\bm{w}^*) \ge \lambda \cdot dist(\bm{w},\mathcal{W}^*)^{1/\theta}$$
where $dist(\bm{w},\mathcal{W}^*) = \inf \left\|\bm{w}-\bm{w}^*\right\|$ and $\mathcal{W}^*$ is the optimal set for problem \eqref{lp:another}.
\end{assumption}

This H\"{o}lder Error Bound assumption characterizes an upper bound when the point is away from the optimal set. In polyhedral settings such as linear programming, this property typically follows from Hoffman-type error bounds and thus holds under mild regularity assumptions. The dual error bound is a key regularity condition in convergence analysis: it allows one to convert the decay of a residual or objective gap into a corresponding decrease in the distance to the dual solution set, which in turn facilitates linear convergence guarantees for first order algorithms.

\subsection{Accelerate Stage}

We now introduce the first part of our algorithm: the Accelerate stage. In traditional first-order methods, the initial point is often set to $\bm{w}_1 = 0$ and is frequently far from the optimal solution $\bm{w}^*$, which typically leads to slow convergence. To address this issue, we employ an accelerated gradient descent method to guide the initial point to a neighborhood of the optimal solution  in as few steps as possible. The formal algorithm is presented in \Cref{alg:fast}.

\begin{algorithm}[ht!]
%\small
\caption{Fast Convergence Algorithm - Accelerate Stochastic Subgradient Method}
\label{alg:fast}
%\SetAlgoLined
\begin{algorithmic}[1]
\State \textbf{Input:} $\bm{d} = \frac{\bm{b}}{T}$, total time horizon $T$, probability $\delta$, initial error estimate $\eps_0 = \frac{\overline{r}}{\underline{D}}$.
\State Set $L = \lceil\log{(\eps_0T)}\rceil$, $\hat{\delta} = \log{\frac{\delta}{K}}$, $J \ge \max \{\frac{9(\overline{C}+\overline{D})^2}{\lambda^2}, \frac{1152(\overline{C}+\overline{D})^2\log{(1/\hat{\delta})}}{\lambda^2}\}$ and $T_{\text{fast}} = L\cdot J$
\State Initialize $\bm{w}_0 = \tilde{\bm{w}}_0 = 0$; step size $\eta_1 = \frac{\eps_0}{3\overline{r}^2}$ and $\gamma_t = \frac{1}{\log{T}}$; $V_1 \ge \frac{\eps_0}{\lambda}$; $\bm{B}_0 = \bm{b}^\top\Phi$.
\For{$l=1,...L$}
\State Set $\tilde{\bm{w}}_1^l$ = $\tilde{\bm{w}}_{l-1}$ and domain $\mathcal{K} = \mathbb{R}_{\ge0}\cap\mathcal{B}(\tilde{\bm{w}}_{l-1},V_k)$.
\For{$j=1,...J$} 

\State Compute current time step $t = (l-1)*J + (j-1)$.
\State Observe input data $(r_t, \bm{a}_t)$ and set \[\tilde{x}_t = 
            \begin{cases}
            1, ~~ r_t > \bm{a_t}^\top \Phi \bm{w_t}\\
            0, ~~ r_t \le \bm{a_t}^\top \Phi \bm{w_t}
            \end{cases}\]
\State Set $x_t = \tilde{x}_t \cdot \mathbb{I}\{\bm{a}_t^\top\Phi \tilde{x}_t \le \bm{B}_t\}$
\State Update $\bm{B}_{t+1} = \bm{B}_t - \bm{a}_t^\top\Phi x_t$
\State Update $
\bm{w}_{t+1} = \max\{
\bm{w}_t +\gamma_t(\bm{a_t}^\top\Phi \tilde{x}_t - \bm{d}^\top\Phi), \bm{0}\}$

\State Update $\tilde{\bm{w}}_{j+1}^l = \prod_{\mathcal{K}}\left[\tilde{\bm{w}}_j^l + \eta_l\left(\bm{a}_t^\top\Phi \tilde{x}_t - \bm{d}^\top\Phi\right)\right]$

\EndFor
\State Set $\tilde{\bm{w}}_l = \frac{1}{J}\sum_{j=1}^J \tilde{\bm{w}}_j^l$.
\State Set $\eta_{l+1} = \frac{\eta_l}{2}$ and $V_{l+1} = \frac{V_l}{2}$.
\EndFor
\State \textbf{Output}: $(x_1,...,x_{T_{\text{fast}}}),\tilde{\bm{w}}_L, T_{\text{fast}}$.
\end{algorithmic}
\end{algorithm}

Note that our \Cref{alg:fast} maintains two variables: $\bm{w}_t$ and $\tilde{\bm{w}}^l_j$. The accelerated gradient descent mechanism ensures that $\tilde{\bm{w}}^l_j$ converges rapidly to a near-optimal solution. 
The key point of this step is that we not only project $\bm{w}_t$ onto $\mathbb{R}_{\ge0}$ but also onto a ball whose radius is updated synchronously, which helps to limit any oscillations that $\bm{w}_t$ may experience during the update process. Since $\bm{w}_t$ gets closer to the optimal value with each update, we treat the previous $\bm{w}_t$ as the center of the ball for the next update and simultaneously decrease the radius of the projection ball to ensure that $\bm{w}_t$ converges rapidly to a region around the optimal value while minimizing oscillations and improving stability.
However, this fast convergence has no guarantee towards the regret of decision making and may sometimes result in bad performance. To address this issue, we also maintain the variable $\bm{w}_t$ to more reliably approximate the optimal dual variable while making decisions.

Another distinctive feature of \Cref{alg:fast} compared to previous methods is the handling of constraints in Steps $9$. In this part of the algorithm, the procedure ensures that all constraints are continuously satisfied throughout the process. This is crucial because, unlike in traditional algorithms where constraints may occasionally be violated, \Cref{alg:fast} guarantees that constraints are satisfied at all times, thus improving the robustness and reliability of our decisions.

\begin{theorem} \label{thm:fast}
Under the stochastic input model, \Cref{asp:bounded_data}, and \Cref{asp:holder}, suppose $\bm{w}^*$ is the optimal solution for the problem \eqref{lp:another}, the output of \Cref{alg:fast} satisfies that
\[
\left\|\tilde{\bm{w}}_L - \bm{w}^*\right\| \le \frac{1}{T}
\]
with high probability $1-\delta$. In addition, the regret of \Cref{alg:const_reg} untill time $T_{\text{fast}}$ satisfies that
\[
\text{Reg}_{T_{\text{fast}}}(\pi_{\text{ALG3}}) \le O\left(q\log{T}\right).
\]    
\end{theorem}

\subsection{Refine Stage}

Now we comes to the second stage and present it in \Cref{alg:const_reg}. The key innovation of \Cref{alg:const_reg} compared to previous algorithms lies in its use of an initial dual variable that is a near-optimal solution to the fluid relaxation problem, rather than starting from $0$ as in prior approaches.
In this way, the algorithm does not waste time converging from an arbitrary starting point and maintain more accurate solutions throughout. 
Further, this adaptive design also offers a significant advantage over directly using the expected optimal solution, which is known to be an $O(\sqrt{T})$ regret. By dynamically adjusting based on the observed data at each step, the algorithm avoids rigidly relying on a fixed expectation, allowing it to respond more flexibly to variations in the data. 
This approach ensures that the algorithm remains closely aligned with underlying problem dynamics, preventing large deviations from expected optimum and resulting in a substantial improvement in our constant regret bound.
In addition, \Cref{alg:const_reg} also adopts a virtual decision $\tilde{x}_t$ as in \Cref{alg:fast} to ensure that all constraints are satisfied and maintain the overall consistency of the final algorithm.

\begin{algorithm}[ht!]
%\small
\caption{Refine Algorithm}
\label{alg:const_reg}
%\SetAlgoLined
\begin{algorithmic}[1]
\State \textbf{Input:} $\bm{d} = \frac{\bm{b}}{T}$, near optimal dual solution $\hat{\bm{w}}^*$, $T_{\text{refine}} = T - T_{\text{fast}}$.
%\State For each state-action pair $(s,a)$, we sample the model $\mathcal{M}$ for $N_0$ times.
\State Parameters: step size $\gamma_t$.
\State Initialize $\bm{w}_1 = \hat{\bm{w}}^*$ and $\bm{B}_1 = \bm{d}^\top\Phi\cdot T$.
\For{$t=1,...T_{\text{refine}}$}
\State Observe input data $(r_t, \bm{a}_t)$ and set:
\[\tilde{x}_t = 
            \begin{cases}
            1, ~~ r_t > \bm{a}_t^\top \Phi \bm{w}_t \\
            0, ~~ r_t \le \bm{a}_t^\top \Phi \bm{w}_t
            \end{cases}\]
\State Set $x_t = \tilde{x}_t \cdot \mathbb{I}\{\bm{a}_t^\top\Phi \tilde{x}_t \le \bm{B}_t\}$
\State Update $ \label{const_reg:update_w}
\bm{w}_{t+1} = \max\{ \bm{w}_t + \gamma_t\left(\bm{a}_t^\top\Phi \tilde{x}_t - \bm{d}^\top\Phi\right), \bm{0}\}$
\State Update $\bm{B}_{t+1} = \bm{B}_t - \bm{a}_t^\top\Phi x_t$
\EndFor
\State \textbf{Output}: $\bm{w}=(\bm{w}_1,...\bm{w}_{T_{\text{refine}}})$.
\end{algorithmic}
\end{algorithm}

The following \Cref{thm:const_reg_bound} gives the regret bound of \Cref{alg:const_reg} and the full proof is presented in \Cref{appendix:const_reg}.
\begin{theorem} \label{thm:const_reg_bound}
    Under the stochastic input model, \Cref{asp:bounded_data}, and \Cref{asp:GPG}, if selecting step size $\gamma_t \le \frac{1}{T}$ for $t=T_{\text{fast}}+1,...,T$, the regret of \Cref{alg:const_reg} satisfies
    \[
    \text{Reg}_{T_{\text{refine}}}(\pi_{\text{ALG4}}) \le  O\left(\frac{q}{\eps}\right).
    \]
\end{theorem}

\subsection{Putting it together}
%In previous sections, we have discussed two stages of our algorithm separately. Now we put them together and present our final algorithm in \Cref{alg:log_regret}. In \Cref{alg:fast}, we successfully obtain a near optimal solution with $O(\log{T})$ regret and we input it to our \Cref{alg:const_reg}. 
In the previous sections, we have discussed the two distinct stages of our algorithm. Now, we integrate them into a unified framework, which is outlined in \Cref{alg:log_regret}. In \Cref{alg:fast}, we have obtained a near-optimal solution with an $O(\log{T})$ regret bound using $O(\log{T})$ time steps. This solution is then fed into \Cref{alg:const_reg}, which further refines the dual variable $\bm{w}_t$ within the remaining time and achieves a constant regret. By combining the strengths of these two algorithms, we successfully achieves a $O(\log{T}+1/\eps)$ regret bound, beating previous $O(\sqrt{T})$ regret. We present our final result in \Cref{thm:log_reg_bound}, which is an immediate result combining \Cref{thm:fast} and \Cref{thm:const_reg_bound}.

\begin{algorithm}[ht!]
%\small
\caption{Accelerate-Then-Refine Algorithm}
\label{alg:log_regret}
%\SetAlgoLined
\begin{algorithmic}[1]
\State \textbf{Input:} $\bm{d} = \frac{\bm{b}}{T}$, time horizon $T$.
\State Compute $T_\text{fast}$ and $T_\text{refine} = T - T_\text{fast}$.
\For{$t=1,...T_\text{fast}$}
\State Run Accelerate \Cref{alg:fast}.
\EndFor
\For{$t=T_\text{fast}+1,...T$}
\State Run Refine \Cref{alg:const_reg}.
\EndFor
\State \textbf{Output}: $\bm{x}=(x_1,...x_T)$, $\bm{w}=(\bm{w}_1,...\bm{w}_T)$.
\end{algorithmic}
\end{algorithm}

\begin{theorem} \label{thm:log_reg_bound}
    Under the stochastic input model, \Cref{asp:bounded_data}, \Cref{asp:GPG} with parameter $\eps$ and \Cref{asp:holder}, if selecting step size $\gamma_t = \frac{1}{\log{T}}$ for $t=1,...,T_{\text{fast}}$ and $\gamma_t \le \frac{1}{T}$ for $t=T_{\text{fast}+1,...,T}$ in \Cref{alg:log_regret}, the regret of \Cref{alg:log_regret} satisfies that with high probability $1-\delta$:
    \[
    \text{Reg}_T(\pi_{\text{ALG5}}) \le  O\left(q\log{T+\frac{q}{\eps}}\right).
    \]
\end{theorem}

\section{Extension to General Function Settings} \label{sec:general}
In previous sections, we have proposed two first-order algorithms for online linear programming. Now we extend our algorithms to a more general setting: at each time step $t$, we observe a reward function $f(x_t; \bm{\theta_t}): \mathcal{X} \rightarrow \mathbb{R}$ and cost functions $\bm{g}(x_t; \bm{\theta_t}): \mathcal{X} \rightarrow \mathbb{R}^m$, where $\mathcal{X}$ is a convex and compact set and $\bm{\theta_t} \in \Theta \subset \mathbb{R}^l$ is the parameter to be revealed at time $t$. In particular, when $\bm{\theta_t} = (r_t, \bm{a}_t)$ the optimization problem reduces to the online LP problem discussed in the previous sections. We now formally present this general function setting in the below:

\begin{equation} \label{general:primal}
\begin{aligned}
    \max\ &\sum_{t=1}^T f(x_t; \bm{\theta_t}) \\ 
    \mathrm{s.t.\ } &\sum_{t=1}^T\bm{g}_i(x_t; \bm{\theta_t}) \le \bm{b}_i \\
    & \bm{x} \in \mathcal{X}
\end{aligned}
\end{equation}

Similar to \Cref{asp:bounded_data}, we make following assumptions throughout the whole section:
\begin{assumption} \label{asp:general_bounded}
    Without loss of generality, we assume that

    i) The functions $f(x_t; \bm{\theta_t})$ are concave over all $x_t \in \mathcal{X}$ and $\bm{g}(x_t; \bm{\theta_t})$ are convex over all $x_t \in \mathcal{X}$ for all possible $\bm{\theta_t} \in \Theta$.
    
    ii) $\left|f(x_t; \bm{\theta_t})\right| \le \overline{F}$ for all $x_t \in \mathcal{X},~\bm{\theta_t} \in \Theta$.

    iii) $\left\|\bm{g}(x_t; \bm{\theta_t})^\top\Phi\right\|_\infty \le \overline{G}$ for all $x_t \in \mathcal{X},~\bm{\theta_t} \in \Theta$. In addition, $0 \in \mathcal{X}$ and $g_i(0; \bm{\theta}) = 0$ for all $\bm{\theta} \in \Theta$.

    iv) $\mathcal{\theta}_t$ are $i.i.d.$ sampled from some distribution $\mathcal{P}$.

    v) There exist two positive constants $\underline{D}$ and $\overline{D}$ such that, $\underline{D} \le \left\|\bm{d}^\top \Phi\right\|_\infty \le \overline{D}$.
\end{assumption}

We define the following expectation for  and a probability measure $\mathcal{P}$ in the parameter space $\Theta$,
We then evaluate the function $u(x; \bm{\theta}): \mathcal{X} \rightarrow \mathbb{R}$ in expectation over $\Theta$ by defining the operator
\[
\mathcal{P} u(x(\bm{\theta}_t); \bm{\theta}) = \int_{\bm{\theta}'\in\Theta} u(x(\bm{\theta}'_t); \bm{\theta}')d\mathcal{P}(\bm{\theta}')
\]
where $x(\bm{\theta}) : \Theta \rightarrow \mathcal{X}$ is a measurable function. This construction aggregates the values $u(x(\bm{\theta}'_t); \bm{\theta}')$ across the parameter space according to $\mathcal{P}$.
Consequently $\mathcal{P}u(\cdot)$ can be interpreted as a deterministic functional: given the mapping $x(\cdot)$, it returns a single scalar obtained by averaging with respect to the underlying parameter distribution. With this operator in place, we next introduce a function that characterizes the optimization problem once the parameter $\bm{\theta}_t$ has been observed.
%Define a following function representing the optimization once a parameter $\bm{\theta}_t$ is observed:
\[
s(\bm{w};\bm{\theta}_t) = \max_{w\ge0} \{f(x(\bm{\theta}_t);\bm{\theta}_t) - \bm{g}(x(\bm{\theta}_t);\bm{\theta}_t)^\top\Phi \bm{w}\}
\]
Then we can directly apply our new reformulation in \Cref{sec:new_lp} and obtain the following approximation:
\begin{equation} \label{general:approx_dual}
\begin{aligned}
    \min_{\bm{w}}\ &h_{T,\Phi}(\bm{w}) = \bm{d}^{\top}\Phi \bm{w} + \frac{1}{T}\sum^T_{t=1}\mathcal{P}_ts(\bm{w};\bm{\theta}_t) \\
    \mathrm{s.t.\ } &\bm{w} \ge 0
\end{aligned}
\end{equation}
where $\bm{x}(\bm{\theta}_t) = (x_1(\bm{\theta}_t),...,x_t(\bm{\theta}_t))$ are primal decisions. Note that all primal decisions $\bm{x}(\bm{\theta}_t)$ are dependent on the parameter $\bm{\theta}_t$ since our algorithm always observes new input data first and then make a decision. 

\subsection{Mirror-Descent-Based Algorithm}
In this subsection we extend our \Cref{alg:mirror_sto} to this general function settings and we present it in \Cref{alg:general_md}. This extension highlights that the core mechanism of our approach is not tied to the linear programming structure, but instead relies on the mirror-descent template with a suitable choice of potential function $\psi$. Therefore we omit the extension of \Cref{alg:main}, as it can be viewed as a special case of \Cref{alg:mirror_sto} when selecting $\psi(u) = \frac{1}{2}\left\|u\right\|_2^2$. Moreover, in \Cref{sec:mirror} we have made two additional common assumptions and we will not repeat them here to avoid redundancy. 
Compared with the linear case, the main technical subtlety in this general function setting lies in Step $4$: we now need to find some $x$ to maximize function $s(x;\bm{\theta})$, while in the linear case we can immediately find that $x=1$ is the optimal solution to this optimization problem.
\begin{algorithm}[ht!]
%\small
\caption{Mirror-descent-based algorithm under general function setting}
\label{alg:general_md}
%\SetAlgoLined
\begin{algorithmic}[1]
\State \textbf{Input:} $\bm{d} = \frac{\bm{b}}{T}$.
%\State For each state-action pair $(s,a)$, we sample the model $\mathcal{M}$ for $N_0$ times.
\State Initialize $\bm{w}_1 = 0$.

\For{$t=1,...T$}
\State Observe input parameter $\bm{\theta}_t$ and solve 
\begin{equation} \label{eq:def_x}
    x_t = \arg \max_{x\in\mathcal{X}} \{f(x;\bm{\theta}_t) - \bm{g}(x;\bm{\theta}_t)^\top\Phi \bm{w}_t\}
\end{equation}

\State Update $\bm{y}_t = \bm{d}^\top\Phi - \bm{g}(x_t;\bm{\theta}_t)^\top\Phi x_t$.
\State Update $\bm{w}_{t+1} = \arg \min\limits_{\bm{w}\ge0} \{ \langle \gamma_t \bm{y}_t, \bm{w} \rangle + D_{\psi}(\bm{w}||\bm{w}_t) \}$.
\EndFor
\State \textbf{Output}: $\bm{x}=(\bm{x}_1,...\bm{x}_T)$.

\end{algorithmic}
\end{algorithm}

Importantly, \Cref{alg:general_md} achieves the same order of regret bound as in the linear case. This shows that our analysis does not hinge on linearity, and that the performance guarantee carries over to a significantly broader class of objectives and constraint structures. In other words, the linear-programming formulation should be viewed as an instantiation of a more general methodology rather than a restriction of our framework. We formalize our result in the following:
\begin{theorem} \label{thm:general_md}
    Under the stochastic input model, \Cref{assump:sepa} and \Cref{asp:general_bounded}, if the step size $\gamma_t = \frac{1}{\sqrt{T}}$ for $t=1,..,T$ in \Cref{alg:general_md}, the regret and expected constraints violation of \Cref{alg:general_md} satisfy
    \begin{equation}
        \text{Reg}_T(\pi_{\text{ALG6}}) \le O(q\sqrt{T})~~\text{and}~~v(\pi_{\text{ALG6}}) \le O(\sqrt{qT}). \nonumber
    \end{equation}
\end{theorem}

\subsection{Two Stage Algorithm}

Now we extend our two stage algorithm into the general function settings and present it in \Cref{alg:general_log}. 
Before analyzing its performance, we first establish an upper bound for the dual variables $\bm{w}_t$ generated in \Cref{alg:general_log}. Concretely, we prove that $\left\|\bm{w}_t\right\|_2$ remains bounded by a problem-dependent constant that does not grow with $t$ or $T$. This boundedness result is crucial: it prevents the dual iterates from exploding, ensures that the primal updates are performed with well-controlled effective step sizes, and allows us to bound the magnitude of the terms in the Lagrangian that involve the dual variables $\bm{w}_t$, which appear in the regret decomposition.

\begin{lemma} \label{lem:general_w_upper_bound}
Under \Cref{asp:general_bounded}, the dual variable $\bm{w}_t$ in \Cref{alg:general_log} can be upper bounded by
\begin{equation}
\begin{aligned}
    ||w_t||_2 & \le \frac{q(\overline{G}+\overline{D})^2 +4\overline{F}}{2\underline{D}} +q(\overline{G}+\overline{D})
\end{aligned}
\end{equation}
\end{lemma}

\begin{algorithm}[ht!]
%\small
\caption{$\log{T}$ regret Algorithm under general function setting}
\label{alg:general_log}
%\SetAlgoLined
\begin{algorithmic}[1]
\State \textbf{Input:} $\bm{d} = \frac{\bm{b}}{T}$, total time horizon $T$, probability $\delta$, initial error estimate $\eps_0 = \frac{\overline{r}}{\underline{D}}$.
\State Set $L = \lceil\log{(\eps_0T)}\rceil$, $\hat{\delta} = \log{\frac{\delta}{K}}$, $J \ge \max \{\frac{9(\overline{G}+\overline{D})^2}{\lambda^2}, \frac{1152(\overline{G}+\overline{D})^2\log{(1/\hat{\delta})}}{\lambda^2}\}$ and $T_{\text{fast}} = L\cdot J$
\State Initialize $\bm{w}_0 = \tilde{\bm{w}}_0 = 0$; step size $\eta_1 = \frac{\eps_0}{3\overline{r}^2}$ and $\gamma_t = \frac{1}{\log{T}}$; $V_1 \ge \frac{\eps_0}{\lambda}$; $\bm{B}_0 = \bm{b}^\top\Phi$.
\For{$l=1,...L$}
\State Set $\tilde{\bm{w}}_1^l$ = $\tilde{\bm{w}}_{l-1}$ and domain $\mathcal{K} = \mathbb{R}_{\ge0}\cap\mathcal{B}(\tilde{\bm{w}}_{l-1},V_k)$.
\For{$j=1,...J$} 

\State Compute current time step $t = (l-1)*J + (j-1)$.
\State Observe input parameter $\bm{\theta}_t$ and solve 
\begin{equation} \label{eq:def_x}
    \tilde{x}_t = \arg \max_{x\in\mathcal{X}} \{f(x;\bm{\theta}_t) - \bm{g}(x;\bm{\theta}_t)^\top\Phi \bm{w}_t\}
\end{equation}
\State Set $x_t = \tilde{x}_t \cdot \mathbb{I}\{\bm{g}(\tilde{x}_t;\bm{\theta}_t)^\top\Phi \tilde{x}_t \le \bm{B}_t\}$
\State Update $\bm{B}_{t+1} = \bm{B}_t - \bm{a}_t^\top\Phi x_t$
\State Update $
\bm{w}_{t+1} = \max\{
\bm{w}_t +\gamma_t(\bm{g}(\tilde{x}_t;\bm{\theta}_t)^\top\Phi \tilde{x}_t - \bm{d}^\top\Phi), \bm{0}\}$

\State Update $\tilde{\bm{w}}_{j+1}^l = \prod_{\mathcal{K}}\left[\tilde{\bm{w}}_j^l + \eta_l\left(\gamma_t(\bm{g}(\tilde{x}_t;\bm{\theta}_t)^\top\Phi \tilde{x}_t - \bm{d}^\top\Phi\right)\right]$

\EndFor
\State Set $\tilde{\bm{w}}_l = \frac{1}{J}\sum_{j=1}^J \tilde{\bm{w}}_j^l$.
\State Set $\eta_{l+1} = \frac{\eta_l}{2}$ and $V_{l+1} = \frac{V_l}{2}$.
\EndFor

\State Compute $T_{\text{refine}} = T - T_{\text{fast}}$ and set $\gamma_t = \frac{1}{T}, \bm{w}_1 = \tilde{\bm{w}}_L, \bm{B}_1 = \bm{d}^\top\Phi\cdot T$.
%\State For each state-action pair $(s,a)$, we sample the model $\mathcal{M}$ for $N_0$ times.
\For{$t=1,...T_{\text{refine}}$}
\State Observe input parameter $\bm{\theta}_t$ and solve 
\begin{equation} \label{eq:def_x}
    \tilde{x}_t = \arg \max_{x\in\mathcal{X}} \{f(x;\bm{\theta}_t) - \bm{g}(x;\bm{\theta}_t)^\top\Phi \bm{w}_t\}
\end{equation}
\State Set $x_t = \tilde{x}_t \cdot \mathbb{I}\{\bm{g}(\tilde{x}_t;\bm{\theta}_t)^\top\Phi \tilde{x}_t \le \bm{B}_t\}$
\State Update $
\bm{w}_{t+1} = \max\{ \bm{w}_t + \gamma_t\left(\bm{g}(\tilde{x}_t;\bm{\theta}_t)^\top\Phi \tilde{x}_t - \bm{d}^\top\Phi\right), \bm{0}\}$
\State Update $\bm{B}_{t+1} = \bm{B}_t - \bm{g}(\tilde{x}_t;\bm{\theta}_t)^\top\Phi x_t$
\EndFor
\State \textbf{Output}: $\bm{x}=(x_1,...x_T)$.
\end{algorithmic}
\end{algorithm}

Building on this boundedness property, we now analyze the regret of \Cref{alg:general_log}. Using the same two-stage decomposition as in \Cref{alg:log_regret}, we show that the cumulative error terms contributed by the primal and dual updates can be dominated by a logarithmic function of the horizon. As a result, \Cref{alg:general_log} achieves an $\log T$ regret bound in the general function setting as well, demonstrating that the improved regret guarantee is not an artifact of linearity but a consequence of the algorithmic structure and the stability of the dual sequence.
\begin{theorem} \label{thm:general_log}
    Under the stochastic input model, \Cref{asp:GPG} with parameter $\eps$, \Cref{asp:holder} and \Cref{asp:general_bounded}, if selecting step size $\gamma_t = \frac{1}{\log{T}}$ for $t=1,...,T_{\text{fast}}$ and $\gamma_t \le \frac{1}{T}$ for $t=T_{\text{fast}+1,...,T}$ in \Cref{alg:log_regret}, the regret of \Cref{alg:general_log} satisfies that with high probability $1-\delta$:
    \[
    \text{Reg}_T(\pi_{\text{ALG7}}) \le  O(q\log{T}+\frac{q}{\eps}). 
    \]
\end{theorem}

\section{Numerical Experiments}
\subsection{Experiment Setup}
A key step in our algorithm is the use of \textit{non-negative} basis and now we explain how to construct it in our experiments in detail. The dimension of weights $w$ is $10$, thereby the basis function $\Phi$ is a $m \times 10$ matrix. We select Gaussian-kernel RBF, which has the inherent property of non-negativity:
\[
\phi(u_i;c,\sigma) = \exp (-\frac{(u_i - c)^2}{2\sigma^2})
\]
where $u_i$ is a one-dimensional continuous coordinate, $c$ is the center of RBF and $\sigma$ is the scale parameter. We embed the constraint index into $u_i$:
\[
u_i = \frac{i - 0.5}{m}, ~i = 1,...,m
\]
where $m$ is the dimension of constraints. The RBF columns adopt a dual-resolution setup, with the coarse layer accounting for $\alpha=0.6$ and the fine layer for $1-\alpha=0.4$. Denote the number of coarse layer centers by $K_c$ and the number of fine layer centers by $K_f$, we have:
\[
K_c = \lceil \alpha K \rceil,~~K_f = K - K_c
\]
Denote the adjacent overlap degree by $\rho$ and the intervals between centers by $\Delta = \frac{1}{K_c - 1}$, and then we can calculate the bandwidth according to the following formula:
\[
\sigma(\rho) = \frac{\Delta}{2\ln(\frac{1}{\rho})}
\]
where we select $\rho_{\text{coarse}}=0.6$ and $\rho_{\text{fine}}=0.3$ to obtain both coarse and fine bandwidths. 
Further, we determine the coarse layer centers $\mathcal{C}_c = \{c_k^{(c)}\}_{k=1}^{K_c}$ and fine layer centers $\mathcal{C}_f = \{c_l^{(f)}\}_{l=1}^{K_f}$ as follows:
\[
c_k^{(c)} = \frac{k - 1}{K_c - 1},~~c_l^{(f)} = (\frac{\Delta}{2}+(l - 1)\Delta) - \lfloor (\frac{\Delta}{2}+(l - 1)\Delta) \rfloor
\]
In our experiments, the dual variable $p$ is Finally, we can assign a specific bandwidth for each center:
\[
\sigma_j = 
\begin{cases}
    \sigma_c, ~c_j \in \mathcal{C}_c \\
    \sigma_f, ~c_j \in \mathcal{C}_f
\end{cases}
\]
where $\sigma_c = \sigma(\rho_{\text{coarse}})$ and $\sigma_f = \sigma(\rho_{\text{fine}})$.

\subsection{Experimental Results}
\subsubsection{Stochastic Input Results}
In our experiments, we adopt the linear setting for simplicity. For stochastic input model, first we $i.i.d.$ sample right-hand-side $d$ from uniform distribution $U[2,3]$. To satisfy the GPG assumption, we consider finite support setting for input data $\{(r_t,\bm{a}_t)\}_{t=1}^T$, where we first sample an input set $\{(r_k,\bm{a}_k)\}_{k=1}^K$ from some distribution and then sample our actual input data from this input set.
In addition, we repeat each trial for $100$ times, compute its average result and construct the nominal $95\%$ confidence interval $C^{(i)}_{0.95}$ for each replication $i$. We compare \Cref{alg:log_regret} (Ours-log) with \Cref{alg:mirror_sto} (Ours-MD) with potential function $\psi(u)=\frac{1}{2}||u||_2^2$ and the simple and fast algorithm proposed by \citet{li2023simple} (Simple-GD). 
We conduct three different experiments under a fixed number of constraints $M=2000$: i)
$r_k,\bm{a}_k$ are $i.i.d.$ sampled from $U[0,1]$ and $U[0,4]$; ii) $r_k,\bm{a}_k$ are $i.i.d.$ sampled from $N[1,1]$ and $N[4,1]$; iii) $r_k,\bm{a}_k$ are $i.i.d.$ sampled from $\text{Cauchy}[0,1]$ and $\text{Cauchy}[2,1]$. Note that the first and second experiment are intended to verify the robustness of three algorithms and the third experiment is aimed to validate their robustness under extreme numerical conditions, given that the Cauchy distribution has a heavy-tailed nature.

\begin{figure}[htbp]
    \centering 
    \caption{Uniform Distribution with $M = 2000$}
    \label{fig:uniform_fixM}
    \begin{subfigure}[t]{0.45\linewidth}
    \centering
    \includegraphics[width=\linewidth]{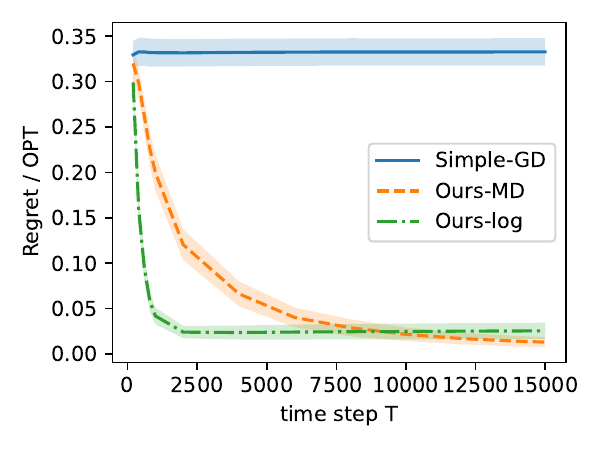}
    \caption{Regret}
    \label{fig:uniform_fix_M_regret}
    \end{subfigure}\hfill
    \begin{subfigure}[t]{0.45\linewidth}
    \centering
    \includegraphics[width=\linewidth]{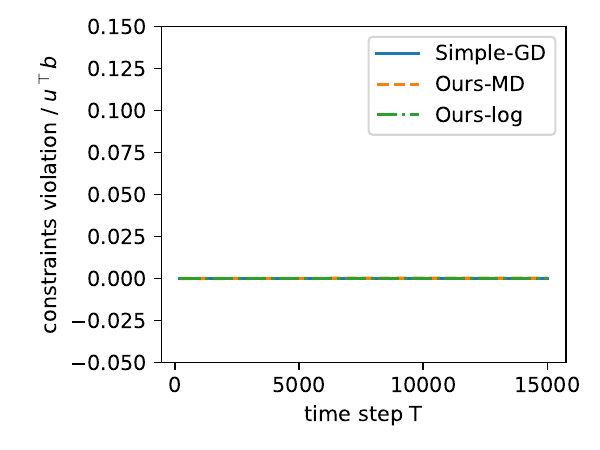}
    \caption{Constraints violation}
    \label{fig:uniform_fix_M_constr}
    \end{subfigure}\hfill
\end{figure}

\begin{figure}[htbp]
    \centering 
    \caption{Normal Distribution with $M = 2000$}
    \label{fig:normal_fixM}
    \begin{subfigure}[t]{0.45\linewidth}
    \centering
    \includegraphics[width=\linewidth]{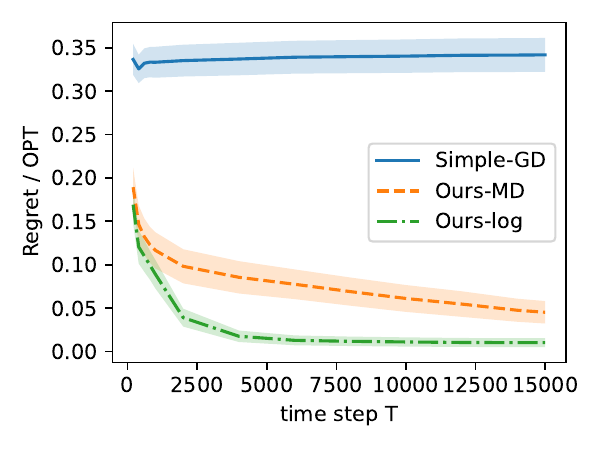}
    \caption{Regret}
    \label{fig:normal_fix_M_regret}
    \end{subfigure}\hfill
    \begin{subfigure}[t]{0.45\linewidth}
    \centering
    \includegraphics[width=\linewidth]{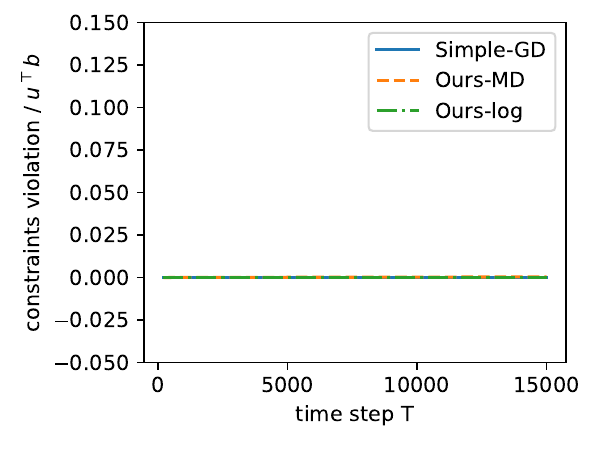}
    \caption{Constraints violation}
    \label{fig:normal_fix_M_constr}
    \end{subfigure}\hfill
\end{figure}

\begin{figure}[htbp]
    \centering 
    \caption{Cauchy Distribution with $M = 2000$}
    \label{fig:cauchy_fixM}
    \begin{subfigure}[t]{0.45\linewidth}
    \centering
    \includegraphics[width=\linewidth]{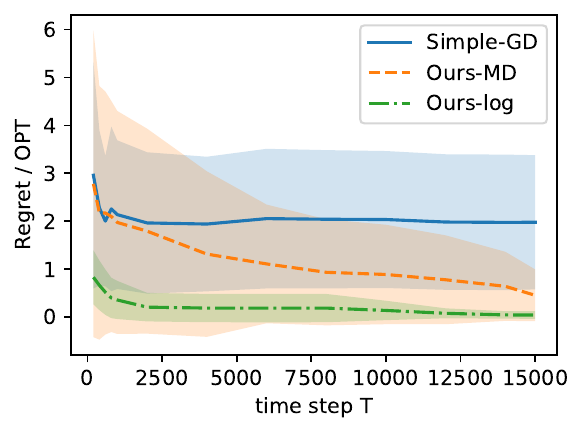}
    \caption{Regret}
    \label{fig:cauchy_fix_M_regret}
    \end{subfigure}\hfill
    \begin{subfigure}[t]{0.45\linewidth}
    \centering
    \includegraphics[width=\linewidth]{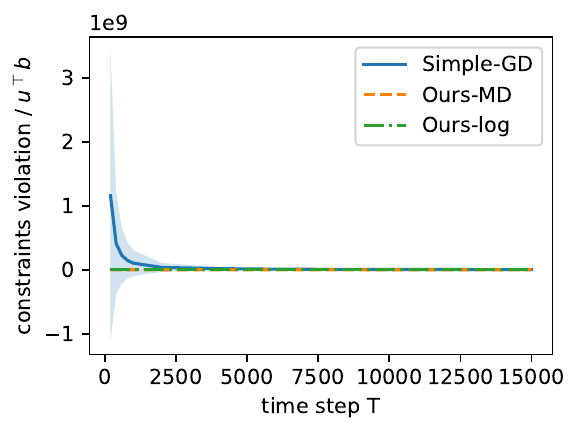}
    \caption{Constraints violation}
    \label{fig:cauchy_fix_M_constr}
    \end{subfigure}\hfill
\end{figure}

\Cref{fig:uniform_fixM} presents the results of three algorithms under uniform distribution. We can observe that the ratio of regret and optimal value of both our \Cref{alg:log_regret} and \Cref{alg:mirror_sto} can converge to $0$, and \Cref{alg:log_regret} converges faster, which is consistent to our main results. In contrast, the Simple-GD algorithm has poor performance and the curve of it in \Cref{fig:uniform_fix_M_regret} implies an $O(T)$ regret. This is because the regret bound of Simple-GD algorithm is $O(\sqrt{T})$ and linearly dependent on the number of constraints, which is quite large and dominates the regret in our setting. In addition, \Cref{fig:uniform_fix_M_constr} shows that all three algorithms can satisfy the constraints quite well. 
We can also observe similar results in \Cref{fig:normal_fixM} \Cref{fig:cauchy_fixM}, which validates the effectiveness and robustness of our two algorithms.

\begin{figure}[htbp]
    \centering
    \caption{Cauchy Distribution with $T = 5000$}
    \label{fig:cauchy_fixT}
    \begin{subfigure}[t]{0.45\linewidth}
    \centering
    \includegraphics[width=\linewidth]{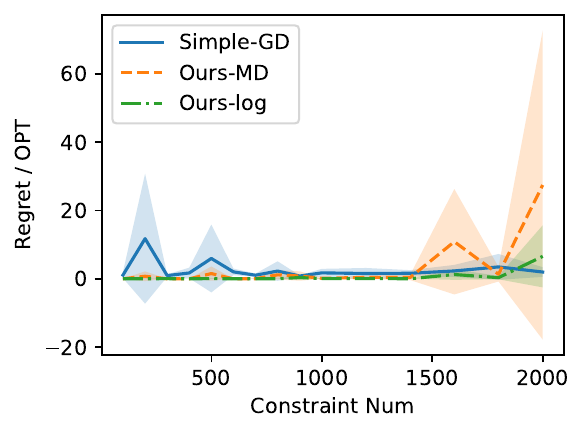}
    \caption{Regret}
    \label{fig:cauchy_fix_T_regret}
    \end{subfigure}\hfill
    \begin{subfigure}[t]{0.45\linewidth}
    \centering
    \includegraphics[width=\linewidth]{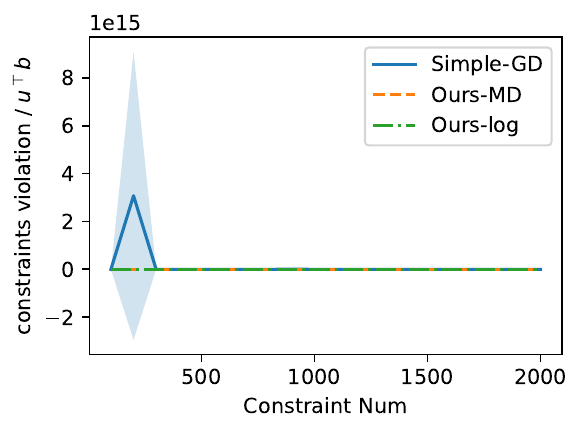}
    \caption{Constraints violation}
    \label{fig:cauchy_fix_T_constr}
    \end{subfigure}\hfill
\end{figure}

Furthermore, we conduct experiments under fixed time horizon $T=5000$ and observe both regret and constraint violation under different number of constraints. We sample input data $\{(r_t,\bm{a}_t)\}_{t=1}^T$ in the same way of previous fixed number of constraints experiments. We can observe that for the uniform and normal distribution experiment, even when the number of constraints is relative small (less than $500$), both Ours-log and Ours-MD enjoys good performance, whereas Simple-GD already has poor performance.
When the number of constraints become larger, Ours-log is quite robust with stable performance. In addition, Simple-GD is prone to severe performance degradation when faced with extreme data as shown in \Cref{fig:cauchy_fixT}.

\begin{figure}[htbp]
    \centering
    \caption{Uniform Distribution with $T = 5000$}
    \label{fig:uniform_fixT}
    \begin{subfigure}[t]{0.45\linewidth}
    \centering
    \includegraphics[width=\linewidth]{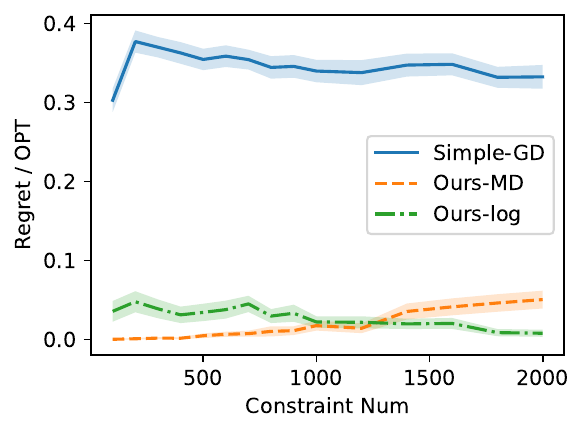}
    \caption{Regret}
    \label{fig:uniform_fix_T_regret}
    \end{subfigure}\hfill
    \begin{subfigure}[t]{0.45\linewidth}
    \centering
    \includegraphics[width=\linewidth]{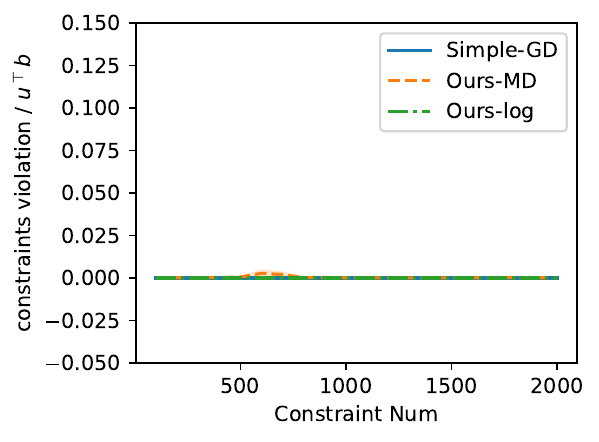}
    \caption{Constraints violation}
    \label{fig:uniform_fix_T_constr}
    \end{subfigure}\hfill
\end{figure}

\begin{figure}[htbp]
    \centering
    \caption{Normal Distribution with $T = 5000$}
    \label{fig:normal_fixT}
    \begin{subfigure}[t]{0.45\linewidth}
    \centering
    \includegraphics[width=\linewidth]{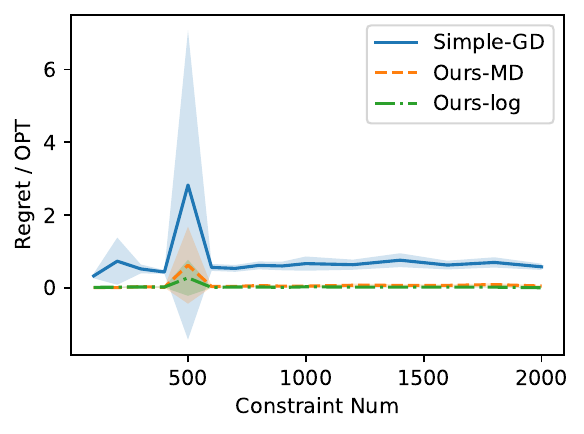}
    \caption{Regret}
    \label{fig:normal_fix_T_regret}
    \end{subfigure}\hfill
    \begin{subfigure}[t]{0.45\linewidth}
    \centering
    \includegraphics[width=\linewidth]{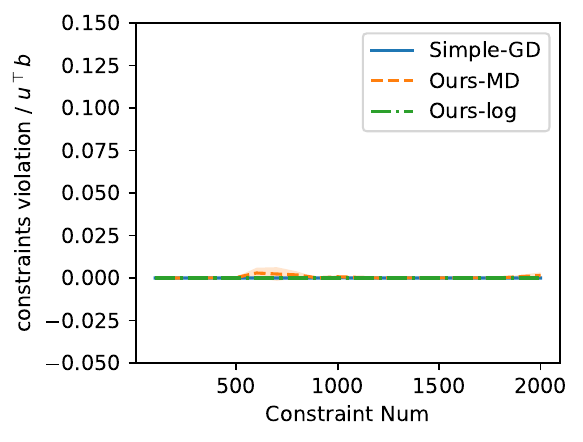}
    \caption{Constraints violation}
    \label{fig:normal_fix_T_constr}
    \end{subfigure}\hfill
\end{figure}

%Cref{fig:uniform_fix_M_regret} shows the results of fixed number of constraints $2000$. We can observe that although both Algorithm SF and our Algorithm MD have $O(\sqrt{T})$ theoretical regret bound, the curve of our Algorithm MD converges to zero at a rate of $O(1 / \sqrt{T})$, while Algorithm SF can only converge to $0.5$. This superiority stems from the fact that our Algorithm MD depends only on the dimension of weight $w$, which is $100$ is this experiment, rather than on the number of constraints. Conversely, the regret bound of Algorithm SF is linearly dependent on the number of constraints, leading a gap to our Algorithm MD.

\subsubsection{Random Permutation Results}

\begin{figure}[htbp]
    \centering
    \caption{Constraint Num = $2000$}
    \begin{subfigure}[htbp]{0.45\linewidth}
    \centering
    \includegraphics[width=\linewidth]{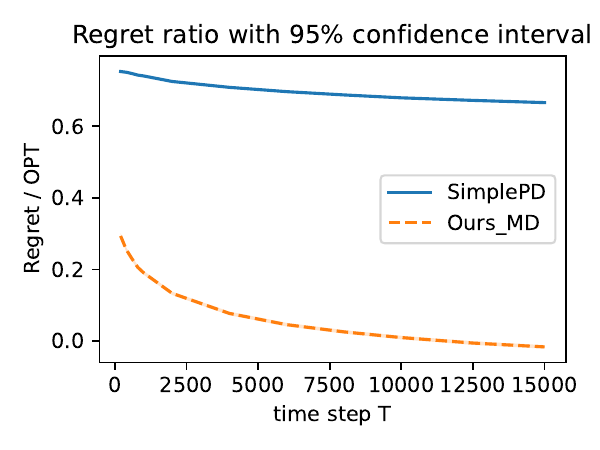}
    \caption{Regret}
    \label{fig:random_regret_fixM}
    \end{subfigure}\hfill
    \begin{subfigure}[htbp]{0.45\linewidth}
    \centering
    \includegraphics[width=\linewidth]{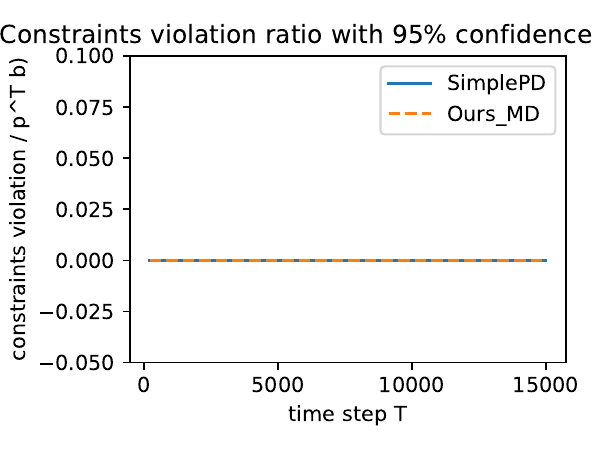}
    \caption{Constraints Violation}
    \label{fig:random_constr_fixM}
    \end{subfigure}\hfill
\end{figure}

\begin{figure}[htbp]
    \centering
    \caption{Time Horizon T = $5000$}
    \begin{subfigure}[htbp]{0.45\linewidth}
    \label{fig:random_fixT}
    \centering
    \includegraphics[width=\linewidth]{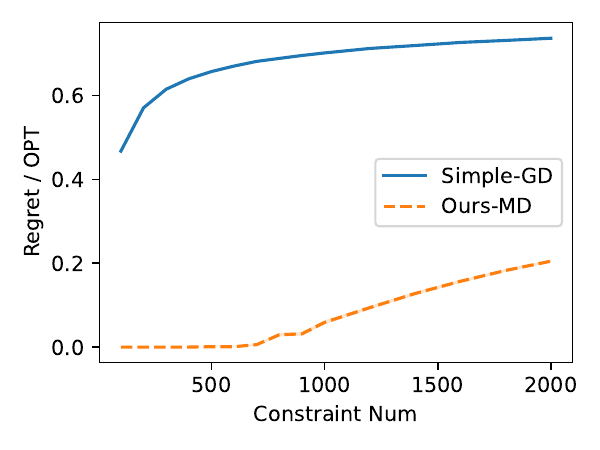}
    \caption{Regret}
    \label{fig:random_regret_fixT}
    \end{subfigure}\hfill
    \begin{subfigure}[htbp]{0.45\linewidth}
    \centering
    \includegraphics[width=\linewidth]{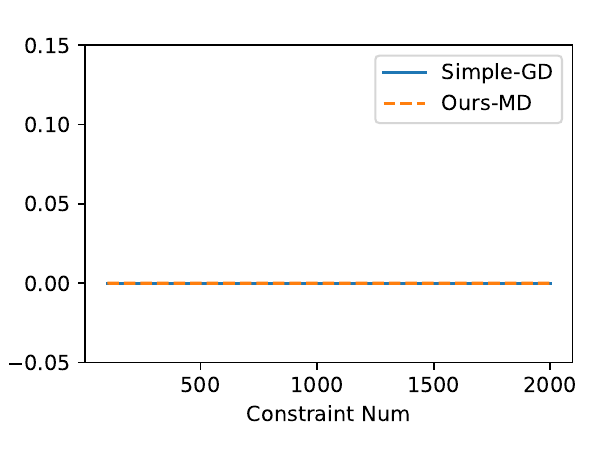}
    \caption{Constraints Violation}
    \label{fig:random_constr_fixT}
    \end{subfigure}\hfill
\end{figure}

For the random permutation model, we sample the input data $\bm{a}_{ij}$ from following five categories: i) Uniform distribution $U[0,3]$; ii) Normal distribution $N[2,1]$; iii) Normal distribution $N(1,1)$ with different mean; iv) Normal distribution with different standard deviation $N(0,1)$; v) an integer set $\{-1,0,1\}$. Besides, the input coefficients $r_t$ are sampled from $U[0,1]$ while the right-hand-side $d$ are sampled from $U[\frac{2}{5},\frac{4}{5}]$. 
\Cref{fig:random_regret_fixM} shows that both Algorithm SF and our Algorithm MD have $O(\sqrt{T})$ convergence rate, while our Algorithm MD converges quite faster and can converge to $0$. \Cref{fig:random_constr_fixM}
further shows that both two algorithms have excellent performance of constraints violation. The constraints violation of Algorithm GR and SF are all $0$ and our Algorithm MD violates the constraints very small. We also conduct experiments with fixed time horizon and various number of constraints. \Cref{fig:random_fixT} shows that Ours-MD can maintain good performance when the number of constraints is not that large, while Simple-GD's performance deteriorates steadily as the number of constraints increases.

\section{Conclusion}
In this paper, we investigate the online semi-infinite LP and develop a primal-dual algorithm. To address the most significant challenge posed by high-dimensional constraints, we developed a new LP formulation via function approximations, which successfully reduces the number of constraints to a constant. Then we develop a mirror descent algorithm and achieve $O(\sqrt{T})$ and $(q+q\log{T})\sqrt{T})$ regret bounds under the stochastic input model and random input model respectively. Further, we develop a novel two-stage algorithm for stochastic input model, achieving the $O(\log{T} + 1/\eps)$ regret bound and improving previous $O(\sqrt{T})$ regret bound. Numerical results demonstrate that our algorithms have excellent performance especially confronted with large-scale problem.

\bibliographystyle{abbrvnat}
\bibliography{bibliography}

\clearpage

\begin{APPENDICES}
\crefalias{section}{appendix}

\section{Auxiliary results}
\begin{lemma}[Azuma's Inequality]
Let $(\mathcal{F}_k)_{k=0}^n$ be a filtration and let $(d_k,\mathcal{F}_k)_{k=1}^n$ be a martingale difference sequence, i.e.,
\[
\mathbb{E}[d_k \mid \mathcal{F}_{k-1}] = 0 \quad \text{a.s.}, \qquad k=1,\dots,n.
\]
Define $S_n := \sum_{k=1}^n d_k$. Assume that there exist constants $c_1,\dots,c_n \ge 0$ such that
\[
|d_k| \le c_k \quad \text{a.s.}, \qquad k=1,\dots,n.
\]
Then for any $t>0$,
\[
\mathbb{P}(|S_n| \ge t) \le 2\exp\!\left(-\frac{t^2}{2\sum_{k=1}^n c_k^2}\right).
\]
\end{lemma}

\begin{lemma}[Negative drift \citep{gupta2024greedy}] \label{lem:gupta}
Under \Cref{asp:GPG}, let $\Phi_t$ be the $\mathcal{F}_t$-measurable stochastic process and suppose the following conditions hold:

i) Bounded Variation: $\left\|\Phi_{t+1} - \Phi_t\right\| \le Z$.

ii) Expected Decrease: When $\Phi_t \ge V$, $\mathbb{E}\left[\Phi_{t+1} - \Phi_t | \Phi_t\right] \le -2\eta Z$.

Then the stochastic process $\Phi_t$ can be upper bounded as:
\[
\mathbb{E}\left[\Phi_t\right] \le Z\left(1+\lceil\frac{V}{Z}\rceil\right) + \frac{Z^2}{2\eta} - \frac{Z}{2}.
\]

\end{lemma}

\section{Proof of \Cref{sec:simple}}

\subsection{Proof of \Cref{lemma:w_upper_bound}}

From the updating formula \eqref{main:update_w}, we have:
\begin{equation} \label{ieq:w}
\begin{aligned}
    ||\bm{w}_{t+1}||_2^2 & \le ||\bm{w}_t + \gamma_t(\bm{a}^\top_t\Phi x_t -\bm{d}^\top \Phi)||_2^2\\
    &= ||\bm{w}_t||_2^2 - 2\gamma_t(\bm{d}^\top\Phi - \bm{a}_t^\top\Phi x_t) \bm{w}_t +\gamma_t^2||\bm{d}^\top\Phi - \bm{a}^\top_t\Phi x_t||_2^2 \\
    &\le ||\bm{w}_t||_2^2 - 2\gamma_t\bm{d}^\top\Phi \bm{w}_t +2\gamma_t\bm{a}_t^\top\Phi x_t \bm{w}_t +\gamma_t^2q(\overline{C}+\overline{D})^2 \\
    &= ||\bm{w}_t||_2^2 +\gamma_t^2q(\overline{C}+\overline{D})^2 + 2\gamma_t\bm{a}_t^\top\Phi \bm{w}_t\mathbb{I}\{r_t > \bm{a}_t^\top\Phi \bm{w}_t\} - 2\gamma_t\bm{d}^\top\Phi \bm{w}_t \\
    &\le ||\bm{w}_t||_2^2 +\gamma_t^2q(\overline{C}+\overline{D})^2 + 2\gamma_tr_t - 2\gamma_t\bm{d}^\top\Phi \bm{w}_t
\end{aligned}
\end{equation}
where the first inequality follows from the update formula \eqref{ieq:w}, the second inequality relies on \Cref{asp:bounded_data} and the third inequality utilizes the condition of indicator function $\mathbb{I}$.
Next, we will show that when $||\bm{w}_t||_2$ is large enough, there must have $||\bm{w}_{t+1}||_2 \le ||\bm{w}_t||_2$. To be specific, when $||\bm{w}_t||_2 \ge \frac{q(\overline{C}+\overline{D})^2 +2\overline{r}}{2\underline{D}}$, we have the following inequalities:
\begin{equation} \label{w_bound:first}
\begin{aligned}
    ||\bm{w}_{t+1}||_2^2 - ||\bm{w}_t||_2^2 &\le \gamma_t^2q(\overline{C}+\overline{D})^2 + 2\gamma_tr_t - 2\gamma_t\bm{d}^\top\Phi \bm{w}_t \\
    &\le \gamma_t^2q(\overline{C}+\overline{D})^2 + 2\gamma_tr_t - 2\gamma_t \underline{D} \bm{w}_t \\
    &\le 0
\end{aligned}
\end{equation}
Otherwise, when $||\bm{w}_t||_2 < \frac{q(\overline{C}+\overline{D})^2 +2\overline{r}}{2\underline{D}}$, we have:
\begin{equation} \label{w_bound:second}
\begin{aligned}
    ||\bm{w}_{t+1}||_2 & \le ||\bm{w}_t + \gamma_t(\bm{a}_t^\top\Phi x_t -\bm{d}^\top \Phi)||_2\\
    &\le ||\bm{w}_t||_2 + ||\gamma_t(\bm{a}_t^\top\Phi x_t -\bm{d} \Phi)||_2 \\
    &\le \frac{q(\overline{C}+\overline{D})^2 +2\overline{r}}{2\underline{D}} +q(\overline{C}+\overline{D})
\end{aligned}
\end{equation}
Combining the two cases above, suppose $t=\tau$ is the first time that $||\bm{w}_t||_2 \ge \frac{q(\overline{C}+\overline{D})^2 +2\overline{r}}{2\underline{D}}$. Since $||\bm{w}_{\tau-1}||_2 < \frac{q(\overline{C}+\overline{D})^2 +2\overline{r}}{2\underline{D}}$, from the inequality \eqref{w_bound:second} we know that $||\bm{w}_{\tau}||_2 \le \frac{q(\overline{C}+\overline{D})^2 +2\overline{r}}{2\underline{D}}+q(\overline{C}+\overline{D})$. Besides, from the inequality \eqref{w_bound:first}, we know that $\bm{w}$ will decrease until it falls below the threshold $\frac{q(\overline{C}+\overline{D})^2 +2\overline{r}}{2\underline{D}}+q(\overline{C}+\overline{D})$.

Overall, we can conclude that:
\[
||\bm{w}_t||_2 \le \frac{q(\overline{C}+\overline{D})^2 +2\overline{r}}{2\underline{D}} +q(\overline{C}+\overline{D})
\]
which completes our proof.

\subsection{Proof of \Cref{thm:simple}}

First we have the following inequalities:
\begin{equation}
\begin{aligned}
    \text{Reg}_T(\pi_{\text{ALG1}}) &\le Tf(\bm{u}^*) - \sum^T_{t=1}r_t\mathbf{\textit{I}}(r_t> \bm{a}^\top_t\Phi \bm{w}_t) \\
           &\le Tf_\Phi(\bm{w}^*) - \sum^T_{t=1}r_t\mathbf{\textit{I}}(r_t> \bm{a}^\top_t\Phi \bm{w}_t) \\
           &\le \sum^T_{t=1}\mathbb{E}[f_\Phi(\bm{w}_t)] - \sum^T_{t=1}r_t\mathbf{\textit{I}}(r_t> \bm{a}^\top_t\Phi \bm{w}_t) \\
           &= \sum^T_{t=1}\mathbb{E}[(\bm{d}^\top\Phi-\bm{a}_t^\top\Phi x_t)\bm{w}_t]
\end{aligned}
\end{equation}
where the second inequality stems from \Cref{lemma:w_p} and third inequality depends on the fact that $\bm{w}^*$ is the optimal solution. Substitute $\gamma_t = \frac{1}{\sqrt{T}}$ into the inequality \eqref{ieq:w}, we have:
\begin{equation}
\begin{aligned}
    ||\bm{w}_{t+1}||_2^2 
    & \le ||\bm{w}_t||_2^2 - \frac{2}{\sqrt{T}}(\bm{d}^\top\Phi - \bm{a}_t^\top\Phi x_t) \bm{w}_t +\frac{q(\overline{C}+\overline{D})^2}{T}
\end{aligned}
\end{equation}
Combining two inequalities above, we have 
\begin{equation}
\begin{aligned}
    \text{Reg}_T(\pi_{\text{ALG1}}) &\le \sum^T_{t=1}\mathbb{E}[(\bm{d}^\top\Phi-\bm{a}_t^\top\Phi x_t)\bm{w}_t] \\
    &= \sqrt{T}\sum^T_{t=1}(||\bm{w}_{t}||_2^2 -  ||\bm{w}_{t+1}||_2^2) + q(\overline{C}+\overline{D})^2\sqrt{T} \\
    &\le q(\overline{C}+\overline{D})^2\sqrt{T}
\end{aligned}
\end{equation}
where the last inequality relies on the fact that $||\bm{w}_1||=0$. 

For the constraints violation, we first revisit the updating formula \eqref{main:update_w} in \Cref{alg:main} and it holds that:
\[
\bm{w}_{t+1} \ge \bm{w}_t +\gamma_t(\bm{a_t}^\top\Phi x_t - \bm{d}^\top\Phi)
\]
Therefore, if we set $\gamma_t = \frac{1}{\sqrt{T}}$, we can obtain:
\[ \label{ieq:thm1_constr}
\Phi(A\bm{x} - B) = \sum_{t=1}^T \bm{a_t}^\top\Phi x_t - Td\Phi \le \sqrt{T}(\bm{w}_{t+1} - \bm{w}_{1}) = \sqrt{T}\bm{w}_{t+1}.
\]
Finally, we have:
\[
\begin{aligned}
v(\pi_{\text{ALG1}}) &= \max_{\bm{w}}\mathbb{E}\left[ ||\left[\bm{w}_t^\top\Phi^\top(A\bm{x}-B)\right]^+||_2 \right] \\
&\le \mathbb{E}\left[ ||\left[\Phi^\top(A\bm{x}-B)\right]^+||_2 \right] \max_{\bm{w}}\mathbb{E}\left[||\bm{w}_t||_2\right] \\
&\le \sqrt{n}\mathbb{E}\left[||\bm{w}_{t+1}||_2\right]\mathbb{E}\left[||\bm{w}_t||_2\right] \\
& \le \left( \frac{q(\overline{C}+\overline{D})^2 +2\overline{r}}{2\underline{D}} +q(\overline{C}+\overline{D}) \right)^2 \sqrt{T}\\
& = O(q^2\sqrt{T})
\end{aligned}
\]
where the first inequality utilizes the property of $2$-norm; the second inequality follows from \eqref{ieq:thm1_constr} and the last inequality depends on \Cref{lemma:w_upper_bound}.
Our proof is thus completed.

\section{Proof of \Cref{sec:mirror}}

\subsection{Proof of \Cref{thm:mirror_bound}} \label{appendix:mirror_sto}

First, according to the KKT condition of the Mirror Descent, we have
\begin{equation}
\begin{aligned}
    \gamma_t \langle \bm{g}_t, \bm{w}_{t+1} - \bm{u} \rangle &\le \langle \nabla \psi(w_{t+1}) - \nabla\psi(w_t), u - w_{t+1}\rangle \\
    &=D_\psi(u||w_t) - D_\psi(u||w_{t+1}) - D_\psi(w_{t+1}||w_t)
\end{aligned}
\end{equation}
for any $\bm{u} \in \mathbb{R}^m_{\ge0}$. Adding the term $\langle \bm{g}_t,\bm{w}_t - \bm{w}_{t+1} \rangle$ to both sides and choosing $\gamma_t=\frac{1}{\sqrt{T}}$, we have
\begin{equation}
\begin{aligned}
    \langle \bm{g}_t, \bm{w}_t - \bm{u} \rangle &\le \frac{1}{\sqrt{T}} \left(D_\psi(\bm{u}||\bm{w}_t) - D_\psi(\bm{u}||\bm{w}_{t+1}) - D_\psi(\bm{w}_{t+1}||\bm{w}_t) + \langle \bm{g}_t, \bm{w}_t - \bm{w}_{t+1} \rangle\right) \\
    &\le \frac{1}{\sqrt{T}} \left(D_\psi(\bm{u}||\bm{w}_t) - D_\psi(\bm{u}||\bm{w}_{t+1}) - \frac{\alpha}{2}||\bm{w}_{t+1}-\bm{w}_t||^2 + \langle \bm{g}_t, \bm{w}_t - \bm{w}_{t+1} \rangle\right) \\
    &\le \frac{1}{\sqrt{T}} \left(D_\psi(\bm{u}||\bm{w}_t) - D_\psi(\bm{u}||\bm{w}_{t+1}) - \frac{\alpha}{2}||\bm{w}_{t+1}-\bm{w}_t||^2 + \frac{\alpha}{2}||\bm{w}_{t+1}-\bm{w}_t||^2 + \frac{1}{2\alpha}||\bm{g}_t||_*^2\right) \\
    &= \frac{1}{\sqrt{T}} \left(D_\psi(\bm{u}||\bm{w}_t) - D_\psi(\bm{u}||\bm{w}_{t+1}) + \frac{1}{2\alpha}||\bm{g}_t||_*^2\right)
\end{aligned}
\end{equation}
where the second inequality depends on the $\alpha$-strongly convexity of the potential function $\psi$ and the third inequality holds since
\[
\langle \bm{g}_t, \bm{w}_t - \bm{w}_{t+1} \rangle \le \left\|\bm{g}_t\right\|\cdot \left\|\bm{w}_t - \bm{w}_{t+1}\right\| \le \frac{\alpha}{2}||\bm{w}_{t+1}-\bm{w}_t||^2 + \frac{1}{2\alpha}||\bm{g}_t||_*^2.
\]

Furthermore, similar to the proof of \Cref{thm:simple}, we have:
\[
\text{Reg}_T(\pi_{\text{ALG2}}) \le \sum^T_{t=1}\mathbb{E}[(\bm{d}^\top\Phi-\bm{a}_t^\top\Phi x_t)\bm{w}_t] = \sum^T_{t=1}\mathbb{E}[\langle \bm{g}_t, \bm{w}_t\rangle]
\]
Combing these two inequalities above and choosing $\bm{u} = \bm{0}$, it holds that:
\begin{equation}
\begin{aligned}
    \text{Reg}_T(\pi_{\text{ALG2}}) &\le \sum^T_{t=1}\mathbb{E}[\langle \bm{g}_t, \bm{w}_t\rangle] \\
    &\le \frac{1}{\sqrt{T}} \sum^T_{t=1} \left(D_\psi(\bm{0}||\bm{w}_t) - D_\psi(\bm{0}||\bm{w}_{t+1}) + \frac{1}{2\alpha}||\bm{g}_t||_*^2\right) \\
    &= \frac{1}{\sqrt{T}}  \left(D_\psi(\bm{0}||\bm{w}_1) - D_\psi(\bm{0}||\bm{w}_{T+1}) + \sum^T_{t=1}\frac{1}{2\alpha}||\bm{g}_t||_*^2\right) \\
    &\le \frac{1}{\sqrt{T}}  \sum^T_{t=1}\frac{1}{2\alpha}||\bm{g}_t||_*^2 \\
    &= O(q\sqrt{T})
\end{aligned}
\end{equation}
where $||\bm{g}_t||_*^2\le K||\bm{g}_t||_2^2 =K ||\bm{d}^\top\Phi - \bm{a}_t^\top\Phi x_t||_2^2\le Kq(\overline{C}+\overline{D})^2$ and $K$ is only dependent on $q$ and norm $\left\|\cdot\right\|$.

For constraints violation, define a stopping time $\tau$ for the first time that there exist a resource $j$ such that $\sum_{t=1}^\tau (a_t)_j^\top\Phi_j (x_{t})_j + \overline{C} \ge b_j^\top\Phi_j=(d_j^\top\Phi_j)\cdot T$. Therefore, the constraints violation can be bounded as:
\[
v(\pi_\text{ALG2}) = \mathbb{E}\left[\max_{p\ge0} \left\|\left[p\left(Ax - b\right)\right]^+\right\|_2\right] = \mathbb{E}\left[\max_{w\ge0}\left\|\left[w(\Phi^\top Ax - \Phi^\top b)\right]^+\right\|_2\right] \le \max_{w\ge0} \left\|w\right\|_2\cdot  \mathbb{E}\left[\left\|\left[\Phi^\top Ax - \Phi^\top b\right]^+\right\|_2\right]
\]
where \Cref{assump:sepa} further gives an upper bound to $\left\|\bm{w}\right\|_2$ as $\overline{W}$.

According to the definition of stopping time $\tau$, we know that:
\begin{equation} \label{ieq:constr_viol}
\mathbb{E}\left[\left\|\left(\Phi^\top Ax - \Phi^\top b\right)^+\right\|_2\right] \le \sqrt{\sum_{j=1}^q \left(\overline{C}\cdot\left(T - \tau\right)\right)^2} = \overline{C}\cdot\left(T - \tau\right)\sqrt{q}
\end{equation}

Therefore, we only need to bound the term $T - \tau$. Since we use mirror descent (Step $7$ in \Cref{alg:mirror_sto}) to update $\bm{w}$, according to the KKT condition, it holds that
\[
\nabla\psi(\bm{w}_{t+1}) = \nabla\psi(\bm{w}_{t}) - \gamma_t\bm{g}_t - \bm{v}_{t+1}
\]
where $\bm{v}_{t+1} \in \mathcal{N}_{\mathbb{R}_\ge0}$ and $\mathcal{N}_{\mathbb{R}_\ge0}$ is the normal core of the set $\mathbb{R}_\ge0$ at point $\bm{w}_{t+1}$. Further, the definition of $\mathcal{N}_{\mathbb{R}_\ge0}$:
\[
\mathcal{N}_{\mathbb{R}_\ge0}(x) = \{\bm{v}\in\mathbb{R}^d:v_j = 0 ~~\text{if}~~ x_j > 0; v_j \le 0 ~~\text{if}~~ x_j = 0\}
\]
implies that the condition $\bm{v}_{t+1}$ always holds. Therefore, under \Cref{assump:sepa}, it holds that
\[
\gamma_t\bm{g}_t \ge \nabla\psi(\bm{w}_{t}) - \nabla\psi(\bm{w}_{t+1}) = \nabla\psi_j(\bm{w}_{t}) - \nabla\psi_j(\bm{w}_{t+1})
\]
where $j$ is the index of the first constraint violation happens at time $\tau$. In addition, according to the definition of $\bm{g}_t$, we know that
\begin{equation} \label{eq:def_gt}
\sum_{t=1}^\tau(\bm{g}_t)_j = \sum_{t=1}^\tau (\bm{d}^\top\Phi - \bm{a}_t^\top\Phi x_t)_j \le \sum_{t=1}^\tau (\bm{d}^\top\Phi)_j -  (\bm{d}^\top\Phi)_j\cdot T  +\overline{C}=(\bm{d}^\top\Phi)_j\cdot (\tau-T) + \overline{C}
\end{equation}
Therefore, we can obtain
\begin{equation} \label{ieq:T_tau}
\begin{aligned}
    T - \tau &\le \frac{\overline{C} - \sum_{t=1}^\tau(\bm{g}_t)_j}{(\bm{d}^\top\Phi)_j} \\
    &\le \frac{1}{\sqrt{T}}\cdot\frac{\sqrt{T}\overline{C} + \sum_{t=1}^\tau (\nabla\psi_j(\bm{w}_{t+1}) - \nabla\psi_j(\bm{w}_{t}))}{\underline{D}} \\
    &\le \frac{1}{\sqrt{T}}\cdot\frac{\sqrt{T}\overline{C} + \nabla\psi_j(\overline{W}) - \nabla\psi_j(\bm{0})}{\underline{D}} \\
    &=O(\sqrt{T})
\end{aligned}
\end{equation}
where the first inequality follows from \eqref{eq:def_gt}, the second one depends on the definition of $\tau$ and \Cref{asp:bounded_data}, and the last one utilizes the monotonicity of potential function $\psi_j$.

Combining two inequalities \eqref{ieq:constr_viol} and \eqref{ieq:T_tau}, we can conclude that:
\[
v(\pi_\text{ALG2}) \le \max_{w\ge0} \left\|w\right\|_2\cdot  \mathbb{E}\left[\left\|\left[\Phi^\top Ax - \Phi^\top b\right]^+\right\|_2\right] \le \overline{W} \cdot O(\sqrt{qT}) = O(\sqrt{qT}).
\]

Thus we complete our proof.

\subsection{Proof of \Cref{lemma:optimal_value}}

See proof in of Lemma 1 in \cite{agrawal2014dynamic}.

\subsection{Proof of \Cref{prop:permut}}

First, define SLP$(s,\bm{B}_0)$ as the following LP:
\begin{equation}
\begin{aligned}
\max\ &\sum_{i=1}^sr_ix_i \\ 
    \mathrm{s.t.\ } &\Phi^\top\sum_{i=1}^s a_{ji}x_i \le \Phi^\top(\frac{sb_j}{T} + b_{0j}) \\
    & 0 \le x_i \le 1,~~i=1,...,T 
\end{aligned}
\end{equation}
In addition, we denote $x_i(\bm{w})=\mathbb{I}(r_i > \bm{a}_i\top\Phi \bm{w})$ based on the weight $\bm{w}$, and the optimal value of SLP$(s,b_0)$ as $R^*(s,b_0)$.

We can prove the following two results:

i) When $s \ge \max \{16\overline{C}^2, e^{16\overline{C}^2},e\}$, then the optimal dual solution $\bm{w}^*$ is a feasible solution to SLP$(s,\frac{\log{s}}{\sqrt{s}}\bm{1})$ with high probability no less than $1-\frac{q}{s}$.

ii)
\[
R_s^* \ge R^*(s,\frac{\log{s}}{\sqrt{s}}\bm{1}) - \frac{q\overline{r}\sqrt{s}\log{s}}{\underline{D}}.
\]

%For Part i), this inequality directly comes from \Cref{lemma:optimal_value}.

For the first result, let $\alpha_{ji} = a_{ji}^\top\Phi\mathbb{I}(r_i > a_i^\top\Phi w^*)$ and we have the following inequalities:
\begin{equation} \label{bound_alpha}
\begin{aligned}
    &c_\alpha = \max_{j,i} \alpha_{ji} - \min_{j,i} \alpha_{ji} \le 2 \overline{C} \\
    &\bar{\alpha_j} = \frac{1}{T} \sum^n_{i=1}\alpha_{ji} = \frac{1}{T} \sum^T_{i=1}a_{ji}^\top\Phi x_i(w^*) \le d_j^\top \Phi \\
    &\sigma_j^2 = \frac{1}{T} \sum^T_{i=1}(\alpha_{ji}-\bar{\alpha}_j)^2 \le 4\overline{C}^2
\end{aligned}
\end{equation}
The first and third inequalities utilize the bound of $\bm{a}_i^\top \Phi$ in \Cref{asp:bounded_data}, and the second inequality comes from the feasibility of the optimal solution for problem\eqref{lp:approx_stochastic}.

Therefore, when $k > \max\{16 \overline{C}^2, e^{16 \overline{C}^2}, e\}$, by applying Hoeffding-Bernstein's Inequality, we have
\begin{equation}
\begin{aligned}
    \mathbb{P}\left(\sum^k_{i=1}\alpha_{ji} - kd_j \ge \sqrt{k}\log{k}\right) &\le \mathbb{P}\left(\sum^k_{j=1}\alpha_{ji} - k \bar{\alpha}_j \ge \sqrt{k}\log{k}\right) \\
    &\le \exp \left(-\frac{k \log{k}^2}{8k\overline{C}^2+2\overline{C}\sqrt{k}\log{k}}\right) \\
    &\le \frac{1}{k}
\end{aligned}
\end{equation}
for $i=1,...,q$. The first inequality comes from \eqref{bound_alpha}, the second inequality utilizes the Hoeffding-Bernstein's Inequality and the third inequality is conditioned on $k > \max\{16 \overline{C}^2, e^{16 \overline{C}^2}, e\}$.

Define a event
\[
E_j = \left\{ \sum^s_{i=1}\alpha_{ji} - sd_j^\top\Phi < \sqrt{s}\log{s} \right\}
\]
and $E = \bigcap\limits_{j=1}^{q} E_j$. The above derivation shows that $\mathbb{P}(E_j) \ge 1 - \frac{1}{s}$. Applying union bound property, we have $\mathbb{P}(E) \ge 1 - \frac{q}{s}$, which completes the proof of the first result.

For the second result, similar to the construction of problem \eqref{lp:approx_stochastic}, denote the optimal dual solution to SLP$(s,\frac{\log{s}}{\sqrt{s}}\bm{1})$ as $\tilde{\bm{w}}_s$. Then we have the following inequalities:
\begin{equation}
\begin{aligned}
R^*(s,\frac{\log{s}}{\sqrt{s}}\bm{1}) &= s(\bm{d}+\frac{\log{s}}{\sqrt{s}}\bm{1})^\top \tilde{\bm{w}}^*_s + \sum^s_{i=1}(r_i - a_i^\top \Phi \tilde{\bm{w}}^*_s)^+ \\
&\le \sqrt{s}\log{s}\bm{1}^\top \bm{w}^*_s + s\bm{d}^\top \bm{w}^*_s + \sum^s_{i=1}(r_i - a_i^\top \Phi \bm{w}^*_s)^+ \\
&\le \frac{q\overline{r}\sqrt{s}\log{s}}{\underline{D}} + R_s^*
\end{aligned}
\end{equation}
where the first inequality comes from dual optimality of $\tilde{\bm{w}}_s$ and the second inequality comes from the upper bound of $||\bm{w}^*_s||_\infty$ in following \Cref{lemma:optimal_w_bound} and the strong duality of LP \eqref{lp:approx_dual}. 

\begin{lemma} \label{lemma:optimal_w_bound}
    Any optimal solution $w^*$ of $\min_{w\ge0} f_\Phi(w)$ satisfies:
    \[
    \left\|\bm{w}^*\right\|_2 \le \frac{\sqrt{q}\overline{r}}{\underline{D}}.
    \]
\end{lemma}

\begin{myproof}[Sketch of proof]
To bound $\left\|\bm{w}^*\right\|_2$, we first focus on the $\infty$-norm of $\bm{w}^*$ and then use the inequality $\left\|\bm{w}^*\right\|_2\le\sqrt{q}\left\|\bm{w}^*\right\|_\infty$ to fetch our final result. Assume that $\bm{w}^*$ is the optimal solution for the LP \eqref{lp:approx_dual}, since $\bm{w}=0$ is a feasible solution, it holds that:
\[
f_\Phi(\bm{w}^*) \le f_\Phi(0) = \frac{1}{T}\sum_{t=1}^T(r_t)^+ = \frac{1}{T}\sum_{t=1}^Tr_t\le \overline{r}
\]
since $\overline{r} > 0$ is the upper bound for all $r_t$. Furthermore, it holds that
\[ \label{ieq:w_star_mid}
\bm{d}^\top\Phi \bm{w}^*\le \bm{d}^\top\Phi \bm{w}^* + \frac{1}{T}\sum_{t=1}^T(r_t - \bm{a}_t^\top\Phi \bm{w}^*)^+ = f_\Phi(\bm{w}^*)\le \overline{r}
\]

Next, we will show that for all $j\in[Q]$, $\bm{w}^*_j \le \frac{\overline{r}}{(\bm{d}^\top\Phi)_j}$. If $\bm{w}^*_j > \frac{\overline{r}}{(\bm{d}^\top\Phi)_j}$, since $\bm{d}\ge0,\Phi\ge0,\bm{w}\ge0$, we have
\[
\bm{d}^\top\Phi \bm{w} \ge (\bm{d}^\top\Phi)_j\cdot \bm{w}^*_j > \overline{r}
\]
which contradicts to the inequality \eqref{ieq:w_star_mid}. Therefore, it holds that
\[
\bm{w}^*_j \le \frac{\overline{r}}{(\bm{d}^\top\Phi)_j} \le \frac{\overline{r}}{\underline{D}}
\]

Finally, we can conclude that
\[
\left\|\bm{w}^*\right\|_2 \le \sqrt{q}\left\|\bm{w}^*\right\|_\infty \le \frac{\sqrt{q}\overline{r}}{\underline{D}}
\]
which completes our proof.
\end{myproof}

Therefore, we have
\[
R_s^* \ge R^*(s,\frac{\log{s}}{\sqrt{s}}\bm{1}) - \frac{q\overline{r}\sqrt{s}\log{s}}{\underline{D}}
\]
which completes our proof of the second result.

Finally we can complete the proof with the help of the above two results. Denote $\mathbb{I}_E$ as an indicator function of event $E$, and we have
\begin{equation}
\begin{aligned}
    \frac{1}{s}\mathbb{E}[\mathbb{I}_ER^*_s] &\ge \frac{1}{s}\mathbb{E}[\mathbb{I}_ER^*(s,\frac{\log{s}}{\sqrt{s}}\bm{1})] - \frac{q\overline{r}\sqrt{s}\log{s}}{\underline{D}} \\ 
    &\ge \frac{1}{s}\mathbb{E}[\mathbb{I}_E\sum^s_{i=1}r_ix_i(\bm{w}^*)] - \frac{q\overline{r}\sqrt{s}\log{s}}{\underline{D}}
\end{aligned}
\end{equation}
where the first inequality comes from the part iii) and the second inequality comes from the feasiblity of $\bm{w}^*$ on event $E$. Then we have
\begin{equation}
\begin{aligned}
    \frac{1}{s}\mathbb{E}[R^*_s] &\ge \frac{1}{s}\mathbb{E}[\sum^s_{i=1}r_ix_i(\bm{w}^*)] - \frac{q\overline{r}\sqrt{s}\log{s}}{\underline{D}} - \frac{q\overline{r}}{s} \\
    &= \frac{1}{T}\mathbb{E}[\sum^T_{i=1}r_ix_i(\bm{w}^*)] - \frac{q\overline{r}\sqrt{s}\log{s}}{\underline{D}} - \frac{q\overline{r}}{s} \\
    &\ge \frac{1}{T}R^*_\Phi - \frac{q\overline{r}\sqrt{s}\log{s}}{\underline{D}} - \frac{q\overline{r}}{s} - \frac{q\overline{r}}{T}
\end{aligned}
\end{equation}
where the first inequality comes from our first result and the second inequality comes from \Cref{lemma:optimal_value}. The equality holds since we take expectation for the term $\frac{1}{s}\sum^s_{i=1}r_ix_i(\bm{w}^*)$ and $s$ is an arbitrary positive integer. Thus we complete the proof.

\subsection{Proof of \Cref{thm:permu}}

First, we relate $R_T^*$ to $R^*_\Phi$. Note that the optimal solution $\bm{x}^*$for LP \eqref{lp:primal} is a feasible solution for the approximate LP \eqref{lp:another}, then we have the following inequality:
\[
R_T^* \le R^*_\Phi.
\]
Therefore we have
\begin{equation}
\begin{aligned}
    \text{Reg}_T(\pi_{\text{ALG2}}) &\le R^*_\Phi - \sum^T_{t=1}\mathbb{E}[r_tx_t] \nonumber \\
    &=R^*_\Phi - \sum^T_{t=1}\frac{1}{t}\mathbb{E}[R_t^*] + \sum^T_{t=1}\frac{1}{t}\mathbb{E}[R^*_t] - \sum^n_{t=1}\mathbb{E}[r_tx_t] \nonumber \\
    &=\sum^T_{t=1}(\frac{1}{T}R^*_T - \frac{1}{t}\mathbb{E}[R^*_t]) + \sum^G_{t=1}\mathbb{E}[\frac{1}{T-t+1}\tilde{R}^*_{T-t+1} - r_tx_t] \label{bound:theorem2}
\end{aligned}
\end{equation}
where $\bm{x}_t$ are specified in \Cref{alg:mirror_sto} and $\tilde{R}^*_{T-t+1}$ is defined as the optimal value of the following LP:
\begin{equation}
\begin{aligned}
    \max &\sum^T_{j=t}r_jx_j \\
    \mathrm{s.t.} &\Phi^\top\sum^T_{i=t}a_{ji}x_i \le \Phi^\top(\frac{T-t+1}{T}b_j) \\
    &0 \le x_i \le 1 ~~~~\text{for $i=1,..,m$}
\end{aligned}
\end{equation}
For the first part of \eqref{bound:theorem2}, we can apply \Cref{prop:permut}. Meawhile, the analyses of the second part takes a similar form as the previous stochastic imput model. To be specific, we have
\[
\mathbb{E}[\frac{1}{T-t+1}\tilde{R}^*_{T-t+1} - r_tx_t] \le (\bm{d}^\top\Phi - \bm{a}_t^\top\Phi\mathbb{I}(r_t > \bm{a}_t^\top\Phi \bm{w}_t))^\top \bm{w}_t = \langle \bm{g}_t, \bm{w}_t\rangle.
\]
Similar to the proof of stochastic input model, we have
\begin{equation}
\begin{aligned}
    \langle \bm{g}_t, \bm{w}_t - \bm{u} \rangle 
    &= \frac{1}{\sqrt{n}} \left(D_\psi(\bm{u}||\bm{w}_t) - D_\psi(\bm{u}||\bm{w}_{t+1}) + \frac{1}{2\alpha}||\bm{g}_t||_*^2\right)
\end{aligned}
\end{equation}
for any $u \in R^m_{\ge0}$. Letting $\bm{u}=\bm{0}$ and we have the following inequality:
\[
\begin{aligned}
\sum^T_{t=1}\mathbb{E}[\langle \bm{g}_t, \bm{w}_t - \bm{0} \rangle]
&\le \frac{1}{\sqrt{T}} \sum^T_{t=1} \left(D_\psi(\bm{0}||\bm{w}_t) - D_\psi(\bm{0}||\bm{w}_{t+1}) + \frac{1}{2\alpha}||\bm{g}_t||_*^2\right) \\
&\le \frac{1}{\sqrt{T}}\sum^T_{t=1}\frac{1}{2\alpha}||\bm{g}_t||_*^2\\
&=O(\sqrt{T})
\end{aligned}
\]
where $\bm{g}_t$ are specified in \Cref{alg:mirror_sto}.
Combining two results above, we can have
\begin{equation}
\begin{aligned}
\text{Reg}_T(\pi_{\text{ALG2}}) &\le R_\Phi^* - \mathbb{E}[R_T] \le q\overline{r} + \frac{q\overline{r}\sqrt{T}\log{T}}{\overline{D}} + q\overline{r}\log{T} + \frac{\max\{16\overline{C}^2,e^{16\overline{C}^2},e\}\overline{r}}{n} + \frac{1}{\sqrt{T}}\sum^T_{t=1}\frac{1}{2\alpha}||\bm{g}_t||_*^2 \\
&=O((q+q\log{T})\sqrt{T})
\end{aligned}
\end{equation}

For the constraints violation, the proof is exactly the same as it is in the stochastic input model in \Cref{appendix:mirror_sto}.
Our proof is thus completed.

\section{Proof of \Cref{sec:log_regret}}

\subsection{Proof of \Cref{thm:fast}}
First, we focus on the Step 16 in \Cref{alg:fast} to prove the first result. To begin with, we define the following $\eps$-interval set:
\[
\mathcal{S}_\eps = \{\bm{w} \ge 0:f(\bm{w}) \le f(\bm{w}^*) + \eps\}.
\]
Let $\bm{w}_\eps$ be the closest point in the $\eps$-interval set $\mathcal{S}_\eps$ to a given $\bm{w}$, that is:
\[
\bm{w}_\eps = \arg \min_{\bm{u} \in \mathcal{S}_\eps} \left\|\bm{u} - \bm{w}\right\|_2.
\]
We can observe that when $\bm{w} \notin \mathcal{S}_\eps$, it holds that $f(\bm{w}) = f(\bm{w}^*) + \eps$.
Consider each stage $l$ in \Cref{alg:fast}, and define the following event with $\eps_l = \frac{\eps_0}{2^l}$:
\[
\mathcal{A}_l = \{f(\tilde{\bm{w}}_l) - f(\bm{w}^*) \le \eps_l + \eps\}.
\]
Then we can bound $f(\tilde{\bm{w}}_l) - f(\tilde{\bm{w}}_{l,\eps})$ according to the following lemma:
\begin{lemma} \label{lem:stage}
For each stage $l$ in \Cref{alg:fast}, conditioned on the event $\mathcal{A}_{l-1}$, denote $\tilde{\bm{w}}_l = \frac{1}{J}\sum_{j=1}^J\tilde{\bm{w}}_j^l$. For any $\hat{\delta} \in (0,1)$, it holds that
\begin{equation} \label{ieq:tilde_wl}
f(\tilde{\bm{w}}_l) - f(\tilde{\bm{w}}_{l-1,\eps}) \le \frac{\eta_l(\overline{C}+\overline{D})^2}{2} + \frac{\eps_{l-1}^2}{2\eta_lJ\lambda^2} + \frac{4\eps_{l-1}(\overline{C}+\overline{D})\sqrt{2\log{(1/\hat{\delta})}}}{\lambda\sqrt{J}}
\end{equation}
with a high probability at least $1-\hat{\delta}$.
\end{lemma}

Next, we will prove that conditioned on the event $\mathcal{A}_{l-1}$, the event $\mathcal{A}_l$ happens with a high probability at least $1 - \hat{\delta}$. Note that in \Cref{alg:fast}, we have $\tilde{\bm{w}}^l_1 = \tilde{\bm{w}}_{l-1}$. Then it holds that

Therefore, if we select $\eta_l = \frac{2\eps_l}{3(\overline{C}+\overline{D})^2}=\frac{\eps_{l-1}}{3(\overline{C}+\overline{D})^2}$ and $J \ge \max \{\frac{9(\overline{C}+\overline{D})^2}{\lambda^2}, \frac{1152(\overline{C}+\overline{D})^2\log{(1/\hat{\delta})}}{\lambda^2}\}$, it holds that:
\[
\begin{aligned}
f(\tilde{\bm{w}}_l) - f(\tilde{\bm{w}}_{l-1,\eps}) 
&\le \frac{\eta_l(\overline{C}+\overline{D})^2}{2} + \frac{\eps_{l-1}^2}{2\eta_lJ\lambda^2} + \frac{4\eps_{l-1}(\overline{C}+\overline{D})\sqrt{2\log{(1/\hat{\delta})}}}{\lambda\sqrt{J}} \\
&= \frac{\eps_l}{3} + \frac{3(\overline{C}+\overline{D})^2\eps_{l-1}}{2\lambda^2}\cdot\frac{1}{J} + \frac{4\eps_{l-1}(\overline{C}+\overline{D})\sqrt{2\log{(1/\hat{\delta})}}}{\lambda} \cdot \frac{1}{\sqrt{J}} \\
&\le \frac{\eps_l}{3} + \frac{\eps_l}{3} + \frac{\eps_l}{3} \\
&= \eps_l
\end{aligned}
\]
Finally, we have 
\[
f(\tilde{\bm{w}}_l) - f(\bm{w}^*) = f(\tilde{\bm{w}}_l) - f(\tilde{\bm{w}}_{l,\eps}) + f(\tilde{\bm{w}}_{l,\eps}) - f(\bm{w}^*) \le \eps_l + \eps
\]
which means that the event $\mathcal{A}_l$ happens with a high probability at least $1 - \hat{\delta}$ conditioned on event $\mathcal{A}_{l-1}$. 
Define the “stage-$l$ success event” $B_l$ to be the event that the high-probability bound in \Cref{lem:stage} holds at stage $l$, and we can observe that conditioned on event $A_{l-1}$,
\[
\mathbb{P}(B_l | \text{history}) \ge 1 - \hat{\delta},~~A_{l-1}\bigcap B_l = A_l.
\]
Therefore, as long as all events $B_l$ for $l=1,..,L$ happen, that is the event $B := \bigcap\limits^L_{l=1} B_l$ happens, we have
\[
A_1 \Rightarrow A_2 \Rightarrow ... \Rightarrow A_L, ~~\mathcal{A}_L = \{f(\tilde{\bm{w}}_L) - f(\bm{w}^*) \le \eps_L + \eps\}.
\]
Overall, we can conclude that with high probability $\mathbb{P}(B) \ge (1 - \hat{\delta})^L \ge 1 - \delta$, it holds that
\[
f(\tilde{\bm{w}}_L) - f(\bm{w}^*) \le \eps_L + \eps \le 2\eps.
\]
Finally, with the help of \Cref{asp:holder} and selecting $\lambda=1$ and $\eps = \frac{1}{2T}$, we can obtain that
\[
\left\|\tilde{\bm{w}}_L - \bm{w}^*\right\| \le f(\tilde{\bm{w}}_L) - f(\bm{w}^*) \le \frac{1}{T}
\]
which completes our proof of the first result.

For our second result, since \Cref{alg:fast} is dependent on a virtual decision $\tilde{x}_t$, we know that
\begin{equation} \label{ieq:two_bounds}
\begin{aligned}
\text{Reg}_{T_{\text{fast}}}(\pi_{\text{ALG3}}) &\le \mathbb{E}\left[R_\Phi^* - \sum_{t=1}^{T_{\text{fast}}} r_tx_t\right] \\
&\le {T_{\text{fast}}}\cdot f(\bm{w}^*) - \mathbb{E}\left[\sum_{t=1}^{T_{\text{fast}}} r_t\tilde{x}_t\right] + \mathbb{E}\left[\sum_{t=1}^{T_{\text{fast}}} r_t\tilde{x}_t\right] - \mathbb{E}\left[\sum_{t=1}^{T_{\text{fast}}} r_tx_t\right] \\
&\le \sum_{t=1}^{T_{\text{fast}}} f(\bm{w}_t) - \mathbb{E}\left[\sum_{t=1}^{T_{\text{fast}}} r_t\tilde{x}_t\right] + \mathbb{E}\left[\sum_{t=1}^{T_{\text{fast}}} r_t\tilde{x}_t\right] - \mathbb{E}\left[\sum_{t=1}^{T_{\text{fast}}} r_tx_t\right] \\
&\le \underbrace{\sum_{t=1}^{T_{\text{fast}}}\mathbb{E}\left[\left(\bm{d}^\top\Phi - \bm{a}_t^\top \Phi\tilde{x}_t\right)\bm{w}_t\right]}_{\textbf{BOUND I}} + \underbrace{\sum_{t=1}^{T_{\text{fast}}}\mathbb{E}\left[r_t\left(\tilde{x}_t - x_t\right)\right]}_{\textbf{BOUND II}}
\end{aligned}
\end{equation}
We then separately bound these two parts.

For \textbf{BOUND I}, according to the update rule of $\bm{w}_t$ in \Cref{alg:fast}, which is the Step $15$, we know that
\begin{equation} \label{ieq:fast_update_w}
\begin{aligned}
\left\|\bm{w}_{t+1}\right\|_2 &\le \left\|\bm{w}_t\right\|_2 - 2\gamma_t\left(\bm{d}^\top\Phi - \bm{a}_t^\top \Phi\tilde{x}_t\right)\bm{w}_t + \gamma_T^2\left\|\bm{d}^\top\Phi - \bm{a}_t^\top \Phi\tilde{x}_t\right\|^2 \\
&\le \left\|\bm{w}_t\right\|_2 - \frac{2}{\log{T}}\left(\bm{d}^\top\Phi - \bm{a}_t^\top \Phi\tilde{x}_t\right)\bm{w}_t + \frac{q(\overline{C}+\overline{D})^2}{\log^2{T}}^2 
\end{aligned}
\end{equation}
Therefore, it holds that
\[
\begin{aligned}
\textbf{BOUND I} &= \sum_{t=1}^{T_{\text{fast}}}\mathbb{E}\left[\left(\bm{d}^\top\Phi - \bm{a}_t^\top \Phi\tilde{x}_t\right)\bm{w}_t\right] \\
&\le \mathbb{E}\left[\frac{\log{T}}{2}\sum_{t=1}^{T_{\text{fast}}} \left(\bm{w}_t - \bm{w}_{t+1}\right) + T_{\text{fast}}\cdot\frac{q(\overline{C}+\overline{D})^2}{2\log{T}}\right]\\
&= \mathbb{E}\left[\frac{\log{T}}{2}\bm{w}_0+T_{\text{fast}}\cdot\frac{q(\overline{C}+\overline{D})^2}{2\log{T}}\right]\\
&=O(q\log{T}).
\end{aligned}
\]
where the first inequality comes from the inequality \eqref{ieq:fast_update_w} and the second equality depends on the fact that $\bm{w}_0 = 0$ and $T_{\text{fast}} = O(\log^2{T})$.

For \textbf{BOUND II}, first it holds that:
\[
r_t\left(\tilde{x}_t - x_t\right) \le r_t\tilde{x_t}\cdot\mathbb{I}\{\exists j : B_{t,j} < (\bm{a}_t^\top\Phi\tilde{x}_t)_j\}
\]
where $B_{t,j}$ denotes the $j$-th component of $B_t$ and $\mathbb{I}$ denotes the indicator function. In addition, we can observe that:
\[
\mathbb{I}\{\exists j : B_{t,j} < \bm{a}_t^\top\Phi\tilde{x}_t\} < \sum_{j=1}^q \mathbb{I}\left\{\sum_{i=1}^t(\bm{a}_i^\top\Phi\tilde{x}_i)_j > B_{j}\right\}
\]
Therefore, we have:
\[
\begin{aligned}
\sum_{t=1}^{T_{\text{fast}}}r_t(\tilde{x}_t - x_t) &\le \sum_{t=1}^{T_{\text{fast}}}r_t\tilde{x}_t\cdot\mathbb{I}\{\exists j : B_{t,j} < (\bm{a}_t^\top\Phi\tilde{x}_t)_j\} \\
&\le\sum_{t=1}^{T_{\text{fast}}}\overline{r}\tilde{x}_t\cdot\sum_{j=1}^q\mathbb{I}\left\{\sum_{i=1}^t(\bm{a}_i^\top\Phi\tilde{x}_i)_j > B_{j}\right\} \\
&\le \overline{r}\sum_{j=1}^q\sum_{t=1}^{T_{\text{fast}}}\mathbb{I}\left\{\sum_{i=1}^t(\bm{a}_i^\top\Phi\tilde{x}_i)_j > B_{j}\right\}
\end{aligned}
\]

Define that $S_j(t) = \sum_{i=1}^t (\bm{a}_i^\top\Phi x_i)_j$ and $\tau = \inf\{\tau:S_j(t)>(\bm{b}^\top\Phi)_j\}$. We consider the following two conditions:

i) If $\tau = \infty$, then for any $j \in [Q]$, $\sum_{i=1}^t (\bm{a}_i^\top\Phi \tilde{x}_i)_j \le (\bm{b}^\top\Phi)_j$ will always satisfy. That is to say, the constraints will not be violated. Therefore the term $\mathbb{I}\left\{\sum_{i=1}^t(\bm{a}_i^\top\Phi\tilde{x}_i)_j > B_{j}\right\}$ will be $0$.

ii) If $\tau < \infty$, since $\bm{a}_t^\top\Phi$ has an upper bound $\overline{C}$, it must hold that $S_j(\tau-1) >(\bm{b}^\top\Phi)_j - \overline{C}$. Then we have:
\[
S_j(\tau) - S_J(\tau-1) \le S_j(\tau) - (\bm{b}^\top\Phi)_j + \overline{C}
\]

Combining two conditions above, we can conclude that:
\[
\begin{aligned}
\sum_{t=1}^{T_{\text{fast}}}r_t(\tilde{x}_t - x_t)
&\le \overline{r}\sum_{j=1}^q\sum_{t=1}^{T_{\text{fast}}}\mathbb{I}\left\{\sum_{i=1}^t(\bm{a}_i^\top\Phi\tilde{x}_i)_j > B_{j}\right\} \\
&\le \overline{r}\sum_{j=1}^q\left[\sum_{t=1}^{T_{\text{fast}}}(\bm{a}_t^\top\Phi\tilde{x}_t)_j  - (\bm{b}^\top\Phi)_j + \overline{C}\right]^+
\end{aligned}
\]

Revisiting the Step $15$ in \Cref{alg:fast}, we know that:
\[
\sum_{t=1}^{T_{\text{fast}}}\bm{a}_t^\top\Phi\tilde{x}_t - \bm{d}^\top\Phi \le \frac{1}{\gamma_t}\sum_{t=1}^{T_{\text{fast}}}\bm{w}_{t+1} - \bm{w}_t = \frac{1}{\gamma_t}\cdot\bm{w}_{T_{\text{fast}}+1}
\]
Therefore, we have:
\begin{equation} 
\begin{aligned} 
\textbf{BOUND II} &\le \overline{r}\sum_{j=1}^q\left[\sum_{t=1}^T(\bm{a}_t^\top\Phi\tilde{x}_t)_j  - (\bm{b}^\top\Phi)_j + \overline{C}\right]^+ \\
&\le \overline{r}q\left(\frac{1}{\gamma_t}\cdot\bm{w}_{T_{\text{fast}}+1} + \overline{C}\right) \\
&= O(q\log{T})
\end{aligned}
\end{equation}
since $\bm{w}_{T_{\text{fast}}+1}$ can be similarly bounded by \Cref{lemma:w_upper_bound} and $\gamma_t = \frac{1}{\log{T}}$ .

Combining \textbf{BOUND I} and \textbf{BOUND II} above, we can obtain our final results:
\[
\begin{aligned}
\text{Reg}_{T_{\text{fast}}}(\pi_{\text{ALG3}}) \le \textbf{BOUND I} + \textbf{BOUND II} \le O(q\log{T}).
\end{aligned}
\]
if we select $\gamma_t \le \frac{1}{log{T}}$ for $t=1,...,T_{\text{fast}}$. Overall, our proof is thus completed.

\subsection{Proof of \Cref{lem:stage}}
We focus on the stage $l$. Let $\hat{\bm{g}}_t = \bm{a}_t^\top\Phi \tilde{x}_t - \bm{d}^\top\Phi$, and we know that $\hat{\bm{g}}_j$ is an unbiased estimator for a subgradient $\bm{g}_j$ of $f(\bm{w})$ at time step $j$, since our input data $(r, \bm{a})$ are $i.i.d.$ sampled. Denote $\mathcal{F}_{j-1}$ as the history information from $\tau = 1,...,j-1$, and it holds that
\[
\mathbb{E}\left[\hat{\bm{g}}_j|\mathcal{F}_{j-1}\right] = \bm{g}_j.
\]
Therefore, the following $X_j$ behaves as a martingale difference sequence:
\[
X_j = \bm{g}_j(\tilde{\bm{w}}_j^{l} -\tilde{\bm{w}}_{{l-1},\eps}) - \hat{\bm{g}}_j(\tilde{\bm{w}}_j^{l} -\tilde{\bm{w}}_{{l-1},\eps})
\]
Then, it holds that
\[
\begin{aligned}
\left|X_j\right| &\le \left(\left\|\bm{g}_j\right\| + \left\|\hat{\bm{g}}_j\right\|\right)\left\|\tilde{\bm{w}}_j^{l} -\tilde{\bm{w}}_{{l-1},\eps}\right\| \le 2(\overline{C}+\overline{D})\cdot\left\|\tilde{\bm{w}}_j^{l} -\tilde{\bm{w}}_{{l-1},\eps}\right\| \\&\le 2(\overline{C}+\overline{D})\left(\left\|\tilde{\bm{w}}_j^{l} -\tilde{\bm{w}}_{l-1}\right\| + \left\|\tilde{\bm{w}}_{l-1} -\tilde{\bm{w}}_{{l-1},\eps}\right\|\right)
\end{aligned}
\]
where $\left\|\tilde{\bm{w}}_j^{l} -\tilde{\bm{w}}_{l-1}\right\|$ can be bounded by $V_l = \frac{\eps_{l-1}}{\lambda}$ since $\tilde{\bm{w}}_j^{l} \in \mathcal{B}_{(\tilde{\bm{w}}_{l-1}, V_l)}$.
Next, we will prove that $\left\|\tilde{\bm{w}}_{l-1} -\tilde{\bm{w}}_{{l-1},\eps}\right\| \le V_l$.
First, if $\tilde{\bm{w}}_{l-1} \in \mathcal{S}_\eps$, then according to the definition of $\tilde{\bm{w}}_{{l-1},\eps}$, $\left\|\tilde{\bm{w}}_{l-1} -\tilde{\bm{w}}_{{l-1},\eps}\right\| = 0$. Then we consider $\tilde{\bm{w}}_{l-1} \notin \mathcal{S}_\eps$. Since the definition of $\tilde{\bm{w}}_{{l-1},\eps}$ is essentially an optimization problem, applying KKT conditions, it holds that
\begin{subequations}
\begin{equation}
\begin{cases}
    &\tilde{\bm{w}}_{{l-1},\eps} - \tilde{\bm{w}}_{l-1} - \alpha + \xi \bm{g} = 0\label{eq:KKT} \\
    &\alpha_i(\tilde{\bm{w}}_{{l-1},\eps})_i \ge 0 ~~ \forall i, ~~\xi (f(\tilde{\bm{w}}_{{l-1},\eps}) - f(\bm{w}^*) - \eps) = 0
\end{cases}
\end{equation}
\end{subequations}

where $\alpha$ and $\xi$ are Lagrangian multipliers.
In addition, revisiting the definition of $\mathcal{S}_\eps$, there must have $f(\tilde{\bm{w}}_{{l-1},\eps}) = f(\bm{w}^*) + \eps$ when $\tilde{\bm{w}}_{l-1} \notin \mathcal{S}_\eps$, which means $\xi > 0$. Then we can rewrite \eqref{eq:KKT} as
\begin{equation} \label{eq:KKT_re}
\tilde{\bm{w}}_{{l-1},\eps} - \tilde{\bm{w}}_{l-1} = \xi(\bm{g} - \frac{\alpha}{\xi})
\end{equation}
Define $s = - \frac{\alpha}{\xi}$ and $\tilde{v} = \bm{g} + s$. Note that for the domain $\mathbb{R}_{\ge0}$, its normal cone is defined as
\[
\mathcal{N}_{\mathbb{R}_{\ge0}}(x) = \{s \in \mathbb{R}: s = 0 ~\text{if}~ x > 0;~s\le0 ~\text{if}~ x = 0\}.
\]
Therefore, if $(\tilde{\bm{w}}_{l,\eps})_i = 0$, then $\alpha_i \le 0$ and $s_i \le 0$; if $(\tilde{\bm{w}}_{l,\eps})_i > 0$, then $\alpha_i = 0$ and $s_i = 0$, which means $s \in \mathcal{N}_{\mathbb{R}_{\ge0}}(\tilde{\bm{w}}_{l,\eps})$. Then it holds that
\[
\tilde{v} = \bm{g} + s \in \partial f_\Phi(\tilde{\bm{w}}_{{l-1},\eps}) + \mathcal{N}_{\mathbb{R}_{\ge0}}(\tilde{\bm{w}}_{{l-1},\eps}) = \partial (f_\Phi + \mathbb{I}_{\mathbb{R}_{\ge0}})(\tilde{\bm{w}}_{{l-1},\eps})
\]
where the indicator function $\mathbb{I}$ is defined as following:
\[
\mathbb{I}_{\mathbb{R}_{\ge0}}(x) = 
\begin{cases}
    0, ~&x\in\mathbb{R}_{\ge0} \\
    \infty,~&x\notin\mathbb{R}_{\ge0}
\end{cases}
\]
Let $\tilde{F} = f_\Phi + \mathbb{I}_{\mathbb{R}_{\ge0}}$, and $\tilde{v}$ is an subgradient of $\tilde{F}$. Then it holds that
\begin{equation} \label{ieq:subg_v}
\tilde{F}(\tilde{\bm{w}}_{l-1}) - \tilde{F}(\tilde{\bm{w}}_{{l-1},\eps}) \ge \tilde{v}^\top(\tilde{\bm{w}}_{l-1} - \tilde{\bm{w}}_{{l-1},\eps}) = \left\|\tilde{v}\right\|_2\left\|\tilde{\bm{w}}_{l-1} - \tilde{\bm{w}}_{{l-1},\eps}\right\|_2
\end{equation}
where the equality holds since the equality \eqref{eq:KKT_re} indicates that $\tilde{\bm{w}}_{l-1} - \tilde{\bm{w}}_{l-1,\eps}$ shares the same direction with $\tilde{v} = \bm{g} - \frac{\alpha}{\xi}$. Furthermore, since $\bm{w}^* , \tilde{\bm{w}}_{{l-1},\eps} \in \mathbb{R}_{\ge0}$ and event $A_{l-1}$ happens, it holds that
\[
f_\Phi(\bm{w}^*) = \tilde{F}(\bm{w}^*) \ge \tilde{F}(\tilde{\bm{w}}_{{l-1},\eps}) + \tilde{v}^\top(\bm{w}^*-\tilde{\bm{w}}_{{l-1},\eps}) = f_\Phi(\bm{w}^*) + \eps_{l-1} + \tilde{v}^\top(\tilde{\bm{w}}_{{l-1},\eps}-\bm{w}^*)
\]
where . This implies that
\begin{equation} \label{ieq:eps_v}
\eps_{l-1} \le \tilde{v}^\top(\bm{w}^*-\tilde{\bm{w}}_{{l-1},\eps}) \le \left\|\tilde{v}\right\|_2\left\|\bm{w}^*-\tilde{\bm{w}}_{{l-1},\eps}\right\|_2= \left\|\tilde{v}\right\|_2\cdot dist(\tilde{\bm{w}}_{{l-1},\eps},\mathcal{W}^*)
\end{equation}
where the second inequality follows from the Cauchy Inequality. Combining inequalities \eqref{ieq:subg_v} and \eqref{ieq:eps_v}, we know that
\begin{equation}
\left\|\tilde{\bm{w}}_{l-1} - \tilde{\bm{w}}_{l-1,\eps}\right\|_2 \le \frac{\tilde{F}(\tilde{\bm{w}}_{l-1}) - \tilde{F}(\tilde{\bm{w}}_{l-1,\eps})}{\tilde{v}} \le \frac{dist(\tilde{\bm{w}}_{l-1,\eps},\mathcal{W}^*)\cdot(\tilde{F}(\tilde{\bm{w}}_{l-1}) - \tilde{F}(\tilde{\bm{w}}_{{l-1},\eps}))}{\eps_{l-1}}
\end{equation}

Then, under \Cref{asp:holder}, we know that the distance of a point away from the optimal set is upper bounded by their objective values, which means that
\begin{equation}
\begin{aligned}
\left\|\tilde{\bm{w}}_{l-1} - \tilde{\bm{w}}_{l-1,\eps}\right\|_2 &\le \frac{dist(\tilde{\bm{w}}_{l-1,\eps}, \mathcal{W}^*)}{\eps}\cdot\left(\tilde{F}(\tilde{\bm{w}}_{l-1}) - \tilde{F}(\tilde{\bm{w}}_{{l-1},\eps}))\right) \\
&\le \frac{f(\tilde{\bm{w}}_{l-1,\eps}) - f(\bm{w}^*)}{\lambda\eps} \cdot \left(f(\tilde{\bm{w}}_{l-1}) - f(\bm{w}^*) + f(\bm{w}^*) - f(\tilde{\bm{w}}_{l-1,\eps})\right) \\
&\le \frac{\eps\cdot(\eps_{l-1}+\eps - \eps)}{\lambda\eps} \\
&=\frac{\eps_{l-1}}{\lambda}
\end{aligned}
\end{equation}

Finally, the martingale difference sequence $X_j$ is upper bounded by:
\[
\left|X_j\right| \le 2(\overline{C}+\overline{D})\left(\left\|\tilde{\bm{w}}_j^{l} -\tilde{\bm{w}}_{l-1}\right\| + \left\|\tilde{\bm{w}}_{l-1} -\tilde{\bm{w}}_{{l-1},\eps}\right\|\right) \le \frac{4(\overline{C}+\overline{D})\eps_{l-1}}{\lambda}
\]

Applying Azuma–Hoeffding Inequality, with probability at least $1-\delta$, it holds that
\begin{equation} \label{ieq:azuma}
\frac{1}{J}\sum_{j=1}^J \bm{g}_j(\tilde{\bm{w}}_j^{l-1} -\tilde{\bm{w}}_{l-1,\eps}) - \frac{1}{J}\sum_{j=1}^J \hat{\bm{g}}_j(\tilde{\bm{w}}_j^{l-1} -\tilde{\bm{w}}_{l-1,\eps}) = \frac{1}{J}\sum^J_{j=1} X_j \le \frac{4\eps_{l-1}(\overline{C}+\overline{D})\sqrt{2\log{(1/\delta)J}}}{\lambda}
\end{equation}

Denote $\hat{\bm{w}}_{j+1}^l = \tilde{\bm{w}}_j^l - \eta_j\hat{\bm{g}_j}$. According to the property of projection, we know that,
\[
\begin{aligned}
\left\|\tilde{\bm{w}}_{j+1}^l - \tilde{\bm{w}}_{l-1,\eps}\right\|^2 \le\left\|\hat{\bm{w}}_{j+1}^l - \tilde{\bm{w}}_{l-1,\eps}\right\|^2 = 
\left\|\tilde{\bm{w}}_j^l - \tilde{\bm{w}}_{l-1,\eps}\right\|^2 - 2\eta_j\hat{\bm{g}_j}(\tilde{\bm{w}}_j^l - \tilde{\bm{w}}_{l-1,\eps}) + \left\|\eta_t\hat{\bm{g}_j}\right\|^2
\end{aligned}
\]
That is,
\begin{equation} \label{ieq:hat_g_bound}
\begin{aligned}
\sum_{j=1}^J \hat{\bm{g}}_j(\tilde{\bm{w}}_j^l - \tilde{\bm{w}}_{l-1,\eps}) &\le \sum_{j=1}^J \frac{\left\|\tilde{\bm{w}}_{j+1}^l - \tilde{\bm{w}}_{l-1,\eps}\right\|^2 - \left\|\tilde{\bm{w}}_{j}^l - \tilde{\bm{w}}_{l-1,\eps}\right\|^2}{2\eta_l} + \frac{1}{2\eta_l}\sum_{j=1}^J\left\|\eta_l\hat{\bm{g}_j}\right\|^2 \\
&\le \frac{\left\|\tilde{\bm{w}}_{1}^l - \tilde{\bm{w}}_{l-1,\eps}\right\|^2}{2\eta_l} + \frac{ \eta_lJ (\overline{C}+\overline{D})^2}{2} \\
&= \frac{\left\|\tilde{\bm{w}}_{l-1} - \tilde{\bm{w}}_{l-1,\eps}\right\|^2}{2\eta_l} + \frac{ \eta_lJ (\overline{C}+\overline{D})^2}{2} \\
&\le \frac{V_l^2}{2\eta_l} + \frac{ J (\overline{C}+\overline{D})^2}{2\eta_l} = \frac{\eps_{l-1}^2}{2\eta_l\lambda^2} + \frac{ \eta_lJ (\overline{C}+\overline{D})^2}{2}
\end{aligned}
\end{equation}
where the equality stems from the Step $5$ in \Cref{alg:fast}.
Note that $\bm{g}_j$ is a subgradient for $\tilde{\bm{w}}_j^{l}$ at time step $j$, we can obtain
\begin{equation} \label{ieq:subg}
f_\Phi(\tilde{\bm{w}}_j^{l}) - f_\Phi(\tilde{\bm{w}}_{{l-1},\eps}) \le \bm{g}_j^\top(\tilde{\bm{w}}_j^{l} - \tilde{\bm{w}}_{{l-1},\eps})
\end{equation}
Combining inequalities \eqref{ieq:azuma},\eqref{ieq:hat_g_bound} and \eqref{ieq:subg}, it holds that
\[
\begin{aligned}
\frac{1}{J}\sum_{j=1}^J f_\Phi(\tilde{\bm{w}}_j^{l}) - f_\Phi(\tilde{\bm{w}}_{l-1,\eps}) 
&\le \frac{1}{J}\sum_{j=1}^J \bm{g}_j^\top(\tilde{\bm{w}}_j^{l-1} - \tilde{\bm{w}}_{l-1,\eps}) \\
&\le \frac{1}{J}\left(\sum_{j=1}^J \hat{\bm{g}}_j(\tilde{\bm{w}}_j^{l-1} -\tilde{\bm{w}}_{l-1,\eps}) + \frac{4\eps_{l-1}(\overline{C}+\overline{D})\sqrt{2\log{(1/\delta)J}}}{\lambda}\right) \\
&\le \frac{\eta_l(\overline{C}+\overline{D})^2}{2} + \frac{\eps_{l-1}^2}{2\eta_lJ\lambda^2} + \frac{4\eps_{l-1}(\overline{C}+\overline{D})\sqrt{2\log{(1/\delta)}}}{\lambda\sqrt{J}}
\end{aligned}
\]

Finally, since $f_\Phi(\bm{w})$ is a convex function, it holds that
\[
\begin{aligned}
f(\tilde{\bm{w}}_l) - f(\tilde{\bm{w}}_{l-1,\eps}) &= f(\frac{1}{J}\sum_{j=1}^J\tilde{\bm{w}}^{l}_j) - f(\tilde{\bm{w}}_{l-1,\eps}) \\
&\le \frac{1}{J}\sum_{j=1}^J f_\Phi(\tilde{\bm{w}}_j^{l}) - f_\Phi(\tilde{\bm{w}}_{l-1,\eps}) \\&\le  \frac{\eta_l(\overline{C}+\overline{D})^2}{2} + \frac{\eps_{l-1}^2}{2\eta_lJ\lambda^2} + \frac{4\eps_{l-1}(\overline{C}+\overline{D})\sqrt{2\log{(1/\delta)}}}{\lambda\sqrt{J}}
\end{aligned}
\]
which completes our proof.

\subsection{Proof of \Cref{thm:const_reg_bound}} \label{appendix:const_reg}

First, since \Cref{alg:const_reg} is dependent on a virtual decision $\tilde{x}_t$, we know that
\[
\begin{aligned}
\text{Reg}_{T_{\text{refine}}}(\pi_{\text{ALG4}}) &\le \mathbb{E}\left[R_\Phi^* - \sum_{t=1}^{T_{\text{refine}}} r_tx_t\right] \\
&\le {T_{\text{refine}}}\cdot f(\bm{w}^*) - \mathbb{E}\left[\sum_{t=1}^{T_{\text{refine}}} r_t\tilde{x}_t\right] + \mathbb{E}\left[\sum_{t=1}^{T_{\text{refine}}} r_t\tilde{x}_t\right] - \mathbb{E}\left[\sum_{t=1}^{T_{\text{refine}}} r_tx_t\right] \\
&\le \sum_{t=1}^{T_{\text{refine}}} f(\bm{w}_t) - \mathbb{E}\left[\sum_{t=1}^{T_{\text{refine}}} r_t\tilde{x}_t\right] + \mathbb{E}\left[\sum_{t=1}^{T_{\text{refine}}} r_t\tilde{x}_t\right] - \mathbb{E}\left[\sum_{t=1}^{T_{\text{refine}}} r_tx_t\right] \\
&\le \underbrace{\sum_{t=1}^{T_{\text{refine}}}\mathbb{E}\left[\left(\bm{d}^\top\Phi - \bm{a}_t^\top \Phi\tilde{x}_t\right)\bm{w}_t\right]}_{\textbf{BOUND I}} + \underbrace{\sum_{t=1}^{T_{\text{refine}}}\mathbb{E}\left[r_t\left(\tilde{x}_t - x_t\right)\right]}_{\textbf{BOUND II}}
\end{aligned}
\]
We then separately bound these two parts.

For \textbf{BOUND I}, similar to proof of \Cref{thm:simple}, we have:
\[
\begin{aligned}
\left\|\bm{w}_{t+1}\right\|_2^2 &\le \left\|\bm{w}_t + \gamma_t\left(\bm{d}^\top - \bm{a}_t^\top\Phi \tilde{x}_t\right)\right\|_2^2 \\
&= \left\|\bm{w}_t\right\|^2_2 + \gamma_t^2\left(\bm{d}^\top\Phi - \bm{a}_t^\top\Phi\tilde{x}_t\right)^2 -2\gamma_t\left(\bm{d}^\top\Phi - \bm{a}_t^\top\Phi\tilde{x}_t\right)\bm{w}_t
\end{aligned}
\]
Therefore, it holds that:
\[
\begin{aligned}
    \textbf{BOUND I} &\le \sum_{t=1}^{T_{\text{refine}}}\frac{1}{2\gamma_t}\left(\left\|\bm{w}_t\right\|^2_2 - \left\|\bm{w}_{t+1}\right\|^2_2\right) + \frac{\gamma_t}{2}\left(\bm{d}^\top\Phi - \bm{a}_t^\top\Phi\tilde{x}_t\right)^2 \\
    &\le \frac{1}{2\gamma_t}\left(\left\|\hat{\bm{w}}^*\right\|^2_2 - \left\|\bm{w}_{{T_{\text{refine}}}+1}\right\|^2_2\right) + \frac{T\gamma_tq}{2}\left(\overline{C}+\overline{D}\right)^2 \\
    &= \frac{1}{2\gamma_t}\left(\langle\hat{\bm{w}}^* +\bm{w}_{{T_{\text{refine}}}+1},\hat{\bm{w}}^*-\bm{w}_{{T_{\text{refine}}}+1} \rangle\right) + \frac{T\gamma_tq}{2}\left(\overline{C}+\overline{D}\right)^2 \\
    &\le \frac{1}{2\gamma_t}\left(\left\|\hat{\bm{w}}^* + \bm{w}_{{T_{\text{refine}}}+1}\right\|_2 \cdot \left\|\hat{\bm{w}}^* -\bm{w}_{{T_{\text{refine}}}+1}\right\|_2\right) + \frac{T\gamma_tq}{2}\left(\overline{C}+\overline{D}\right)^2 \\
    &\le \frac{1}{2\gamma_t}\left(\left(\left\|\hat{\bm{w}}^*\right\|_2 + \left\|\bm{w}_{{T_{\text{refine}}}+1}\right\|_2\right) \cdot \left\|\hat{\bm{w}}^* -\bm{w}_{{T_{\text{refine}}}+1}\right\|_2\right) + \frac{T\gamma_tq}{2}\left(\overline{C}+\overline{D}\right)^2 \\
    &\le \frac{1}{2\gamma_t}\left(\left(\left\|\hat{\bm{w}}^*\right\|_2 + \left\|\bm{w}_{{T_{\text{refine}}}+1}\right\|_2\right) \cdot \left(\left\|\hat{\bm{w}}^* -\bm{w}^*\right\|_2 + \left\|\bm{w}^* -\bm{w}_{{T_{\text{refine}}}+1}\right\|_2\right)\right) + \frac{T\gamma_tq}{2}\left(\overline{C}+\overline{D}\right)^2
\end{aligned}
\]
where the second inequality follows from \Cref{asp:bounded_data} and the fact that $\hat{\bm{w}}^*$ is the start point of $\bm{w}$, the third one utilizes the Cauchy–Schwarz Inequality and the forth and last one is obtained from the triangle inequality for the Euclidean norm.
Note that The upper bound of $\left\|\hat{\bm{w}}^*\right\|_2$ and $\left\|\bm{w}_{{T_{\text{refine}}}+1}\right\|_2$ can be similarly given by \Cref{lemma:w_upper_bound} and the upper bound of $\left\|\hat{\bm{w}}^* -\bm{w}^*\right\|_2$ is given in \Cref{thm:fast}.
Therefore we focus on the upper bound of $\left\|\bm{w}^* -\bm{w}_{{T_{\text{refine}}}+1}\right\|_2$, which is presented in the following \Cref{lemma:w_diff_bound}. 

\begin{lemma} \label{lemma:w_diff_bound}
    Suppose that $\bm{d}$ has a GPG of $\eps$, $\bm{w}^*$ is the optimal solution for \eqref{lp:approx_stochastic} and $\bm{w}_{T+1}$ is the output of \Cref{alg:const_reg}. It holds that:
    \[
    \mathbb{E}\left[\left\|\bm{w}^* - \bm{w}_{T+1}\right\|_2\right] \le \gamma_t\left(2Z + \frac{\eps}{2} + \frac{2Z^2}{\eps}\right)
    \]
    where $Z = \sqrt{q}\left(\overline{C} + \overline{D}\right)$.
\end{lemma}

\begin{myproof}
First, for any $t\ge1$, we have:
\[
\begin{aligned}
    \left|\left\|\bm{w}^* - \bm{w}_{t+1}\right\|_2 - \left\|\bm{w}^* - \bm{w}_t\right\|_2\right| &\le \left\|\bm{w}^* - \bm{w}_{t+1} - \left(\bm{w}^* - \bm{w}_t\right)\right\|_2 \\
    &= \left\|\bm{w}_t - \left(\bm{w}_t + \gamma_t\left(\bm{a}_t^\top\Phi\tilde{x}_t - \bm{d}^\top\Phi\right)\right)^+\right\|_2 \\
    &\le \left\|\bm{w}_t - \left(\bm{w}_t + \gamma_t\left(\bm{a}_t^\top\Phi\tilde{x}_t - \bm{d}^\top\Phi\right)\right)\right\|_2 \\
    &\le \gamma_tZ
\end{aligned}
\]
where the first inequality stems from the triangle inequality.
Now we illustrate that the term $\left\|\bm{w}^* - \bm{w}_t\right\|_2$ has negative drift property, which plays a key role in bounding $\left\|\bm{w}^* - \bm{w}_t\right\|_2$. Similar results have been developed in previous literature \citep{huang2009delay,he2025online} and here we conclude them in the following lemma:

\begin{lemma} \label{lemma:nega_drift}
Denote $\bm{w}_t$ be the output of \Cref{alg:const_reg}, for any $t\ge1$, conditioned on any fixed $\bm{w}_t$, we have
\[
\mathbb{E}\left[\left\|\bm{w}^* - \bm{w}_{t+1}\right\|_2^2\right] - \left\|\bm{w}^* - \bm{w}_t\right\|_2^2 \le \gamma_t^2\left(\overline{C}+\overline{D}\right)^2 + 2\gamma_t\left(L_{\tilde{x}_t}(\bm{w}^*) - L_{\tilde{x}_t}(\bm{w}_t)\right)
\]
In addition, if $d$ has a GPG of $\eps$, then for any constants $\eta$ and $H$ satisfying the condition
\[
0\le\eta\le\eps,~~\text{and}~ \gamma_t^2Z^2 - 2(\eps - \eta)H\le\eta^2,
\]
as long as $\left\|\bm{w}^* - \bm{w}_t\right\|_2 \ge \gamma_tH$,
it holds that
\[
\mathbb{E}\left[\left\|\bm{w}^* - \bm{w}_{t+1}\right\|_2\right] \le \left\|\bm{w}^* - \bm{w}_t\right\|_2 - \gamma_t\eta.
\]
\end{lemma}

The full proof of \Cref{lemma:nega_drift} has been presented in \Cref{appendix:nega_drift}. Select $\eta = \frac{\eps}{2}$ and $H = \max\{\frac{Z^2 - \eta^2}{2(\eps - \eta)},\eta\}$ and it holds that
\[
\mathbb{E}\left[\left\|\bm{w}^* - \bm{w}_{t+1}\right\|_2 - \left\|\bm{w}^* - \bm{w}_t\right\|_2|\bm{w}_t\right] \le -\gamma_t\eta
\]
Besides, we know that $\bm{w}_1 = \bm{w}^*$, which implies that $\left\|\bm{w}^* - \bm{w}_1\right\|_2 = 0$. By applying \Cref{lem:gupta}, it holds that for all $t\ge 1$,
\begin{equation} \label{ieq:w_star_t_bound}
\begin{aligned}
\mathbb{E}\left[\left\|\bm{w}^* - \bm{w}_t\right\|_2\right] &\le \gamma_tZ\left(1 + \lceil \frac{H}{Z} \rceil + \frac{Z - \eta}{2\eta}\right) \\
&\le \gamma_t\left(2Z + H + \frac{Z^2}{2\eta}\right) \\
&= \gamma_t\left(2Z + \max\{\frac{Z^2  -\eps^2/4}{\eps},\frac{\eps}{2}\} + \frac{Z^2}{\eps}\right) \\
&\le \gamma_t\left(2Z + \frac{\eps}{2} + \frac{2Z^2}{\eps}\right)
\end{aligned}
\end{equation}
which completes our proof.
\end{myproof}

Finally, with the help of \Cref{lemma:w_diff_bound} and \Cref{lemma:optimal_w_bound}, we can conclude that
\begin{equation} \label{ieq:bound1_final}
\begin{aligned}
\textbf{BOUND I} &\le\frac{1}{2\gamma_t}\left(\left(\left\|\hat{\bm{w}}^*\right\|_2 + \left\|\bm{w}_{{T_{\text{refine}}}+1}\right\|_2\right) \cdot \left(\left\|\hat{\bm{w}}^* -\bm{w}^*\right\|_2 + \left\|\bm{w}^* -\bm{w}_{{T_{\text{refine}}}+1}\right\|_2\right)\right) + \frac{T\gamma_tq}{2}\left(\overline{C}+\overline{D}\right)^2 \\
&\le \frac{1}{2\gamma_t}\left(2\overline{W} \cdot\left(\frac{1}{T} + \gamma_t\left(2Z + \frac{\eps}{2} + \frac{2Z^2}{\eps}\right)\right)\right)+ \frac{T\gamma_tZ^2}{2} \\
&= \frac{\overline{W}}{\gamma_t}\cdot\gamma_t\left(1+2Z + \frac{\eps}{2} + \frac{2Z^2}{\eps}\right) + \frac{T\gamma_tZ^2}{2}\\
&= \overline{W}\left(1+2Z + \frac{\eps}{2} + \frac{2Z^2}{\eps}\right) + \frac{T\gamma_tZ^2}{2}
\end{aligned} 
\end{equation} 
where $\overline{W} = \left(\frac{q(\overline{C}+\overline{D})^2 +2\overline{r}}{2\underline{D}} +q(\overline{C}+\overline{D})\right)$ is the upper bound of $\left\|\hat{\bm{w}}^*\right\|_2$ and $\left\|\bm{w}_{{T_{\text{refine}}}+1}\right\|_2$, and $Z = \sqrt{q}\left(\overline{C} + \overline{D}\right)$ for simplicity.

For \textbf{BOUND II}, similar to the proof of \Cref{thm:fast}, we know that:
\[
\sum_{t=1}^{T_{\text{refine}}}\bm{a}_t^\top\Phi\tilde{x}_t - \bm{d}^\top\Phi \le \frac{1}{\gamma_t}\sum_{t=1}^{T_{\text{refine}}}\bm{w}_{t+1} - \bm{w}_t = \frac{1}{\gamma_t}(\bm{w}_{T_{\text{refine}}+1} - \hat{\bm{w}}^*)
\]
Therefore, we have:
\begin{equation} 
\label{ieq:bound2_final}
\begin{aligned} 
\textbf{BOUND II} &\le \overline{r}\sum_{j=1}^q\left[\sum_{t=1}^{T_{\text{refine}}}(\bm{a}_t^\top\Phi\tilde{x}_t)_j  - (\bm{b}^\top\Phi)_j + \overline{C}\right]^+ \\
&\le \overline{r}q\left(\frac{1}{\gamma_t}\left\|\bm{w}_{T_{\text{refine}}+1} - \hat{\bm{w}}^*\right\|_2 + \overline{C}\right) \\
&\le \overline{r}q\left(\frac{1}{\gamma_t}\left(\left\|\hat{\bm{w}}^* -\bm{w}^*\right\|_2 + \left\|\bm{w}^* -\bm{w}_{{T_{\text{refine}}}+1}\right\|_2\right) + \overline{C}\right) \\
&\le \overline{r}q\left(1+ 2Z + \frac{\eps}{2} + \frac{2Z^2}{\eps} + \overline{C}\right)
\end{aligned}
\end{equation}
where the third inequality utilizes Cauchy Inequality and last one follows inequality \eqref{ieq:w_star_t_bound}. Combining \textbf{BOUND I} \eqref{ieq:bound1_final} and \textbf{BOUND II} \eqref{ieq:bound2_final}, we can get our final results:
\[
\begin{aligned}
\text{Reg}_{T_{\text{refine}}}(\pi_{\text{ALG4}}) &\le \textbf{BOUND I} + \textbf{BOUND II} \\
&\le \overline{W}\left(1+2Z + \frac{\eps}{2} + \frac{2Z^2}{\eps}\right) + \frac{T\gamma_tZ^2}{2} + \overline{r}q\left(1 + 2Z + \frac{\eps}{2} + \frac{2Z^2}{\eps} + \overline{C}\right) \\
&= \frac{T\gamma_tZ^2}{2} + O(\frac{q}{\eps}) \\
&\le O(\frac{q}{\eps})
\end{aligned}
\]
if we select $\gamma_t \le \frac{1}{T}$. Therefore, our final regret bound is independent of time horizon $T$ and our proof is thus completed.

\subsection{Proof of \Cref{lemma:nega_drift}} \label{appendix:nega_drift}
For any fixed $\bm{w}_t$, we have:
\[
\begin{aligned}
\mathbb{E}\left[\left\|\bm{w}^* - \bm{w}_{t+1}\right\|_2^2\right] - \left\|\bm{w}^* - \bm{w}_t\right\|_2^2 &= \mathbb{E}\left[\left\|\bm{w}^* - \left(\bm{w}_t + \gamma_t\left(\bm{a}_t^\top\Phi \tilde{x}_t - \bm{d}^\top\Phi\right)\right)^+\right\|_2^2\right] - \left\|\bm{w}^* - \bm{w}_t\right\|_2^2 \\
&\le \mathbb{E}\left[\left\|\bm{w}^* - \bm{w}_t - \gamma_t\left(\bm{a}_t^\top\Phi \tilde{x}_t - \bm{d}^\top\Phi\right)\right\|_2^2\right] - \left\|\bm{w}^* - \bm{w}_t\right\|_2^2 \\
&=\mathbb{E}\left[\left\|\gamma_t\left(\bm{a}_t^\top\Phi \tilde{x}_t - \bm{d}^\top\Phi\right)\right\|_2^2\right] - 2\mathbb{E}\left[\langle\bm{w}^*-\bm{w}_t,\gamma_t\left(\bm{a}_t^\top\Phi \tilde{x}_t - \bm{d}^\top\Phi\right)\rangle\right]
\end{aligned}
\]
Now we consider the following Lagrangian function $L(\bm{w})$:
\[
L(\bm{w}) = \max_{0\le x\le1} \mathbb{E}\left[rx + \bm{w}^\top\Phi^\top\left(\bm{d} - \bm{a}x\right)\right]
\]

Note that $\mathbb{E}\left[\tilde{x}_t\right]$ is the optimal solution for $L(\bm{w}_t)$ according to the Step 6 in \Cref{alg:const_reg}, while it is a feasible solution for $L(\bm{w}^*)$. Therefore, we obtain that
\[
\begin{aligned}
L_{\tilde{x}_t}(\bm{w}_t) &= \mathbb{E}\left[r\tilde{x}_t + \bm{w}_t^\top\Phi^\top\left(\bm{d} - \bm{a}\tilde{x}_t\right)\right] \\
L_{\tilde{x}_t}(\bm{w}^*) &\ge \mathbb{E}\left[r\tilde{x}_t + (\bm{w}^*)^\top\Phi^\top\left(\bm{d} - \bm{a}\tilde{x}_t\right)\right]
\end{aligned}
\]
which implies that
\[
\begin{aligned}
L_{\tilde{x}_t}(\bm{w}^*) - L_{\tilde{x}_t}(\bm{w}_t) &\ge \mathbb{E}\left[r\tilde{x}_t + (\bm{w}^*)^\top\Phi^\top\left(\bm{d} - \bm{a}\tilde{x}_t\right)\right] - \mathbb{E}\left[r\tilde{x}_t + \bm{w}_t^\top\Phi^\top\left(\bm{d} - \bm{a}\tilde{x}_t\right)\right] \\
&= \mathbb{E}\left[\langle\bm{w}_t - \bm{w}^*,\Phi^\top\left(\bm{a}\tilde{x}_t - \bm{d}\right)\rangle\right]
\end{aligned}
\]
Therefore, it holds that
\[
\begin{aligned}
    \mathbb{E}\left[\left\|\bm{w}^* - \bm{w}_{t+1}\right\|_2^2\right] - \left\|\bm{w}^* - \bm{w}_t\right\|_2^2 &\le \mathbb{E}\left[\left\|\gamma_t\left(\bm{a}_t^\top\Phi \tilde{x}_t - \bm{d}^\top\Phi\right)\right\|_2^2\right] - 2\mathbb{E}\left[\langle\bm{w}^*-\bm{w}_t,\gamma_t\left(\bm{a}_t^\top\Phi \tilde{x}_t - \bm{d}^\top\Phi\right)\rangle\right] \\
    &\le \gamma_t^2Z^2 + 2\gamma_t\left(L_{\tilde{x}_t}(\bm{w}^*) - L_{\tilde{x}_t}(\bm{w}_t)\right)
\end{aligned}
\]
which completes the proof of the first part.

Then, if $d$ has a GPG of $\eps$, assume that $\hat{d}$ satisfies the condition $\left\|\hat{\bm{d}} - \bm{d}\right\| \le \eps$, we define:
\[
\hat{L}_x(\bm{w}) = \bm{w}^\top\Phi^\top\hat{\bm{d}} + \mathbb{E}\left[\max_{0\le\bm{x} \le1} \{r\cdot x - \bm{w}^\top\Phi^\top\bm{a}x\}\right]
\]
Note that the only difference between $L_x(\bm{w})$ and $\hat{L}_x(\bm{w})$ is that $L_x(\bm{w})$ is based on $d$ and $\hat{L}_x(\bm{w})$ is based on on $\hat{d}$. Then, it holds that
\begin{equation} \label{eq:lem5_mid}
L_x(\bm{w}) - L_x(\bm{w}^*) = \hat{L}_x(\bm{w}) - \hat{L}_x(\bm{w}^*) + \langle\bm{w}^* - \bm{w},\hat{\bm{d}} - \bm{d}\rangle 
\ge \langle\bm{w}^* - \bm{w},\hat{\bm{d}} - \bm{d}\rangle \ge \eps\left\|\bm{w}^* - \bm{w}\right\|_2
\end{equation}
where the first inequality holds since $\bm{w}^*$ is also the optimal solution for the problem $\min_{\bm{w}\ge0} \hat{L}(\bm{w})$ according to our GPG assumption.

Then we can obtain that
\[
\mathbb{E}\left[\left\|\bm{w}^* - \bm{w}_{t+1}\right\|_2^2\right] - \left\|\bm{w}^* - \bm{w}_t\right\|_2^2 
\le \gamma_t^2Z^2 - 2\gamma_t\left(L_{\tilde{x}_t}(\bm{w}_t) - L_{\tilde{x}_t}(\bm{w}^*)\right)
\le \gamma_t^2Z^2 - 2\eps\gamma_t\left\|\bm{w}_t - \bm{w}^*\right\|_2
\]

Therefore, for any constants $\eta$ and $H$ satisfying the condition $0\le\eta\le\eps,~~\text{and}~ \gamma_t^2Z^2 - 2(\eps - \eta)H\le\eta^2$, as long as $\left\|\bm{w}^* - \bm{w}_t\right\|_2 \ge \gamma_tH$, it holds that
\[
\begin{aligned}
\mathbb{E}\left[\left\|\bm{w}^* - \bm{w}_{t+1}\right\|_2^2\right] &\le  \left\|\bm{w}^* - \bm{w}_t\right\|_2^2 + \gamma_t^2Z^2 - 2\eps\gamma_t\left\|\bm{w}_t - \bm{w}^*\right\|_2 \\
&= \left\|\bm{w}^* - \bm{w}_t\right\|_2^2 +  \gamma_t^2Z^2 - 2\gamma_t(\eps-\eta)\left\|\bm{w}_t - \bm{w}^*\right\|_2 - 2\gamma_t\eta\left\|\bm{w}_t - \bm{w}^*\right\|_2 \\
&\le \left\|\bm{w}^* - \bm{w}_t\right\|_2^2  - 2\gamma_t\eta\left\|\bm{w}_t - \bm{w}^*\right\|_2 + \gamma_t^2\eta^2 \\
&= \left(\left\|\bm{w}^* - \bm{w}_t\right\|_2 - \gamma_t\eta\right)^2
\end{aligned}
\]
which implies that
\[
\mathbb{E}\left[\left\|\bm{w}^* - \bm{w}_{t+1}\right\|_2\right] \le \left\|\bm{w}^* - \bm{w}_t\right\|_2 - \gamma_t\eta.
\]

Our proof is thus completed.

\section{Proof of \Cref{sec:general}}

\subsection{Proof of \Cref{thm:general_md}} \label{appendix:general_md}

First, according to the KKT condition of the Mirror Descent, we have
\begin{equation}
\begin{aligned}
    \gamma_t \langle \bm{y}_t, \bm{w}_{t+1} - \bm{u} \rangle &\le \langle \nabla \psi(w_{t+1}) - \nabla\psi(w_t), u - w_{t+1}\rangle \\
    &=D_\psi(u||w_t) - D_\psi(u||w_{t+1}) - D_\psi(w_{t+1}||w_t)
\end{aligned}
\end{equation}
for any $\bm{u} \in \mathbb{R}^m_{\ge0}$. Adding the term $\langle \bm{y}_t,\bm{w}_t - \bm{w}_{t+1} \rangle$ to both sides and choosing $\gamma_t=\frac{1}{\sqrt{T}}$, we have
\begin{equation}
\begin{aligned}
    \langle \bm{y}_t, \bm{w}_t - \bm{u} \rangle &\le \frac{1}{\sqrt{T}} \left(D_\psi(\bm{u}||\bm{w}_t) - D_\psi(\bm{u}||\bm{w}_{t+1}) - D_\psi(\bm{w}_{t+1}||\bm{w}_t) + \langle \bm{y}_t, \bm{w}_t - \bm{w}_{t+1} \rangle\right) \\
    &\le \frac{1}{\sqrt{T}} \left(D_\psi(\bm{u}||\bm{w}_t) - D_\psi(\bm{u}||\bm{w}_{t+1}) - \frac{\alpha}{2}||\bm{w}_{t+1}-\bm{w}_t||^2 + \langle \bm{y}_t, \bm{w}_t - \bm{w}_{t+1} \rangle\right) \\
    &\le \frac{1}{\sqrt{T}} \left(D_\psi(\bm{u}||\bm{w}_t) - D_\psi(\bm{u}||\bm{w}_{t+1}) - \frac{\alpha}{2}||\bm{w}_{t+1}-\bm{w}_t||^2 + \frac{\alpha}{2}||\bm{w}_{t+1}-\bm{w}_t||^2 + \frac{1}{2\alpha}||\bm{y}_t||_*^2\right) \\
    &= \frac{1}{\sqrt{T}} \left(D_\psi(\bm{u}||\bm{w}_t) - D_\psi(\bm{u}||\bm{w}_{t+1}) + \frac{1}{2\alpha}||\bm{y}_t||_*^2\right)
\end{aligned}
\end{equation}
where the second inequality depends on the $\alpha$-strongly convexity of the potential function $\psi$ and the third inequality holds since
\[
\langle \bm{y}_t, \bm{w}_t - \bm{w}_{t+1} \rangle \le \left\|\bm{y}_t\right\|\cdot \left\|\bm{w}_t - \bm{w}_{t+1}\right\| \le \frac{\alpha}{2}||\bm{w}_{t+1}-\bm{w}_t||^2 + \frac{1}{2\alpha}||\bm{y}_t||_*^2.
\]

Furthermore, similar to the proof of \Cref{thm:simple}, we have:
\[
\text{Reg}_T(\pi_{\text{ALG6}}) \le \sum^T_{t=1}\mathbb{E}_{\bm{\theta_t}\sim\mathcal{P}}[(\bm{d}^\top\Phi-\bm{g}(x_t;\bm{\theta}_t)^\top\Phi)\bm{w}_t] = \sum^T_{t=1}\mathbb{E}[\langle \bm{y}_t, \bm{w}_t\rangle]
\]
Combing these two inequalities above and choosing $\bm{u} = \bm{0}$, it holds that:
\begin{equation}
\begin{aligned}
    \text{Reg}_T(\pi_{\text{ALG6}}) &\le \sum^T_{t=1}\mathbb{E}[\langle \bm{y}_t, \bm{w}_t\rangle] \\
    &\le \frac{1}{\sqrt{T}} \sum^T_{t=1} \left(D_\psi(\bm{0}||\bm{w}_t) - D_\psi(\bm{0}||\bm{w}_{t+1}) + \frac{1}{2\alpha}||\bm{y}_t||_*^2\right) \\
    &= \frac{1}{\sqrt{T}}  \left(D_\psi(\bm{0}||\bm{w}_1) - D_\psi(\bm{0}||\bm{w}_{T+1}) + \sum^T_{t=1}\frac{1}{2\alpha}||\bm{y}_t||_*^2\right) \\
    &\le \frac{1}{\sqrt{T}}  \sum^T_{t=1}\frac{1}{2\alpha}||\bm{y}_t||_*^2 \\
    &= O(\sqrt{qT})
\end{aligned}
\end{equation}
where $||\bm{y}_t||_*^2\le K ||\bm{d}^\top\Phi - \bm{g}(x_t;\bm{\theta}_t)^\top\Phi||_2^2\le Kq(\overline{G}+\overline{D})^2$ and $K$ is only dependent on $q$ and norm $\left\|\cdot\right\|$.

For constraints violation, define a stopping time $\tau$ for the first time that there exist a resource $j$ such that $\sum_{t=1}^\tau \bm{g}_j(x_t;\bm{\theta}_t)^\top\Phi_j + \overline{G} \ge b_j^\top\Phi_j=(d_j^\top\Phi_j)\cdot T$. Therefore, the constraints violation can be bounded as:
\[
\begin{aligned}
v(\pi_\text{ALG6}) &= \mathbb{E}_{\bm{\theta_t}\sim\mathcal{P}}\left[\max_{p\ge0} \left\|\left[p^\top\left(\sum_{t=1}^T\bm{g}(x_t;\bm{\theta}_t) - \bm{b}\right)\right]^+\right\|_2\right] \\&= \mathbb{E}_{\bm{\theta_t}\sim\mathcal{P}}\left[\max_{w\ge0}\left\|\left[w(\Phi^\top \sum_{t=1}^T\bm{g}(x_t;\bm{\theta}_t) - \Phi^\top \bm{b})\right]^+\right\|_2\right] \\ &\le \max_{w\ge0} \left\|w\right\|_2\cdot  \mathbb{E}_{\bm{\theta_t}\sim\mathcal{P}}\left[\left\|\left[\Phi^\top \sum_{t=1}^T\bm{g}(x_t;\bm{\theta}_t) - \Phi^\top \bm{b}\right]^+\right\|_2\right]
\end{aligned}
\]
where \Cref{assump:sepa} further gives an upper bound to $\left\|\bm{w}\right\|_2$ as $\overline{W}$.

According to the definition of stopping time $\tau$, we know that:
\begin{equation} \label{ieq:general_constr_viol}
\mathbb{E}_{\bm{\theta_t}\sim\mathcal{P}}\left[\left\|\left(\Phi^\top \sum_{t=1}^T\bm{g}(x_t;\bm{\theta}_t) - \Phi^\top b\right)^+\right\|_2\right] \le \sqrt{\sum_{j=1}^q \left(\overline{G}\cdot\left(T - \tau\right)\right)^2} = \overline{G}\cdot\left(T - \tau\right)\sqrt{q}
\end{equation}

Therefore, we only need to bound the term $T - \tau$. Since we use mirror descent (Step $5-6$ in \Cref{alg:general_md}) to update $\bm{w}$, according to the KKT condition, it holds that
\[
\nabla\psi(\bm{w}_{t+1}) = \nabla\psi(\bm{w}_{t}) - \gamma_t\bm{y}_t - \bm{v}_{t+1}
\]
where $\bm{v}_{t+1} \in \mathcal{N}_{\mathbb{R}_\ge0}$ and $\mathcal{N}_{\mathbb{R}_\ge0}$ is the normal core of the set $\mathbb{R}_\ge0$ at point $\bm{w}_{t+1}$. Further, the definition of $\mathcal{N}_{\mathbb{R}_\ge0}$:
\[
\mathcal{N}_{\mathbb{R}_\ge0}(x) = \{\bm{v}\in\mathbb{R}^d:v_j = 0 ~~\text{if}~~ x_j > 0; v_j \le 0 ~~\text{if}~~ x_j = 0\}
\]
implies that the condition $\bm{v}_{t+1}$ always holds. Therefore, under \Cref{assump:sepa}, it holds that
\[
\gamma_t\bm{y}_t \ge \nabla\psi(\bm{w}_{t}) - \nabla\psi(\bm{w}_{t+1}) = \nabla\psi_j(\bm{w}_{t}) - \nabla\psi_j(\bm{w}_{t+1})
\]
where $j$ is the index of the first constraint violation happens at time $\tau$. In addition, according to the definition of $\bm{g}_t$, we know that
\begin{equation} \label{eq:gneral_def_gt}
\sum_{t=1}^\tau(\bm{y}_t)_j = \sum_{t=1}^\tau (\bm{d}^\top\Phi - \bm{g}(x_t;\bm{\theta}_t)^\top\Phi)_j \le \sum_{t=1}^\tau (\bm{d}^\top\Phi)_j -  (\bm{d}^\top\Phi)_j\cdot T  +\overline{G}=(\bm{d}^\top\Phi)_j\cdot (\tau-T) + \overline{G}
\end{equation}
Therefore, we can obtain
\begin{equation} \label{ieq:general_T_tau}
\begin{aligned}
    T - \tau &\le \frac{\overline{G} - \sum_{t=1}^\tau(\bm{y}_t)_j}{(\bm{d}^\top\Phi)_j} \\
    &\le \frac{1}{\sqrt{T}}\cdot\frac{\sqrt{T}\overline{G} + \sum_{t=1}^\tau (\nabla\psi_j(\bm{w}_{t+1}) - \nabla\psi_j(\bm{w}_{t}))}{\underline{D}} \\
    &\le \frac{1}{\sqrt{T}}\cdot\frac{\sqrt{T}\overline{G} + \nabla\psi_j(\overline{W}) - \nabla\psi_j(\bm{0})}{\underline{D}} \\
    &=O(\sqrt{T})
\end{aligned}
\end{equation}
where the first inequality follows from \eqref{ieq:general_T_tau}, the second one depends on the definition of $\tau$ and \Cref{asp:bounded_data}, and the last one utilizes the monotonicity of potential function $\psi_j$.

Combining two inequalities \eqref{ieq:general_constr_viol} and \eqref{ieq:general_T_tau}, we can conclude that:
\[
v(\pi_\text{ALG6}) \le \max_{w\ge0} \left\|w\right\|_2\cdot  \mathbb{E}_{\bm{\theta_t}\sim\mathcal{P}}\left[\left\|\left[\Phi^\top \bm{g}(x_t;\bm{\theta}_t) - \Phi^\top b\right]^+\right\|_2\right] \le \overline{W} \cdot O(\sqrt{qT}) = O(\sqrt{qT}).
\]

Thus we complete our proof.

\subsection{Proof of \Cref{lem:general_w_upper_bound}}

Note that in \Cref{alg:general_log}, there are two update formula for $\bm{w}_t$; however, the only difference between them is their step size $\gamma_t$. Therefore we introduce a parameter $\gamma$ to represent the general case, and this parameter $\gamma$ will be removed in the subsequent proof. According to the update formula, we can obtain that
\begin{equation} \label{ieq:general_w}
\begin{aligned}
    ||\bm{w}_{t+1}||_2^2 & \le ||\bm{w}_t + \gamma(\bm{g}(\tilde{x}_t;\bm{\theta}_t)^\top_t\Phi -\bm{d}^\top \Phi)||_2^2\\
    &= ||\bm{w}_t||_2^2 - 2\gamma(\bm{d}^\top\Phi - \bm{g}(\tilde{x}_t;\bm{\theta}_t)^\top\Phi) \bm{w}_t +\gamma_t^2||\bm{d}^\top\Phi - \bm{g}(\tilde{x}_t;\bm{\theta}_t)^\top\Phi||_2^2 \\
    &\le ||\bm{w}_t||_2^2 - 2\gamma\bm{d}^\top\Phi \bm{w}_t +2\gamma\bm{g}(\tilde{x}_t;\bm{\theta}_t)^\top\Phi\bm{w}_t +\gamma_t^2q(\overline{G}+\overline{D})^2 \\
    &\le ||\bm{w}_t||_2^2 +\gamma^2q(\overline{G}+\overline{D})^2 + 2\gamma (f(\tilde{x}_t;\bm{\theta}_t) + \overline{F}) - 2\gamma\bm{d}^\top\Phi \bm{w}_t \\
    &\le ||\bm{w}_t||_2^2 +\gamma^2q(\overline{G}+\overline{D})^2 + 4\gamma \overline{F} - 2\gamma\bm{d}^\top\Phi \bm{w}_t
\end{aligned}
\end{equation}
where the first inequality follows from the update formula \eqref{ieq:general_w}, the second and last inequality relies on \Cref{asp:general_bounded}. For the third inequality, from the definition of $\tilde{x}_t$, we know that $f(\tilde{x}_t;\bm{\theta}_t) - \bm{g}(\tilde{x}_t;\bm{\theta}_t)^\top\Phi\bm{w}_t \ge -\overline{F}$; otherwise, $\tilde{x}_t$ will be selected as $0$ and then $f(\tilde{x}_t;\bm{\theta}_t) - \bm{g}(\tilde{x}_t;\bm{\theta}_t)^\top\Phi\bm{w}_t \ge -\overline{F}$ still holds since $\bm{g}(0;\bm{\theta}_t) = 0$.
Next, we will show that when $||\bm{w}_t||_2$ is large enough, there must have $||\bm{w}_{t+1}||_2 \le ||\bm{w}_t||_2$. To be specific, when $||\bm{w}_t||_2 \ge \frac{q(\overline{G}+\overline{D})^2 +4\overline{F}}{2\underline{D}}$, we have the following inequalities:
\begin{equation} \label{general_w_bound:first}
\begin{aligned}
    ||\bm{w}_{t+1}||_2^2 - ||\bm{w}_t||_2^2 &\le \gamma^2q(\overline{G}+\overline{D})^2 + 4\gamma\overline{F} - 2\gamma\bm{d}^\top\Phi \bm{w}_t \\
    &\le \gamma_t^2q(\overline{G}+\overline{D})^2 + 4\gamma\overline{F} - 2\gamma \underline{D} \bm{w}_t \\
    &\le q(\overline{G}+\overline{D})^2 + 4\overline{F} - 2 \underline{D} \bm{w}_t \\
    &\le 0
\end{aligned}
\end{equation}
where the first inequality comes from \eqref{ieq:general_w}, the second inequality depends on \Cref{asp:general_bounded} and the last one utilizes the fact that $\gamma \le 1$. Otherwise, when $||\bm{w}_t||_2 < \frac{q(\overline{G}+\overline{D})^2 +4\overline{F}}{2\underline{D}}$, we have:
\begin{equation} \label{general_w_bound:second}
\begin{aligned}
    ||\bm{w}_{t+1}||_2 & \le ||\bm{w}_t + \gamma_t(\bm{a}_t^\top\Phi x_t -\bm{d}^\top \Phi)||_2\\
    &\le ||\bm{w}_t||_2 + ||\gamma_t(\bm{a}_t^\top\Phi x_t -\bm{d} \Phi)||_2 \\
    &\le \frac{q(\overline{G}+\overline{D})^2 +4\overline{F}}{2\underline{D}} +q(\overline{G}+\overline{D})
\end{aligned}
\end{equation}
Combining the two cases above, suppose $t=\tau$ is the first time that $||\bm{w}_t||_2 \ge \frac{q(\overline{G}+\overline{D})^2 +4\overline{F}}{2\underline{D}}$. Since $||\bm{w}_{\tau-1}||_2 < \frac{q(\overline{G}+\overline{D})^2 +4\overline{F}}{2\underline{D}}$, from the inequality \eqref{general_w_bound:second} we know that $||\bm{w}_{\tau}||_2 \le \frac{q(\overline{G}+\overline{D})^2 +4\overline{F}}{2\underline{D}}$. Besides, from the inequality \eqref{general_w_bound:first}, we know that $\bm{w}$ will decrease until it falls below the threshold $\frac{q(\overline{G}+\overline{D})^2 +4\overline{F}}{2\underline{D}}+q(\overline{G}+\overline{D})$.

Overall, we can conclude that:
\[
||\bm{w}_t||_2 \le \frac{q(\overline{G}+\overline{D})^2 +4\overline{F}}{2\underline{D}} +q(\overline{G}+\overline{D})
\]
which completes our proof.

\subsection{Proof of \Cref{thm:general_log}}
Similar to \Cref{sec:log_regret}, we split \Cref{alg:general_log} into two parts: Accelerate Stage (Step $4-16$) and Refine Stage (Step $18-23$), and the following two lemmas give their regrets respectively. 
\begin{lemma} \label{lem:general_fast}
    Under \Cref{asp:GPG} with parameter $\eps$, \Cref{asp:holder} and \Cref{asp:general_bounded}, considering only Step $4-16$ in \Cref{alg:general_log}, which is the Accelerate Stage, if we select step size $\gamma_t = \frac{1}{\log{T}}$, the regret of Step $4-16$ is upper bounded by $O(q\log{T})$.
\end{lemma}

\begin{lemma} \label{lem:general_const_reg_bound}
    Under \Cref{asp:GPG} with parameter $\eps$, \Cref{asp:holder} and \Cref{asp:general_bounded}, considering only Step $18-23$ in \Cref{alg:general_log}, which is the Refine Stage, if we select step size $\gamma_t \le \frac{1}{T}$, the regret of Step $18-23$ is upper bounded by $O(q/\eps)$ with high probability $1-\delta$.
\end{lemma}

Then we can immediately obtain \Cref{thm:general_log}, which completes our proof.

\subsection{Proof of \Cref{lem:general_fast}}
First, since Step $4-16$ in \Cref{alg:general_log} are dependent on a virtual decision $\tilde{x}_t$ and collectively consume time $T_{\text{fast}}$, we know that
\begin{equation} \nonumber
\begin{aligned}
\text{Reg}_{T_{\text{fast}}}(\pi_{\text{ALG7}}) &\le \mathbb{E}_{\bm{\theta_t}\sim\mathcal{P}}\left[R_\Phi^* - \sum_{t=1}^{T_{\text{fast}}} f(x_t;\bm{\theta}_t)\right] \\
&\le {T_{\text{fast}}}\cdot h(\bm{w}^*) - \mathbb{E}_{\bm{\theta_t}\sim\mathcal{P}}\left[\sum_{t=1}^{T_{\text{fast}}} f(\tilde{x}_t;\bm{\theta}_t)\right] + \mathbb{E}_{\bm{\theta_t}\sim\mathcal{P}}\left[\sum_{t=1}^{T_{\text{fast}}} f(\tilde{x}_t;\bm{\theta}_t)\right] - \mathbb{E}_{\bm{\theta_t}\sim\mathcal{P}}\left[\sum_{t=1}^{T_{\text{fast}}} f(x_t;\bm{\theta}_t)\right] \\
&\le \sum_{t=1}^{T_{\text{fast}}} h(\bm{w}_t) - \mathbb{E}_{\bm{\theta_t}\sim\mathcal{P}}\left[\sum_{t=1}^{T_{\text{fast}}} f(\tilde{x}_t;\bm{\theta}_t)\right] + \mathbb{E}_{\bm{\theta_t}\sim\mathcal{P}}\left[\sum_{t=1}^{T_{\text{fast}}} f(\tilde{x}_t;\bm{\theta}_t)\right] - \mathbb{E}_{\bm{\theta_t}\sim\mathcal{P}}\left[\sum_{t=1}^{T_{\text{fast}}} f(x_t;\bm{\theta}_t)\right] \\
&\le \underbrace{\sum_{t=1}^{T_{\text{fast}}}\mathbb{E}_{\bm{\theta_t}\sim\mathcal{P}}\left[\left(\bm{d}^\top\Phi - \bm{g}(\tilde{x}_t;\bm{\theta}_t)^\top \Phi\right)\bm{w}_t\right]}_{\textbf{BOUND I}} + \underbrace{\mathbb{E}_{\bm{\theta_t}\sim\mathcal{P}}\left[\sum_{t=1}^{T_{\text{fast}}} f(\tilde{x}_t;\bm{\theta}_t) -f(x_t;\bm{\theta}_t)\right]}_{\textbf{BOUND II}}
\end{aligned}
\end{equation}
We then separately bound these two parts.

For \textbf{BOUND I}, according to the Step $11$ of \Cref{alg:general_log}, we know that
\begin{equation} \label{ieq:general_fast_update_w}
\begin{aligned}
\left\|\bm{w}_{t+1}\right\|_2 &\le \left\|\bm{w}_t\right\|_2 - 2\gamma_t\left(\bm{d}^\top\Phi - \bm{g}(\tilde{x}_t;\bm{\theta}_t)^\top \Phi\right)\bm{w}_t + \gamma_T^2\left\|\bm{d}^\top\Phi - \bm{g}(\tilde{x}_t;\bm{\theta}_t)^\top \Phi\right\|^2 \\
&\le \left\|\bm{w}_t\right\|_2 - \frac{2}{\log{T}}\left(\bm{d}^\top\Phi - \bm{g}(\tilde{x}_t;\bm{\theta}_t)^\top \Phi\right)\bm{w}_t + \frac{q(\overline{G}+\overline{D})^2}{\log^2{T}}^2 
\end{aligned}
\end{equation}
Therefore, it holds that
\[
\begin{aligned}
\textbf{BOUND I} &= \sum_{t=1}^{T_{\text{fast}}}\mathbb{E}_{\bm{\theta_t}\sim\mathcal{P}}\left[\left(\bm{d}^\top\Phi - \bm{g}(\tilde{x}_t;\bm{\theta}_t)^\top \Phi\right)\bm{w}_t\right] \\
&\le \mathbb{E}\left[\frac{\log{T}}{2}\sum_{t=1}^{T_{\text{fast}}} \left(\bm{w}_t - \bm{w}_{t+1}\right) + T_{\text{fast}}\cdot\frac{q(\overline{G}+\overline{D})^2}{2\log{T}}\right]\\
&= \mathbb{E}\left[\frac{\log{T}}{2}\bm{w}_0+T_{\text{fast}}\cdot\frac{q(\overline{G}+\overline{D})^2}{2\log{T}}\right]\\
&=O(q\log{T}).
\end{aligned}
\]
where the first inequality comes from the inequality \eqref{ieq:general_fast_update_w} and the second equality depends on the fact that $\bm{w}_0 = 0$ and $T_{\text{fast}} = O(\log^2{T})$.

For \textbf{BOUND II}, first it holds that:
\[
f(\tilde{x}_t;\bm{\theta}_t) - f(x_t;\bm{\theta}_t) \le f(\tilde{x}_t;\bm{\theta}_t)\cdot\mathbb{I}\{\exists j : B_{t,j} < (\bm{g}(\tilde{x}_t;\bm{\theta}_t)^\top\Phi)_j\}
\]
where $B_{t,j}$ denotes the $j$-th component of $B_t$ and $\mathbb{I}$ denotes the indicator function. In addition, we can observe that:
\[
\mathbb{I}\{\exists j : B_{t,j} < \bm{g}(\tilde{x}_t;\bm{\theta}_t)^\top\Phi\} < \sum_{j=1}^q \mathbb{I}\left\{\sum_{i=1}^t(\bm{g}_i(\tilde{x}_i;\bm{\theta}_i)^\top\Phi)_j > B_{j}\right\}
\]
Therefore, we have:
\[
\begin{aligned}
\sum_{t=1}^{T_{\text{fast}}}f(\tilde{x}_t;\bm{\theta}_t) - f(x_t;\bm{\theta}_t) &\le \sum_{t=1}^{T_{\text{fast}}}f(\tilde{x}_t;\bm{\theta}_t)\cdot\mathbb{I}\{\exists j : B_{t,j} < (\bm{g}(\tilde{x}_t;\bm{\theta}_t)^\top\Phi)_j\} \\
&\le\sum_{t=1}^{T_{\text{fast}}}\overline{F}\cdot\sum_{j=1}^q\mathbb{I}\left\{\sum_{i=1}^t(\bm{g}_i(\tilde{x}_i;\bm{\theta}_i)^\top\Phi)_j > B_{j}\right\} \\
&\le \overline{F}\sum_{j=1}^q\sum_{t=1}^{T_{\text{fast}}}\mathbb{I}\left\{\sum_{i=1}^t(\bm{g}_i(\tilde{x}_i;\bm{\theta}_i)^\top\Phi)_j > B_{j}\right\}
\end{aligned}
\]

Define that $S_j(t) = \sum_{i=1}^t (\bm{g}_i(\tilde{x}_i;\bm{\theta}_i)^\top\Phi)_j$ and $\tau = \inf\{\tau:S_j(t)>(\bm{b}^\top\Phi)_j\}$. We consider the following two conditions:

i) If $\tau = \infty$, then for any $j \in [Q]$, $\sum_{i=1}^t (\bm{g}_i(\tilde{x}_i;\bm{\theta}_i)^\top\Phi)_j \le (\bm{b}^\top\Phi)_j$ will always satisfy. That is to say, the constraints will not be violated. Therefore the term $\mathbb{I}\left\{\sum_{i=1}^t(\bm{g}_i(\tilde{x}_i;\bm{\theta}_t)^\top\Phi)_j > B_{j}\right\}$ will be $0$.

ii) If $\tau < \infty$, since $\bm{g}(\tilde{x}_t;\bm{\theta}_t)^\top\Phi$ has an upper bound $\overline{C}$, it must hold that $S_j(\tau-1) >(\bm{b}^\top\Phi)_j - \overline{G}$. Then we have:
\[
S_j(\tau) - S_J(\tau-1) \le S_j(\tau) - (\bm{b}^\top\Phi)_j + \overline{G}
\]

Combining two conditions above, we can conclude that:
\[
\begin{aligned}
\sum_{t=1}^{T_{\text{fast}}}f(\tilde{x}_t;\bm{\theta}_t) - f(x_t;\bm{\theta}_t)
&\le \overline{F}\sum_{j=1}^q\sum_{t=1}^{T_{\text{fast}}}\mathbb{I}\left\{\sum_{i=1}^t(\bm{g}_i(\tilde{x}_i;\bm{\theta}_i)^\top\Phi)_j > B_{j}\right\} \\
&\le \overline{r}\sum_{j=1}^q\left[\sum_{t=1}^{T_{\text{fast}}}(\bm{g}(\tilde{x}_i;\bm{\theta}_i)^\top\Phi)_j  - (\bm{b}^\top\Phi)_j + \overline{G}\right]^+
\end{aligned}
\]

Revisiting the Step $11$ in \Cref{alg:general_log}, we know that:
\[
\sum_{t=1}^{T_{\text{fast}}}\bm{g}(\tilde{x}_t;\bm{\theta}_t)^\top\Phi - \bm{d}^\top\Phi \le \frac{1}{\gamma_t}\sum_{t=1}^{T_{\text{fast}}}\bm{w}_{t+1} - \bm{w}_t = \frac{1}{\gamma_t}\cdot\bm{w}_{T_{\text{fast}}+1}
\]
Therefore, we have:
\begin{equation} 
\begin{aligned} 
\textbf{BOUND II} &\le \overline{F}\sum_{j=1}^q\left[\sum_{t=1}^T(\bm{g}(\tilde{x}_t;\bm{\theta}_t)^\top\Phi)_j  - (\bm{b}^\top\Phi)_j + \overline{G}\right]^+ \\
&\le \overline{F}q\left(\frac{1}{\gamma_t}\cdot\bm{w}_{T_{\text{fast}}+1} + \overline{G}\right) \\ \nonumber
&= O(q\log{T})
\end{aligned}
\end{equation}
since $\bm{w}_{T_{\text{fast}}+1}$ can be bounded by \Cref{lem:general_w_upper_bound} and $\gamma_t = \frac{1}{\log{T}}$ .

Combining \textbf{BOUND I} and \textbf{BOUND II} above, we can obtain our final results:
\[
\begin{aligned}
\text{Reg}_{T_{\text{fast}}}(\pi_{\text{ALG7}}) \le \textbf{BOUND I} + \textbf{BOUND II} \le O(q\log{T}).
\end{aligned}
\]
if we select $\gamma_t \le \frac{1}{log{T}}$ for $t=1,...,T_{\text{fast}}$. Overall, our proof is thus completed.

\subsection{Proof of \Cref{lem:general_const_reg_bound}} \label{appendix:general_const_reg}

Similar to the proof of \Cref{lem:general_fast}, we know that
\[
\begin{aligned}
\text{Reg}_{T_{\text{refine}}}(\pi_{\text{ALG7}}) &\le \mathbb{E}_{\bm{\theta_t}\sim\mathcal{P}}\left[R_\Phi^* - \sum_{t=1}^{T_{\text{refine}}} f(x_t;\bm{\theta}_t)\right] \\
&\le {T_{\text{refine}}}\cdot h(\bm{w}^*) - \mathbb{E}_{\bm{\theta_t}\sim\mathcal{P}}\left[\sum_{t=1}^{T_{\text{refine}}} f(\tilde{x}_t;\bm{\theta}_t)\right] + \mathbb{E}_{\bm{\theta_t}\sim\mathcal{P}}\left[\sum_{t=1}^{T_{\text{refine}}} f(\tilde{x}_t;\bm{\theta}_t)\right] - \mathbb{E}_{\bm{\theta_t}\sim\mathcal{P}}\left[\sum_{t=1}^{T_{\text{refine}}} f(x_t;\bm{\theta}_t)\right] \\
&\le \sum_{t=1}^{T_{\text{refine}}} h(\bm{w}_t) - \mathbb{E}_{\bm{\theta_t}\sim\mathcal{P}}\left[\sum_{t=1}^{T_{\text{refine}}} f(\tilde{x}_t;\bm{\theta}_t)\right] + \mathbb{E}_{\bm{\theta_t}\sim\mathcal{P}}\left[\sum_{t=1}^{T_{\text{refine}}} f(\tilde{x}_t;\bm{\theta}_t)\right] - \mathbb{E}_{\bm{\theta_t}\sim\mathcal{P}}\left[\sum_{t=1}^{T_{\text{refine}}} f(x_t;\bm{\theta}_t)\right] \\
&\le \underbrace{\sum_{t=1}^{T_{\text{refine}}}\mathbb{E}_{\bm{\theta_t}\sim\mathcal{P}}\left[\left(\bm{d}^\top\Phi - \bm{g}(\tilde{x}_t;\bm{\theta}_t)^\top \Phi\right)\bm{w}_t\right]}_{\textbf{BOUND I}} + \underbrace{\mathbb{E}_{\bm{\theta_t}\sim\mathcal{P}}\left[\sum_{t=1}^{T_{\text{refine}}} f(\tilde{x}_t;\bm{\theta}_t) -f(x_t;\bm{\theta}_t)\right]}_{\textbf{BOUND II}}
\end{aligned}
\]
We then separately bound these two parts.

For \textbf{BOUND I}, according to the Step $21$ in \Cref{alg:general_log}, we have:
\[
\begin{aligned}
\left\|\bm{w}_{t+1}\right\|_2 &\le \left\|\bm{w}_t\right\|_2 - 2\gamma_t\left(\bm{d}^\top\Phi - \bm{g}(\tilde{x}_t;\bm{\theta}_t)^\top \Phi\right)\bm{w}_t + \gamma_T^2\left\|\bm{d}^\top\Phi - \bm{g}(\tilde{x}_t;\bm{\theta}_t)^\top \Phi\right\|^2 \\
&\le \left\|\bm{w}_t\right\|_2 - \frac{2}{\log{T}}\left(\bm{d}^\top\Phi - \bm{g}(\tilde{x}_t;\bm{\theta}_t)^\top \Phi\right)\bm{w}_t + \frac{q(\overline{G}+\overline{D})^2}{\log^2{T}}^2 
\end{aligned}
\]
Therefore, it holds that:
\[
\begin{aligned}
    \textbf{BOUND I} &\le \sum_{t=1}^{T_{\text{refine}}}\frac{1}{2\gamma_t}\left(\left\|\bm{w}_t\right\|^2_2 - \left\|\bm{w}_{t+1}\right\|^2_2\right) + \frac{\gamma_t}{2}\left(\bm{d}^\top\Phi - \bm{g}(\tilde{x}_t;\bm{\theta}_t)^\top\Phi\tilde{x}_t\right)^2 \\
    &\le \frac{1}{2\gamma_t}\left(\left\|\tilde{\bm{w}}_L\right\|^2_2 - \left\|\bm{w}_{{T_{\text{refine}}}+1}\right\|^2_2\right) + \frac{T\gamma_tq}{2}\left(\overline{G}+\overline{D}\right)^2 \\
    &= \frac{1}{2\gamma_t}\left(\langle\tilde{\bm{w}}_L +\bm{w}_{{T_{\text{refine}}}+1},\tilde{\bm{w}}_L-\bm{w}_{{T_{\text{refine}}}+1} \rangle\right) + \frac{T\gamma_tq}{2}\left(\overline{G}+\overline{D}\right)^2 \\
    &\le \frac{1}{2\gamma_t}\left(\left\|\tilde{\bm{w}}_L + \bm{w}_{{T_{\text{refine}}}+1}\right\|_2 \cdot \left\|\tilde{\bm{w}}_L -\bm{w}_{{T_{\text{refine}}}+1}\right\|_2\right) + \frac{T\gamma_tq}{2}\left(\overline{G}+\overline{D}\right)^2 \\
    &\le \frac{1}{2\gamma_t}\left(\left(\left\|\tilde{\bm{w}}_L\right\|_2 + \left\|\bm{w}_{{T_{\text{refine}}}+1}\right\|_2\right) \cdot \left\|\tilde{\bm{w}}_L -\bm{w}_{{T_{\text{refine}}}+1}\right\|_2\right) + \frac{T\gamma_tq}{2}\left(\overline{G}+\overline{D}\right)^2 \\
    &\le \frac{1}{2\gamma_t}\left(\left(\left\|\tilde{\bm{w}}_L\right\|_2 + \left\|\bm{w}_{{T_{\text{refine}}}+1}\right\|_2\right) \cdot \left(\left\|\tilde{\bm{w}}_L -\bm{w}^*\right\|_2 + \left\|\bm{w}^* -\bm{w}_{{T_{\text{refine}}}+1}\right\|_2\right)\right) + \frac{T\gamma_tq}{2}\left(\overline{G}+\overline{D}\right)^2
\end{aligned}
\]
where the second inequality follows from \Cref{asp:general_bounded} and the fact that $\tilde{\bm{w}}_L$ is the start point of $\bm{w}$ in the second stage, the third one utilizes the Cauchy–Schwarz Inequality and the forth and last one is obtained from the triangle inequality for the Euclidean norm.
Note that the upper bound of $\left\|\tilde{\bm{w}}_L\right\|_2$ and $\left\|\bm{w}_{{T_{\text{refine}}}+1}\right\|_2$ can be similarly given by \Cref{lem:general_w_upper_bound}.
In addition, similar to \Cref{thm:fast} we can obtain that $\left\|\tilde{\bm{w}}_L -\bm{w}^*\right\|_2$ is upper bounded by $\frac{1}{T}$ with high probability $1-\delta$.
Therefore we now focus on the upper bound of $\left\|\tilde{\bm{w}}_L -\bm{w}_{{T_{\text{refine}}}+1}\right\|_2$, which is presented in the following \Cref{lemma:general_w_diff_bound}. 

\begin{lemma} \label{lemma:general_w_diff_bound}
    Suppose that $\bm{d}$ has a GPG of $\eps$, $\bm{w}^*$ is the optimal solution for \eqref{general:approx_dual} and $\tilde{\bm{w}}_L$ is specified in \Cref{alg:general_log}. It holds that:
    \[
    \mathbb{E}\left[\left\|\bm{w}^* - \tilde{\bm{w}}_L\right\|_2\right] \le \gamma_t\left(2Z + \frac{\eps}{2} + \frac{2Z^2}{\eps}\right)
    \]
    where $Z = \sqrt{q}\left(\overline{G} + \overline{D}\right)$.
\end{lemma}

\begin{myproof}
First, for any $t\ge1$, we have:
\[
\begin{aligned}
    \left|\left\|\bm{w}^* - \bm{w}_{t+1}\right\|_2 - \left\|\bm{w}^* - \bm{w}_t\right\|_2\right| &\le \left\|\bm{w}^* - \bm{w}_{t+1} - \left(\bm{w}^* - \bm{w}_t\right)\right\|_2 \\
    &= \left\|\bm{w}_t - \left(\bm{w}_t + \gamma_t\left(\bm{g}(\tilde{x}_t;\bm{\theta}_t)^\top\Phi - \bm{d}^\top\Phi\right)\right)^+\right\|_2 \\
    &\le \left\|\bm{w}_t - \left(\bm{w}_t + \gamma_t\left(\bm{g}(\tilde{x}_t;\bm{\theta}_t)^\top\Phi - \bm{d}^\top\Phi\right)\right)\right\|_2 \\
    &\le \gamma_tZ
\end{aligned}
\]
where the first inequality stems from the triangle inequality.
Now we illustrate that the term $\left\|\bm{w}^* - \bm{w}_t\right\|_2$ has negative drift property, which plays a key role in bounding $\left\|\bm{w}^* - \bm{w}_t\right\|_2$. Similar results have been developed in previous literature \citep{huang2009delay,he2025online} and here we conclude them in the following lemma:

\begin{lemma} \label{lem:nega_drift}
Denote $\bm{w}_t$ be the output of \Cref{alg:const_reg}, for any $t\ge1$, conditioned on any fixed $\bm{w}_t$, we have
\[
\mathbb{E}\left[\left\|\bm{w}^* - \bm{w}_{t+1}\right\|_2^2\right] - \left\|\bm{w}^* - \bm{w}_t\right\|_2^2 \le \gamma_t^2\left(\overline{G}+\overline{D}\right)^2 + 2\gamma_t\left(L_{\tilde{x}_t}(\bm{w}^*) - L_{\tilde{x}_t}(\bm{w}_t)\right)
\]
In addition, if $d$ has a GPG of $\eps$, then for any constants $\eta$ and $H$ satisfying the condition
\[
0\le\eta\le\eps,~~\text{and}~ \gamma_t^2Z^2 - 2(\eps - \eta)H\le\eta^2,
\]
as long as $\left\|\bm{w}^* - \bm{w}_t\right\|_2 \ge \gamma_tH$,
it holds that
\[
\mathbb{E}\left[\left\|\bm{w}^* - \bm{w}_{t+1}\right\|_2\right] \le \left\|\bm{w}^* - \bm{w}_t\right\|_2 - \gamma_t\eta.
\]
\end{lemma}

The full proof of \Cref{lem:nega_drift} has been presented in \Cref{appendix:general_nega_drift}. Select $\eta = \frac{\eps}{2}$ and $H = \max\{\frac{Z^2 - \eta^2}{2(\eps - \eta)},\eta\}$ and it holds that
\[
\mathbb{E}\left[\left\|\bm{w}^* - \bm{w}_{t+1}\right\|_2 - \left\|\bm{w}^* - \bm{w}_t\right\|_2|\bm{w}_t\right] \le -\gamma_t\eta
\]
Besides, we know that $\bm{w}_1 = \bm{w}^*$, which implies that $\left\|\bm{w}^* - \bm{w}_1\right\|_2 = 0$. By applying \Cref{lem:gupta}, it holds that for all $t\ge 1$,
\begin{equation} \label{ieq:w_star_t_bound}
\begin{aligned}
\mathbb{E}\left[\left\|\bm{w}^* - \bm{w}_t\right\|_2\right] &\le \gamma_tZ\left(1 + \lceil \frac{H}{Z} \rceil + \frac{Z - \eta}{2\eta}\right) \\
&\le \gamma_t\left(2Z + H + \frac{Z^2}{2\eta}\right) \\
&= \gamma_t\left(2Z + \max\{\frac{Z^2  -\eps^2/4}{\eps},\frac{\eps}{2}\} + \frac{Z^2}{\eps}\right) \\
&\le \gamma_t\left(2Z + \frac{\eps}{2} + \frac{2Z^2}{\eps}\right)
\end{aligned}
\end{equation}
which completes our proof.
\end{myproof}

Finally, with the help of \Cref{lemma:general_w_diff_bound} and \Cref{lem:general_w_upper_bound}, we can conclude that
\begin{equation} \label{ieq:bound1_final}
\begin{aligned}
\textbf{BOUND I} &\le \frac{1}{2\gamma_t}\left(\left(\left\|\tilde{\bm{w}}_L\right\|_2 + \left\|\bm{w}_{{T_{\text{refine}}}+1}\right\|_2\right) \cdot \left(\left\|\tilde{\bm{w}}_L -\bm{w}^*\right\|_2 + \left\|\bm{w}^* -\bm{w}_{{T_{\text{refine}}}+1}\right\|_2\right)\right) + \frac{T\gamma_tq}{2}\left(\overline{G}+\overline{D}\right)^2 \\
&\le \frac{1}{2\gamma_t}\left(2\overline{W} \cdot\left(\frac{1}{T} + \gamma_t\left(2Z + \frac{\eps}{2} + \frac{2Z^2}{\eps}\right)\right)\right)+ \frac{T\gamma_tZ^2}{2} \\
&= \frac{\overline{W}}{\gamma_t}\cdot\gamma_t\left(1+2Z + \frac{\eps}{2} + \frac{2Z^2}{\eps}\right) + \frac{T\gamma_tZ^2}{2}\\
&= \overline{W}\left(1+2Z + \frac{\eps}{2} + \frac{2Z^2}{\eps}\right) + \frac{T\gamma_tZ^2}{2}
\end{aligned} 
\end{equation} 
where $\overline{W} = \left(\frac{q(\overline{G}+\overline{D})^2 +4\overline{F}}{2\underline{D}} +q(\overline{G}+\overline{D})\right)$ is the upper bound of $\left\|\tilde{\bm{w}}_L\right\|_2$ and $\left\|\bm{w}_{{T_{\text{refine}}}+1}\right\|_2$, and $Z = \sqrt{q}\left(\overline{G} + \overline{D}\right)$ for simplicity.

For \textbf{BOUND II}, similar to the proof of \Cref{thm:fast}, we know that:
\[
\sum_{t=1}^{T_{\text{refine}}}\bm{g}(\tilde{x}_t;\bm{\theta}_t)^\top\Phi - \bm{d}^\top\Phi \le \frac{1}{\gamma_t}\sum_{t=1}^{T_{\text{refine}}}\bm{w}_{t+1} - \bm{w}_t = \frac{1}{\gamma_t}(\bm{w}_{T_{\text{refine}}+1} - \tilde{\bm{w}}_L)
\]
Therefore, we have:
\begin{equation} 
\label{ieq:bound2_final}
\begin{aligned} 
\textbf{BOUND II} &\le \overline{r}\sum_{j=1}^q\left[\sum_{t=1}^{T_{\text{refine}}}(\bm{g}(\tilde{x}_t;\bm{\theta}_t)^\top\Phi)_j  - (\bm{b}^\top\Phi)_j + \overline{G}\right]^+ \\
&\le \overline{r}q\left(\frac{1}{\gamma_t}\left\|\bm{w}_{T_{\text{refine}}+1} - \tilde{\bm{w}}_L\right\|_2 + \overline{G}\right) \\
&\le \overline{r}q\left(\frac{1}{\gamma_t}\left(\left\|\tilde{\bm{w}}_L -\bm{w}^*\right\|_2 + \left\|\bm{w}^* -\bm{w}_{{T_{\text{refine}}}+1}\right\|_2\right) + \overline{G}\right) \\
&\le \overline{r}q\left(1+ 2Z + \frac{\eps}{2} + \frac{2Z^2}{\eps} + \overline{G}\right)
\end{aligned}
\end{equation}
where the third inequality utilizes Cauchy Inequality and last one follows inequality \eqref{ieq:w_star_t_bound}. Combining \textbf{BOUND I} \eqref{ieq:bound1_final} and \textbf{BOUND II} \eqref{ieq:bound2_final}, we can get our final results:
\[
\begin{aligned}
\text{Reg}_{T_{\text{refine}}}(\pi_{\text{ALG7}}) &\le \textbf{BOUND I} + \textbf{BOUND II} \\
&\le \overline{W}\left(1+2Z + \frac{\eps}{2} + \frac{2Z^2}{\eps}\right) + \frac{T\gamma_tZ^2}{2} + \overline{r}q\left(1 + 2Z + \frac{\eps}{2} + \frac{2Z^2}{\eps} + \overline{G}\right) \\
&= \frac{T\gamma_tZ^2}{2} + O(\frac{q}{\eps}) \\
&\le O(\frac{q}{\eps})
\end{aligned}
\]
if we select $\gamma_t \le \frac{1}{T}$. Therefore, our final regret bound is independent of time horizon $T$ and our proof is thus completed.

\subsection{Proof of \Cref{lem:nega_drift}} \label{appendix:general_nega_drift}
For any fixed $\bm{w}_t$, we have:
\[
\begin{aligned}
&\mathbb{E}\left[\left\|\bm{w}^* - \bm{w}_{t+1}\right\|_2^2\right] - \left\|\bm{w}^* - \bm{w}_t\right\|_2^2 = \mathbb{E}_{\bm{\theta_t}\sim\mathcal{P}}\left[\left\|\bm{w}^* - \left(\bm{w}_t + \gamma_t\left(\bm{g}(\tilde{x}_t;\bm{\theta}_t)^\top\Phi - \bm{d}^\top\Phi\right)\right)^+\right\|_2^2\right] - \left\|\bm{w}^* - \bm{w}_t\right\|_2^2 \\
&\le \mathbb{E}_{\bm{\theta_t}\sim\mathcal{P}}\left[\left\|\bm{w}^* - \bm{w}_t - \gamma_t\left(\bm{g}(\tilde{x}_t;\bm{\theta}_t)^\top\Phi - \bm{d}^\top\Phi\right)\right\|_2^2\right] - \left\|\bm{w}^* - \bm{w}_t\right\|_2^2 \\
&=\mathbb{E}_{\bm{\theta_t}\sim\mathcal{P}}\left[\left\|\gamma_t\left(\bm{g}(\tilde{x}_t;\bm{\theta}_t)^\top\Phi - \bm{d}^\top\Phi\right)\right\|_2^2\right] - 2\mathbb{E}_{\bm{\theta_t}\sim\mathcal{P}}\left[\langle\bm{w}^*-\bm{w}_t,\gamma_t\left(\bm{g}(\tilde{x}_t;\bm{\theta}_t)^\top\Phi - \bm{d}^\top\Phi\right)\rangle\right]
\end{aligned}
\]
Now we consider the following Lagrangian function $L(\bm{w})$:
\[
L(\bm{w}) = \max_{x\in\mathcal{X}} \mathbb{E}_{\bm{\theta}\sim\mathcal{P}}\left[f(x;\bm{\theta}) + \bm{w}^\top\Phi^\top\left(\bm{d} - \bm{g}(x;\bm{\theta})\right)\right]
\]

Note that $\tilde{x}_t$ is the optimal solution for $L(\bm{w}_t)$ according to the Step $19$ in \Cref{alg:general_log}, while it is a feasible solution for $L(\bm{w}^*)$. Therefore, we obtain that
\[
\begin{aligned}
L_{\tilde{x}_t}(\bm{w}_t) &= \mathbb{E}_{\bm{\theta}\sim\mathcal{P}}\left[f(\tilde{x}_t;\bm{\theta}) + \bm{w}_t^\top\Phi^\top\left(\bm{d} - \bm{g}(\tilde{x}_t;\bm{\theta})\right)\right] \\
L_{\tilde{x}_t}(\bm{w}^*) &\ge \mathbb{E}_{\bm{\theta}\sim\mathcal{P}}\left[f(\tilde{x}_t;\bm{\theta}) + (\bm{w}^*)^\top\Phi^\top\left(\bm{d} - \bm{g}(\tilde{x}_t;\bm{\theta})\right)\right]
\end{aligned}
\]
which implies that
\[
\begin{aligned}
L_{\tilde{x}_t}(\bm{w}^*) - L_{\tilde{x}_t}(\bm{w}_t) &\ge \mathbb{E}_{\bm{\theta}\sim\mathcal{P}}\left[f(\tilde{x}_t;\bm{\theta}) + (\bm{w}^*)^\top\Phi^\top\left(\bm{d} - \bm{g}(\tilde{x}_t;\bm{\theta})\right)\right] - \mathbb{E}\left[f(\tilde{x}_t;\bm{\theta}) + \bm{w}_t^\top\Phi^\top\left(\bm{d} - \bm{g}(\tilde{x}_t;\bm{\theta})\right)\right] \\
&= \mathbb{E}_{\bm{\theta}\sim\mathcal{P}}\left[\langle\bm{w}_t - \bm{w}^*,\Phi^\top\left(\bm{g}(\tilde{x}_t;\bm{\theta}) - \bm{d}\right)\rangle\right]
\end{aligned}
\]
Therefore, it holds that
\[
\begin{aligned}
    &\mathbb{E}_{\bm{\theta}\sim\mathcal{P}}\left[\left\|\bm{w}^* - \bm{w}_{t+1}\right\|_2^2\right] - \left\|\bm{w}^* - \bm{w}_t\right\|_2^2 \\&\le \mathbb{E}_{\bm{\theta}\sim\mathcal{P}}\left[\left\|\gamma_t\left(\bm{g}(\tilde{x}_t;\bm{\theta})^\top\Phi - \bm{d}^\top\Phi\right)\right\|_2^2\right] - 2\mathbb{E}_{\bm{\theta}\sim\mathcal{P}}\left[\langle\bm{w}^*-\bm{w}_t,\gamma_t\left(\bm{g}(\tilde{x}_t;\bm{\theta})^\top\Phi - \bm{d}^\top\Phi\right)\rangle\right] \\
    &\le \gamma_t^2Z^2 + 2\gamma_t\left(L_{\tilde{x}_t}(\bm{w}^*) - L_{\tilde{x}_t}(\bm{w}_t)\right)
\end{aligned}
\]
which completes the proof of the first part.

Then, if $d$ has a GPG of $\eps$, assume that $\hat{d}$ satisfies the condition $\left\|\hat{\bm{d}} - \bm{d}\right\| \le \eps$, we define:
\[
\hat{L}_x(\bm{w}) = \bm{w}^\top\Phi^\top\hat{\bm{d}} + \mathbb{E}_{\bm{\theta}\sim\mathcal{P}}\left[\max_{x\in\mathcal{X}} \{f(x;\bm{\theta}) - \bm{w}^\top\Phi^\top\bm{g}(x;\bm{\theta})\}\right]
\]
Note that the only difference between $L_x(\bm{w})$ and $\hat{L}_x(\bm{w})$ is that $L_x(\bm{w})$ is based on $d$ and $\hat{L}_x(\bm{w})$ is based on on $\hat{d}$. Then, it holds that
\begin{equation} \label{eq:general_lem5_mid}
L_x(\bm{w}) - L_x(\bm{w}^*) = \hat{L}_x(\bm{w}) - \hat{L}_x(\bm{w}^*) + \langle\bm{w}^* - \bm{w},\hat{\bm{d}} - \bm{d}\rangle 
\ge \langle\bm{w}^* - \bm{w},\hat{\bm{d}} - \bm{d}\rangle \ge \eps\left\|\bm{w}^* - \bm{w}\right\|_2
\end{equation}
where the first inequality holds since $\bm{w}^*$ is also the optimal solution for the problem $\min_{\bm{w}\ge0} \hat{L}(\bm{w})$ according to our GPG assumption.

Then we can obtain that
\[
\mathbb{E}\left[\left\|\bm{w}^* - \bm{w}_{t+1}\right\|_2^2\right] - \left\|\bm{w}^* - \bm{w}_t\right\|_2^2 
\le \gamma_t^2Z^2 - 2\gamma_t\left(L_{\tilde{x}_t}(\bm{w}_t) - L_{\tilde{x}_t}(\bm{w}^*)\right)
\le \gamma_t^2Z^2 - 2\eps\gamma_t\left\|\bm{w}_t - \bm{w}^*\right\|_2
\]

Therefore, for any constants $\eta$ and $H$ satisfying the condition $0\le\eta\le\eps,~~\text{and}~ \gamma_t^2Z^2 - 2(\eps - \eta)H\le\eta^2$, as long as $\left\|\bm{w}^* - \bm{w}_t\right\|_2 \ge \gamma_tH$, it holds that
\[
\begin{aligned}
\mathbb{E}\left[\left\|\bm{w}^* - \bm{w}_{t+1}\right\|_2^2\right] &\le  \left\|\bm{w}^* - \bm{w}_t\right\|_2^2 + \gamma_t^2Z^2 - 2\eps\gamma_t\left\|\bm{w}_t - \bm{w}^*\right\|_2 \\
&= \left\|\bm{w}^* - \bm{w}_t\right\|_2^2 +  \gamma_t^2Z^2 - 2\gamma_t(\eps-\eta)\left\|\bm{w}_t - \bm{w}^*\right\|_2 - 2\gamma_t\eta\left\|\bm{w}_t - \bm{w}^*\right\|_2 \\
&\le \left\|\bm{w}^* - \bm{w}_t\right\|_2^2  - 2\gamma_t\eta\left\|\bm{w}_t - \bm{w}^*\right\|_2 + \gamma_t^2\eta^2 \\
&= \left(\left\|\bm{w}^* - \bm{w}_t\right\|_2 - \gamma_t\eta\right)^2
\end{aligned}
\]
which implies that
\[
\mathbb{E}\left[\left\|\bm{w}^* - \bm{w}_{t+1}\right\|_2\right] \le \left\|\bm{w}^* - \bm{w}_t\right\|_2 - \gamma_t\eta.
\]

Our proof is thus completed.

\end{APPENDICES}

\end{document}